\documentclass[lettersize,journal]{IEEEtran}
\usepackage{enumerate}
\usepackage[nottoc]{tocbibind}
\usepackage{enumitem,kantlipsum}
\usepackage{indentfirst}
\usepackage{multirow}
\usepackage{pifont}
\usepackage[table,xcdraw,dvipsnames,usenames]{xcolor}
\usepackage{subfloat}
\usepackage[noadjust]{cite} 
\usepackage{amsmath,amssymb,amsfonts}
\usepackage{algorithm}
\usepackage{algpseudocode}
\usepackage{caption}
\usepackage{subcaption}
\DeclareCaptionLabelFormat{ieeetable}{\MakeUppercase{#1~#2}\\}
\usepackage{textcomp}
\usepackage{stfloats}
\usepackage{url}
\usepackage{graphicx}
\usepackage{bbding}
\usepackage{rotating}
\usepackage{gensymb}
\usepackage[bottom,flushmargin]{footmisc}
\usepackage{adjustbox}
\usepackage{xcolor}
\usepackage{siunitx}
\usepackage{lineno}
\usepackage{multicol,siunitx,amssymb,enumitem,ragged2e, pifont}
\usepackage{hyperref}
\usepackage{stmaryrd}
\usepackage{trimclip}

\newcommand{\textupdate}[1]{\textcolor{black}{#1}}
\newcolumntype{K}[1]{>{\centering\arraybackslash\scriptsize}p{#1}}
\newlength{\kw}
\def\BibTeX{{\rm B\kern-.05em{\sc i\kern-.025em b}\kern-.08em
    T\kern-.1667em\lower.7ex\hbox{E}\kern-.125emX}}
\usepackage{balance}
\usepackage{xr}
\begin{document}
\title{
% Zero-Shot Anomaly Explanation for Anonymized Video Crowd Monitoring Using Alert Categorization, Spatial Localization, and Scene Context Generation

% Unified Zero-Shot Anomaly Explanation for Anonymized Crowd Video Monitoring: Context, Categorization, and Spatial Localization
Guided Zero-Shot Anomaly Explanation for Anonymized Crowd Video Monitoring: Context, Categorization, and Spatial Localization
}

\author{Mulugeta~Weldezgina~Asres,
        Lei~Jiao, \textit{Senior Member, IEEE},
        Christian~Walter~Omlin
\thanks{This work was funded by the AI4Citizens project (ID: 320783), under the Research Council of Norway.}
\thanks{M.~W.~Asres is with University~of~Agder (UiA), Norway, mulugetawa@uia.no.}   
\thanks{L.~Jiao is with UiA, Norway, lei.jiao@uia.no.  Corresponding author. 
}   
\thanks{C.~W.~Omlin is with UiA, Norway, christian.omlin@uia.no.}   
 }

% \markboth{IEEE Transactions on Information Forensics and Security}%
\markboth{}%
{Author \MakeLowercase{\textit{et al.}}: Bare Demo of IEEEtran.cls for IEEE Journals}

\maketitle

\begin{abstract}
\textupdate{Despite the promising development of deep learning for video anomaly detection (VAD), explainability (XAI) of these models remains a challenge.
Explainable and privacy-protecting VAD builds transparency and trust, and facilitates adoption of computer vision systems, including safety and security monitoring. 
However, anomaly explainability in privacy-preserving VAD systems is underexplored.  
% Approaches propose VLMs, affected by the hallucination of textual and image entanglement, resulting in high false positive and false negative errors.
In this study, we present the XAVAD approach, a unified descriptive anomaly XAI for an anonymized VAD, that integrates flexible zero-shot anonymization, anomaly type categorization, visual localization, and scene context description. 
We provide empirical and qualitative assessments of multifaceted explanations integrated with full-body visual anonymization for privacy protection.
We adopt a multimodal CLIP embedding for videos, using improved LLM-assisted prompting and pool-scoring mechanisms for anomaly-type identification. 
We enhance Grad-ECLIP for spatial anomaly heatmap localization, leveraging computation and accuracy through our novel prompt-guided patching autofocus mechanism. 
We employ VLMs to generate context and reasoning descriptions of detected anomalies through curated prompt engineering.  
We have evaluated the accuracy and computational cost of XAVAD on several publicly available CCTV and video datasets. 
Our experiments demonstrate substantial gains of our XAVAD over state-of-the-art benchmark approaches and provide insights into privacy protection trade-offs in XAI for VAD. 
We further present ablation studies and assessments across diverse configurations and modeling alternatives to demonstrate the efficacy of our proposed methods.
Code: \href{https://github.com/muleina/XAVAD}{https://github.com/muleina/XAVAD}
}

\end{abstract}


\section*{{Acronyms}} 

\newlength\mylen
\settowidth\mylen{\textsc{GroundedSA}} % widest symbol
\setlist[description]{style=nextline,      % Move description to next line
                      font==\tiny,          % Font size of description text
                      leftmargin=5.9em,       % Overall indentation from the left
                      labelsep=0.0em          % Space between label and description
                    }
\begin{description}[font=\normalfont\space\small, labelwidth=\mylen]
    \item[AN]              \small Anonymization  
    \item[AUC]             Area under the Curve   
    \item[BLEU]        Bilingual Evaluation Understudy 
    \item[CV]              Computer Vision 
     % \item[CLIP]                Pre-trained Contrastive Language Image Pretraining
     % \item[DL]              Deep Learning 
      % \item[Grad-ECLIP]       Gradient-based Visual Explanation for CLIP
    \item[GDSN]      GroundedSAM2 with Neg. Segmentation
    % \item[LLM]               Large Language Model
    \item[mAP]             Mean Average Precision 
    \item[mAR]            Mean Average Recall  
    \item[mF$_1$]            Mean F$_1$-score
     \item[MASKED]    Full-body AN using Solid Color Masking
    \item[MLLMs]           Multimodal LLMs 
     \item[PPA]           Prompt-Guided Patched Autofocus
      \item[ROUGE]        Recall-Oriented Understudy for Gisting Evaluation  
      \item[SAHI]         Slicing-Aided Hyper-Inference
     \item[SAM]       Segment Anything 
     \item[Top$k$]       Top $k^\text{th}$ Prediction
    
    % \item[UCF-Crime]              VAD UCF-Crime Dataset  
    \item[VAD]               Video Anomaly Detection  
     % \item[VLM]           Vision-Language Model
    \item[XAI]           Explainable Artificial Intelligence
    \item[XAVAD]          Our Unified XAI Approach for AN-VAD 
    % \item[XD]              VAD Violence Dataset
    % \item[YOLO]         You Only Look Once
    \item[ZS]           Zero-Shot
    \item[ZSAC]        Zero-Shot Anomaly Categorization
     \item[ZSAD]        Zero-Shot Anomaly Context Description
     \item[ZSAM]        Zero-Shot Anomaly Map
    \item[${C}$]       Set of Anomaly Category Classes 
    % \item[$H$, $W$]            Image Height, Image Width 
    % \item[$h$, $w$]             Subscript for $H$ and $W$, and $\mathbf{z}=[H, W]$  
     \item[$\mathbf{z}$]            Image Size $\mathbf{z}=[H, W]$
    \item[$\hat{\mathbf{X}}$, $\mathbf{X}$]               Video Data, Anonymized Video Data
    \item[$\hat{\mathbf{I}}$, $\mathbf{I}$]              Video Frame, Anonymized Frame 
     \item[$\mathbf{M}$]  Binary Segmentation Mask/s
      \item[$\Psi$]      Object Segmentation Model 
    \item[$\Theta$]                Image AN Function / Algorithm
    \item[$\mathcal{F}_\Theta$]             Video AN System that Applies $\Theta$ 
    \item[$U$, $Q$]             VAD Utility Task, Privacy Detection Task
    \item[$\mathcal{F}_U$]             VAD Utility Model
    \item[$\Gamma$]       Anonymized VAD System 

\item[${A}$]           Alert Score of AN-VAD
\item[$\Omega$]            Our Multifaceted Anomaly Explainer

%  \item[$A_c$]    Category Anomaly XAI
%  \item[$A_d$]    Context Anomaly XAI
% \item[$A_m$]     Heatmap Anomaly XAI
 \item[$A_c$, $A_m$, $A_d$]   Anomaly XAI for Category, Heatmap, and Context
    \item[$M$]     Number of Anomaly Classes
    \item[$T$]     Number of Consecutive Video Frames
    \item[$\mathcal{T}$] Set of Textual Descriptions for $C$
     \item[$\mathcal{F}_\text{clip}$]     Pretrained CLIP Models, $\mathcal{F}_\text{clip}=\{\mathcal{F}_{\text{text}}, \mathcal{F}_\text{image}\}$
  % \item[$\mathcal{F}_{\text{text}}$]     Pretrained CLIP Text Encoder
  \item[$\mathbf{f}^c$]      Embedded Feature of Anomaly Classes using $\mathcal{F}_{\text{text}}$
  % \item[$\mathcal{F}_{\text{image}}$]     Pretrained CLIP Image Encoder
  \item[$\mathbf{f}^v$]      Embedded Feature of Video Frames using $\mathcal{F}_{\text{image}}$

 \item[$\mathbf{s}^{v,c}_{j}$]     Cosine Similarity Scores for the $j^\text{th}$ Frame $\forall C$
 \item[$s^\text{max}_i$]           Maximum Score for the $i^\text{th}$ Anomaly Class 
 \item[$j^\text{max}_i$]          Video Frame Index of $s^\text{max}_i$

\item[$\Lambda$]   Class Probability Scores $\lambda_i \in \Lambda$ after Softmax
\item[$S_c,  \beta$]      Maximum of $\Lambda$, $S_c = \lambda^\text{max}$, Its Class Index $\beta=i^\text{max}$
\item[$j^\text{max}_{\beta}$]     Video Frame Index for XAI Anomaly Class at $\beta$ 
\item[${\mathbf{I}}^\text{max}_\beta$]      ZSAC Localized Frame at ${\mathbf{I}}^\text{max}_\beta ={\mathbf{X}}[j^\text{max}_{\beta}]$

\item[$\mathcal{G}$]                Grad-ECLIP ZSAM Function
\item[$\xi$]                        PPA Hyperparameters for Improved $\mathcal{G}$
\item[$\psi$]   Image Slicer Function using the Sliding Window $w$ 
\item[$w$]                     Sliding Window with $\text{size}=[H_w, W_w], ~\text{step}=G_w$
\item[$\eta^\text{max}_\beta$]          Heatmap Score of Spatial Anomaly XAI for $\beta$ at $j^\text{max}_{\beta}$
% \item[$\mathcal{V}$]                     VLM Model
\item[$\Phi$]                     ZSAD Inference using VLM $\mathcal{V}$
\item[$\mathcal{P}_\text{vlm}$]      VLM XAI Prompt for $\Phi$
\item[$\xi_\text{vlm}$ ]          Inference Hyperparameters of $\mathcal{V}$
\item[$N_\text{vlm}$ ]          Number of Input Video Frames for $\mathcal{V}$

\end{description}

 \section{Introduction}
\label{sec:introduction}

AI-driven video anomaly detection (VAD) has become indispensable for modern CCTV safety and security applications.
However, its effectiveness and acceptance are limited without privacy protection and interpretability~\cite{asres2025low, fioresi2023ted, liu2025privacy, abdalla2025video, gao2025evolution}. 
Conventional black-box VAD models provide detection scores indicating whether an anomaly exists, but safety-critical deployment of AI systems in public surveillance demands explainability beyond the detection score~\cite{zhang2025holmes, wu2024open, du2024uncovering, lin2026unified, doshi2023towards, song2026instance, ye2025vera}.
VAD systems can exhibit bias against specific racial, gender, age, religion, and other privacy attributes due to a lack of balanced and curated training datasets~\cite{asres2025low, fioresi2023ted}. 
Explainable artificial intelligence (XAI) in VAD aims to bridge this gap by enforcing transparency, thereby enabling trustworthy crowd monitoring solutions~\cite{zhang2025holmes, wu2024open, du2024uncovering, lin2026unified, doshi2023towards, song2026instance, ye2025vera}.
Explainability enhances human–AI collaboration by enabling security personnel to verify and understand alerts, thereby reducing false flags and improving response time~\cite{doshi2023towards, song2026instance}. 
XAI can also aid in debugging models, identifying biases, and improving robustness from a technical standpoint. 
Furthermore, as security monitoring systems increasingly intersect with legal and ethical considerations, explainability with privacy protection becomes essential for accountability, auditability, and regulatory compliance~\cite{asres2025low, lin2026unified}. 

Recent progress in multimodal models, vision-language models (VLMs), and large language models (LLMs) has led to several XAI VAD studies~\cite{zhang2025holmes, wu2024open, du2024uncovering, lin2026unified, doshi2023towards, song2026instance, ye2025vera}.
Large multimodal models, through strong visual reasoning and language interaction capabilities, shift the focus from simply generating anomaly scores to establishing a verifiable, human-interpretable rationale of the anomaly in various semantic dimensions~\cite{lin2026unified, song2026instance, ye2025vera}. 
However, we identified two major gaps: 1) limited effort in zero-shot (ZS) unified XAI that combines sematic categorization, spatial heatmap, and contextual localization of anomalies within a VAD model-agnostic framework without the heavy computation of invoking multiple large models, and 2) prior works do not adequately address privacy protection along with XAI and VAD simultaneously~\cite{lin2026unified}.
These gaps necessitate for anonymization-centered approaches that respect compliance regulations by mitigating biases and inaccuracies inherited from models and training data while fostering explainability. 
 
In this paper, we explore unified XAI approach in privacy-preserving VAD context for real-time CCTV security monitoring. 
We address privacy, accuracy, transparency and versatility challenges to enhance trust in VAD systems by describing anomalies on visually anonymized videos using holistic ZS analysis without any additional training. 
We propose a novel multi-faceted XAI study for anonymized VAD (XAVAD) that integrates flexible ZS methods for anomaly type categorization, visual localization, and scene context description. 
The approach adopts contrastive language image pretraining ({CLIP})~\cite{radford2021learning} for videos using improved LLM-assisted prompting and score pooling mechanisms across a sequence of video frames for alert type identification. 
We employ gradient-based visual explanation for CLIP (Grad-ECLIP)~\cite{chenyang_gradeclip} with our scout and focus approach for anomaly heatmap localization. 
We propose a two-stage prompt-guided patched autofocus (PPA) to improve accuracy while mitigating computational overhead for handling high video resolutions. 
The ZS efficacy of multimodal models, such as CLIP, can struggle on real-world CCTV footage that comes with varying resolutions. Image patching, slicing image into a grid, is a widely employed technique that handles size variations but comes with considerable computational overhead.  
We further engage VLMs~\cite{liu2024improved, sellergren2025medgemma, hui2024qwen2, marafioti2025smolvlm, zhang2025videollama} for anomaly context description using our curated prompt engineering design.  
The XAVAD achieves privacy protection through full-body anonymization (AN) using effective segmentation masking. We leverage ZS open-vocabulary methods~\cite{liu2024grounding, kirillov2023segment} to address limitations, such as  segmentation with unknown overlapping objects, of state-of-the-art closed approaches~\cite{redmon2016you, robinson2025rf}. 
% Furthermore, the efficacy of multimodal models, such as CLIP and VLMs, for ZS classification and captioning applications is challenged in real-world CCTV footage that comes with varying image resolutions. Image patching, splitting the image into a grid, is a widely employed technique that handles size variations but comes with considerable computational overhead.  

We have evaluated XAVAD and compared it with benchmarks on several publicly available CCTV and video datasets.
Our experiment demonstrates substantial accuracy and computational gains of our XAVAD over benchmark approaches. 
We also present comparison and ablation studies on diverse configurations and modeling alternatives to demonstrate efficacy of the proposed methods.
Our study reveals trade-offs between AN and XAI in the VAD context, discusses potential considerations that should be taken into account, and addresses challenges to improve robustness.

We highlight the key contributions of this study as follows:
\begin{itemize}
    \item We design a unified XAI for VAD that enhances alert explainability while protecting privacy and VAD quality. To the best of our knowledge, our study is the first to combine AN, VAD, and XAI (textual and visual) in a single pipeline using ZS approaches.
    \item We adapt XAI methods from images to videos, underpinned by several enhancements for real-world considerations. The approaches outperform state-of-the-art methods in accuracy, computation, and privacy protection. 
    \item We propose a PPA to substantially improve accuracy and reduce the computational cost of heatmap generation for handling sources with different video resolutions.  
    \item We propose ZS open-vocabulary negative segmentation (GDSN) that substantially improves full-body AN mask generation, and thus XAI, by mitigating concealment of overlapped objects, such as weapons.

\end{itemize}

The remainder of the paper is organized as follows. We review related explainable VAD studies and multimodal ZS modeling in Section~\ref{sec:background}. Subsequently, we present our methods and evaluation in Section~\ref{sec:methodology} and Section~\ref{sec:results_and_discussion}, respectively. Finally, we conclude the study in Section~\ref{sec:conclusion}, and Appendix~\ref{sec:appendix} provides supplementary materials on methods and results.

\section{Literature Review}
\label{sec:background}

Crowd VAD has garnered significant attention as an essential aspect of CV in the realm of smart cities~\cite{liu2025privacy, abdalla2025video}. 
Multitude innovative deep learning techniques have been introduced, consistently demonstrating superior VAD~\cite{tian2021weakly, pu2024learning, sharif2023deep, liu2022decouple}.
% Recent research focuses on weakly supervised VAD, a training using a limited set of labeled anomaly data, to address overgeneralization of semi-(un)supervised methods while lessening demand for labeled training data of supervised approaches~\cite{tian2021weakly, pu2024learning, sharif2023deep, liu2022decouple}. 
% Conventional black-box VAD models are specifically trained on VAD datasets and generate binary flags indicating whether an anomaly exists in a given video or not~\cite{tian2021weakly, pu2024learning, sharif2023deep}. 
Recent progresses are exploring beyond detection, as safety-critical AI systems necessitate explainable VAD systems~\cite{ zhang2025holmes, wu2024open, du2024uncovering, lin2026unified, doshi2023towards, song2026instance, ye2025vera}.

\subsection{Explainable Video Anomaly Detection}

Exploring the recent success of multimodal models, several VAD studies have leveraged fusing language and visual semantics for reasoning and detecting anomalies~\cite{lin2026unified, joo2023clip, pu2024learning}. 
Before the deep integration of large vision-language models, early attempts at interpretability have enforced structural priors onto neural feature maps using structured semantic representations (from object detectors) and scene graphs~\cite{doshi2023towards}. 
% These logic-based methods bring human-like interpretability by using structured semantic representations, like scene graphs and causal inference~\cite{doshi2023towards}. 
% They achieve this by analyzing foreground objects, action context, and background or environment context to make the system's reasoning transparent. 
However, these frameworks are highly complex, as generating scene knowledge is extensive for crowd monitoring. In addition, their VADs are prone to failure if the underlying object recognition models misclassify a scene element~\cite{lin2026unified}.

The rapid progress of VLMs~\cite{liu2024improved, sellergren2025medgemma, hui2024qwen2, marafioti2025smolvlm, zhang2025videollama} enable XAI through strong visual reasoning and language interaction capabilities~\cite{ye2025vera}. 
Explainable VADs shift the focus from purely generating anomaly score to establishing a verifiable, human-interpretable rationale that explains the decision across spatial, temporal, and semantic dimensions~\cite{lin2026unified, song2026instance, ye2025vera}. 
Ref.~\cite{lin2026unified} proposes a framework to bridge the gap between detection, spatial (bounding box) localization, and textual explanation. The approach involves multiple reasoning (temporal scoring and bounding box generation) and refining stages (focused on the box region for caption generation) using multiple VLMs. The computational overhead of multiple VLM queries despite being fully ZS can pose a challenge for real-time deployment. 
Moreover, even though pretrained VLMs are well-suited for VAD by describing the rationale behind specific anomalies, they often lack spatial heatmap localization~\cite{ye2025vera, song2026instance, zhang2025holmes, wu2024open, du2024uncovering}. 

Embedding entanglement of normal vs. abnormal data and/or visual vs. textual data can increase bias, misdetections, and hallucinations when employing VLMs and LLMs for VAD~\cite{wang2023unveiling, lin2026unified}. 
Foundation models can also introduce semantic biases and inaccuracies inherited from their pretraining data, further enforcing the necessity of anonymized VAD~\cite{lin2026unified}. 
In addition, complex operations like multi-caption prompting using multimodal LLMs, dense graph parsing, and online LLM inference introduce steep memory and latency overhead for real-time and edge VAD deployments~\cite{zanella2024harnessing, ye2025vera, zhang2025holmes, lin2026unified}.

Driven by safety-critical deployment requirements and regulatory compliance in public surveillance, explainable and privacy-aware VAD emerges as a distinct paradigm, a major step toward trustworthy VAD. 
Our study thus integrates AN to ensure privacy protection and mitigate potential bias in both the VAD~\cite{asres2025low, liu2025privacy} and XAI pipelines~\cite{radford2021learning, chenyang_gradeclip, sellergren2025medgemma}. 
Moreover, the study will focus on low cost methods for real-time applications, using small, low-complexity models and training-free XAI ZS approaches.
Ref.~\cite{gao2025evolution} reviews recent advances in VAD from deep neural networks to multimodal LLMs (MLLMs), and Ref.~\cite{liu2025privacy} presents privacy-preserving VAD approaches and their taxonomies. Ref.~\cite{asres2025low} analyzes the performance of different AN methods for crowd monitoring VAD applications. 
Table~\ref{tbl:xai_vad_review_benchmark} in Appendix~\ref{sec:xai_vad_review_appendix} highlights key differences of our XAI approaches against related studies.

\subsection{Multimodel Embedding for Zero-Shot Modeling}

Multimodal embedding models map different data modalities, such as text, images, audio, and video, into a shared representation space.
By enabling semantic measurement across different formats, these embedding models have wide applications in cross-modal search, image and video captioning, ZS classification, and multimodal content moderation. 

Several models exist in literature, including {CLIP}~\cite{radford2021learning} from OpenAI, {ALIGN}~\cite{jia2021scaling} and {SigLIP}~\cite{zhai2023sigmoid} from Google, and ImageBind~\cite{girdhar2023imagebind} from Meta. 
The {CLIP} models, pre-trained vision-language embedding models, align semantically related images and texts in a shared embedding space~\cite{radford2021learning}. 
These models employ vision transformer ({ViT}) or a {ResNet} networks, and trained on pairs of images and textual captions~\cite{radford2021learning}.  
Multimodal embedding models enable ZS classification by aligning image/video embeddings with class-specific textual descriptions, without requiring task-specific training~\cite{balapuwaduge2026prompt}.
The semantic matching is performed by computing cosine similarities between the image and text encodings.
The {CLIP} has demonstrated promising efficacy in establishing ZS anomaly detection~\cite{zhou2023anomalyclip, feng2026cap} and is integrated in bigger VLMs~\cite{radford2021learning}.

We will focus in this study on the {ViT} variants of the CLIP, such as ViT-B/32, ViT-B/16, ViT-L/14, and ViT-L/1433px, for better performance over the ResNet. Appendix~\ref{sec:clip_models_review_appendix} highlights further comparison among the CLIP models. 

\subsection{Zero-Shot Spatial Semantic Localization}

% Although XAI models can be achieved through intrinsically interpretable models, e.g., linear regression, decision trees, and rule-based systems, the need for higher model learning capacity leads to the complexity of black-box models. 
% Hence, post-hoc methods have been developed to provide, mostly per instance, explainability for the predicted results of black-box models~\cite{selvaraju2017grad, jung2021towards, luque2025metric, selvaraju2017grad, chefer2021generic, chefer2021generic, zhou2022extract, wang2023visual, li2025closer, chenyang_gradeclip}. 

% Since visualizing the importance of input features is a straightforward approach to interpret a model, many works visualize the internal representations of CNNs or Transformers with heat maps. 
% Most explanation methods can be categorized as: CAM methods, perturbation methods, Shapley-value methods, or attribution propagation (relevance-based) methods.

Visual saliency maps of CV models can be captured through post-hoc methods and often presented through bounding boxes~\cite{doshi2023towards, lin2026unified}, heatmaps~\cite{selvaraju2017grad, chung2024evaluating, chefer2021generic, zhou2022extract, wang2023visual, li2025closer, chenyang_gradeclip}, and segmentation maps~\cite{song2026instance, gu2025univad, bai2024one}.
The bounding boxes often employ object detection models that can have notable efficacy shortcomings~\cite{chung2024evaluating, lin2026unified}. 
The boxes lack precision, as they often occupy large spaces, including non-target areas~\cite{chung2024evaluating}. 
Segmentation maps improve boundary localization, while heatmaps provide pixel-level scoring, offering more precision~\cite{chung2024evaluating, chenyang_gradeclip}.

Class activation mapping (CAM) methods, including Grad-CAM~\cite{selvaraju2017grad}, LIFT-CAM~\cite{jung2021towards}, and SyCAM~\cite{luque2025metric}, are widely used for generating heatmaps of salient regions in CNN-based models. 
Grad-CAMs have evolved to address the constraints of CAM, such as dependence on a global pooling layer, noisy heatmaps, and scalability across a variety of architectures. 
For transformer models, attention maps leveraged with gradients have been explored for XAI~\cite{chung2024evaluating, wiegreffe2019attention, chenyang_gradeclip}.  
In addition, binary segmentation maps can explain spatial semantic context using vision-language multimodal models, such as segment anything (SAM)~\cite{kirillov2023segment, gu2025univad, song2026instance} and LISA~\cite{bai2024one}. Refs.~\cite{song2026instance, gu2025univad} incorporate SAM for ZS segmentation-based XAI for anomaly detection. Ref.~\cite{bai2024one} introduces VideoLISA, a video-based MLLM, for language-based segmentation in videos.

Several studies have proposed heatmap generation for ViT models from attention maps~\cite{zhou2022extract, wang2023visual, li2025closer, chenyang_gradeclip}. Most explanation methods, relying on classification models, can fall short in delivering satisfactory results with multimodal embedding models, such as CLIP, possibly because of the sparse attention patterns of the softmax map~\cite{chenyang_gradeclip, wang2023visual}. 
{Grad-ECLIP}~\cite{chenyang_gradeclip} proposes an attention-map visual explanation to generate high-quality heatmaps that reveal important regions or words used for CLIP’s scoring of an image-text pair. 
The method delivers better localization accuracy and requires less hyperparameter tuning, with lower computational overhead, as compared to related methods, such as GAME~\cite{chefer2021generic}, MaskCLIP~\cite{zhou2022extract}, M2IB~\cite{wang2023visual}, and CLIPSurgery~\cite{li2025closer}.
In this study, we improve the accuracy and reduce the computational overhead of Grad-ECLIP for VAD XAI in a high-resolution video setting.

\subsection{Zero-Shot Context and Reason Generation}

The vision-language models, such as caption generators and VLMs~\cite{liu2024improved, sellergren2025medgemma, hui2024qwen2, marafioti2025smolvlm, zhang2025videollama}, are gaining attraction for zero- and few-shot context description for CV systems~\cite{lin2026unified, song2026instance, ye2025vera, zhang2025holmes, wu2024open, du2024uncovering}. 

Recent small VLMs, such as SmolVLM~\cite{marafioti2025smolvlm}, InternVL~\cite{wang2025internvl3}, Gemma~\cite{sellergren2025medgemma}, Qwen-VL~\cite{hui2024qwen2}, LLava~\cite{liu2024improved}, and Video-LLaMA~\cite{lin2024video}, carry great potential for real-time vision XAI through their strong visual reasoning and language interaction. 
However, VAD studies employ VLMs not only to generate an explanation but also to detect anomalies. This entanglement can propagate bias from the VLM models into the VAD decisions, as the VLMs are predominantly trained on large, uncurated datasets from the Internet.  
Furthermore, inferencing VLM for every video stream introduces memory and latency challenges for real-time deployment. 
Anomalies are rare events, and VLMs are computationally expensive compared to conventional deep learning models. 
We lower inferencing cost by opting for lightweight VADs for real-time detection and invoking VLMs for XAI only when an anomaly is detected. 
% Furthermore, we explore open-source small VLMs with fewer than 5B parameters to address memory constraints, despite most prior works employing predominantly models with 7B or more parameters.  
\section{Methodology}
\label{sec:methodology}

This section presents the problem formulation, proposed methods, and tools employed in our XAVAD study.

\subsection{Problem Formulation}
\begin{figure}[!t]
\centering
\begin{subfigure}[]{1\columnwidth}
\centering
\includegraphics[width=1\columnwidth]{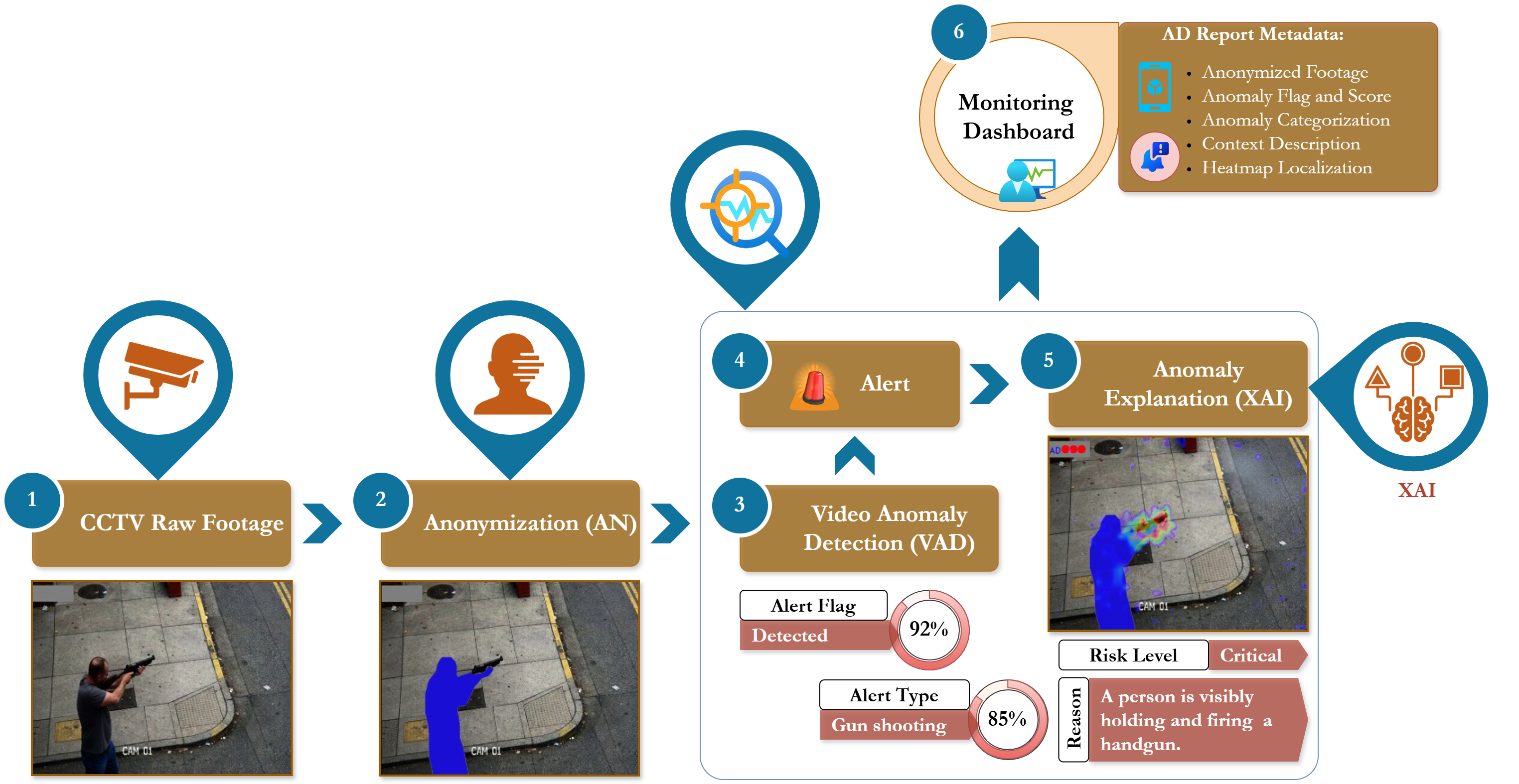} 
% \caption{}
\end{subfigure}
\caption{A unified explainable, full-body privacy-protecting VAD system.
The raw CCTV footage is AI-generated.}
\label{fig:ano_xai_diagrams_with_vis}
\vspace*{-0.5\baselineskip}
\end{figure}

\begin{figure}[!t]
\centering
\begin{subfigure}[]{1\columnwidth}
\centering
\includegraphics[width=1\columnwidth]{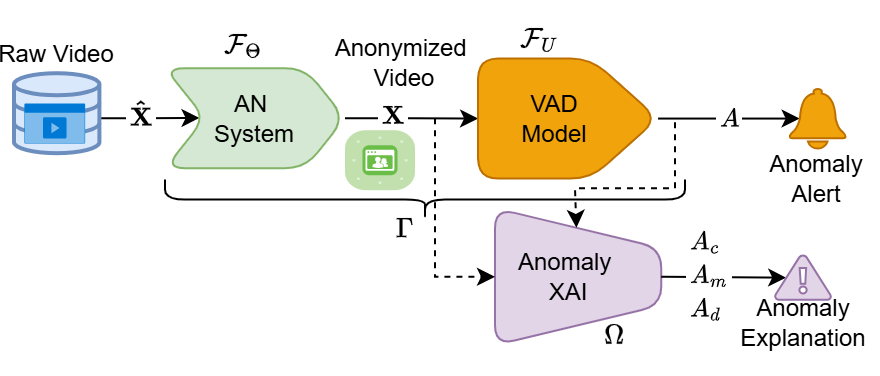} 
% \caption{}
\end{subfigure}
\caption{System pipeline diagram of the XAVAD system. The $\Gamma$ denotes the anonymized VAD system, and the $\Omega$ is the multifaceted anomaly explainer.
}
\label{fig:an_vad_xai_system_diagram}
\vspace*{-0.5\baselineskip}
\end{figure}

This study presents explainable for privacy-aware VAD framework by integrating AN, VAD, and XAI systems (see Fig.~\ref{fig:ano_xai_diagrams_with_vis}). 
Let’s consider a utility VAD task $U$, and a privacy leakage task $Q$. 
The goal of a privacy-preserving VAD system $\Gamma$, shown in Fig.~\ref{fig:an_vad_xai_system_diagram}, is to maintain the efficacy of a VAD system $\mathcal{F}_U$ while reducing $Q$ through an AN system $\mathcal{F}_{\Theta}$ as~\cite{asres2025low}: 
\begin{equation}
    {A} = \Gamma(\hat{\mathbf{X}}) : \arg \underset{Q\downarrow}{\min} \arg \underset{U\uparrow}{\max} ~\mathcal{F}_U \left(\mathcal{F}_{\Theta}(\hat{\mathbf{X}})\right),
\end{equation}
where $\hat{\mathbf{X}} \in \mathbb{Z^+}$ is raw video data, and ${A}$ is the privacy-preserving VAD alert score. 
The $\mathcal{F}_{\Theta}$ remove private sensitive information from $\hat{\mathbf{X}}$ using an AN algorithm $\Theta$, and the $\mathcal{F}_U$ predicts alert score ${A}$.

Our study focuses on full-body masking AN that applies $\Theta$ only to the most privacy-sensitive content (the human subjects), as shown in Fig.~\ref{fig:ano_xai_diagrams_with_vis}. The $\mathcal{F}_{\Theta}$ is formulated as:: 
\begin{equation}
\label{eq:an_non_adaptive}
    \mathbf{{X}} = \mathcal{F}_{\Theta} (\hat{\mathbf{X}}) : \hat{\mathbf{I}} \odot (1 - \mathbf{M}) + \mathbf{M} \odot  \Theta (\hat{\mathbf{I}}), \text{~for}~\forall \hat{\mathbf{I}} \in \hat{\mathbf{X}},
\end{equation}
{where $\hat{\mathbf{I}}$ is a frame in $\hat{\mathbf{X}}$, and $\mathbf{{X}}: \mathbf{I} \in \mathbf{X}$ is the anonymized version after applying AN. 
The $\mathbf{M} = \Psi(\hat{\mathbf{I}})$ is the binary mask of person figures, generated using a segmentation model $\Psi$. 
The $\odot$ represents element-wise multiplication.

The ZS XAI $\Omega$ explains active alerts, where ${A} > \tau$ and $\tau$ is VAD decision threshold, as: 
\begin{equation}
    {A}_c, {A}_d, {A}_m = \Omega(\mathbf{X}, {A}),
\end{equation}
where ${A}_c$, ${A}_d$, and ${A}_m$ are three different anomaly explanation outputs: anomaly type category, anomaly context description, and anomaly heatmap, respectively.

\subsection{Lightweight Full-Body Anonymization}
\label{sec:seg_method}
% The previous full-body AN studies employ a single semantically generated binary $\mathbf{M} \in \mathbb{B}^{[H \times W]}$ per image frame $\mathbf{I} \in \mathbb{Z}^{[H \times W]}$~\cite{ asres2025low, fioresi2023ted}, and applied $\Theta$ without considering the varying transformation effects of $\Theta$ on the target objects in $\mathbf{I}$ (see Fig.~\ref{fig:an_pipe_diags_nonadaptive}). This leads to poor performance in $U$ or $Q$ due to the non-optimal AN.
% In our study, we employ instantly segmented masks $\mathbf{M}={\mathbf{m_0}, \mathbf{m_1},\dots,\mathbf{m_{n-1}}}$ for each target (see Algorithm~\ref{alg:object_detection}), where $n$ is the number of masks, and the AN optimizes $\Theta$ for each target mask $\mathbf{m} \in \mathbf{M}$ (see Fig.~\ref{fig:an_pipe_diags_adaptive}).

 We semantically generated a binary mask $\mathbf{M} \in \mathbb{B}^{[H \times W]}$ per video frame $\hat{\mathbf{I}} \in \mathbb{Z}^{[H \times W]}$, and applied $\Theta$~\cite{asres2025low, fioresi2023ted} for the full-body AN.
% (see Fig.~\ref{fig:an_pipe_diags_nonadaptive}). 
We experiment with closed- and open-vocabulary object segmentation approaches (Algorithm~\ref{alg:object_detection}).

\subsubsection{Close-Vocabulary Segmentation Models}
We employ primarily the you only look once ({YOLO}) models~\cite{redmon2016you} to generate the masks. 
The {YOLO} delivers lightweight, high-performing models for object detection and segmentation applications. 
The {YOLO} models are trained on the MS-COCO dataset, and their segmentation capability is limited to the classes seen in the training set. 
% The model family includes several model versions with varying complexity and accuracy. Although the larger models provide more accurate results, we chose {YOLOv8m-seg} for our study to strike a balance between accuracy and inference speed, and to provide comparable results to previous works in Ref.~\cite{asres2025low, lee2024balancing}. 
% To handle large video resolutions, we adopt the slicing-aided hyper-inference (SAHI) method~\cite{akyon2022slicing} with the YOLO models to handle high-resolution images. SAHI splits an image into smaller patches, applies YOLO to each patch independently, and stitches the detection results back together. 
% It drastically improves the detection of small, distant, or tightly packed objects without requiring model retraining~\cite{akyon2022slicing}.
% A related study in Ref.~\cite{lee2024balancing} has similarly utilized {YOLOv8m-seg} for face and vehicle license plate AN for an object detection application.
For enhanced ZS anomaly heatmap XAI, particularly for weapon threat localization, we find that robust and precise segmentation is essential. Thus, we also experiment with large models and recently improved models from YOLO11, YOLO26~\cite{chakrabarty2026yolo26}, and RF-DETR~\cite{robinson2025rf} (see lines 2--4 in Algorithm~\ref{alg:object_detection}). 
We adopt the slicing-aided hyper-inference (SAHI) method with YOLO to handle high-resolution images, and improves detection of small, distant, or tightly packed objects without requiring model retraining~\cite{akyon2022slicing}.  
SAHI splits an image into smaller patches, applies YOLO to each patch independently, and stitches the detection results back together. 
The YOLO and RF-DETR models are trained on the MS-COCO datasets, which do not include CCTV alert objects, such as weapon threats. 
These models thus can conceal unknown (but relevant) objects when visually overlapped on a person's body.

\subsubsection{Open-Vocabulary Segmentation Models}

To address the challenges of concealment of relevant objects during AN, we employ a ZS (open-vocabulary) segmentation approach using {GroundedSAM}~\cite{ren2024grounded}. 
We augment it with a negative segmentation mechanism, hereafter denoted our approach as GDSN, to exclude objects, such as guns, knives, and bags, when visually overlap the person's body (see line 6 in Algorithm~\ref{alg:object_detection}). GroundedSAM uses {Grounding DINO}~\cite{liu2024grounding} as an open-set object detector and combines it with  SAM~\cite{kirillov2023segment} to enable ZS detection and segmentation based on arbitrary text inputs. Our GDSN employs SAM2~\cite{ravi2025sam} for improved performance, with {Grounding DINO-Tiny}~\cite{liu2024grounding}.  
The GDSN takes prompt inputs from the list of target objects to mask, i.e., $\tau_p=\{``\text{person}"\}$, and the list of negative prompts to exclude from masking, i.e., $\tau_n=\{``\text{gun}", ``\text{knife}", ``\text{bag}"\}$. The {DINO} generates bounding boxes for $\tau_p + \tau_n$, and the boxes of $\tau_n$ are supplied to the SAM as negative regions to exclude. 

% \begin{figure*}[t]
% \centering
% \begin{subfigure}[]{0.7\textwidth}
% \centering
% \includegraphics[width=1\textwidth]{images/LA3D_diagrams-vad_anonymized_non_adaptive_4.png} 
% \end{subfigure}
% \caption{System pipeline diagram of the AN ($\mathcal{F}_\Theta$) systems~\cite{asres2025low}. The $\mathcal{F}_\Theta$ applies the full-body masking AN to all the person subjects in the video frame (see Eq.~\eqref{eq:an_non_adaptive}).}
% \label{fig:an_pipe_diags_nonadaptive}
% % \vspace*{-0.5\baselineskip}
% \end{figure*}

\begin{algorithm}
\caption{$\boldsymbol{\Psi}$: Object of Interest Mask Generation}
\label{alg:object_detection}
\begin{algorithmic}[1]
\small
\State $\boldsymbol{\Psi}$ ({$\mathcal{M}, \mathbf{I}, \tau_p, \tau_n, \lambda=0.25$})
\newline
Instant object segmentation mask generation.
\newline
\Comment $\mathcal{M}$ is the seg. model: {YOLO}, {RF-DETR}, or {GDSN}.
\newline
\Comment $\mathbf{I} \in \mathbb{Z}^{[H \times W]}$ is the input video frame.
\newline
\Comment $\tau_p$ is a list of target objects to detect, e.g., a person.
\newline
\Comment $\tau_p$ is a list of objects to exclude for negative seg. of the {GDSN}, e.g., a gun, a knife, and a bag.
\newline
\Comment $\lambda$ is the detection confidence threshold.
\newline
% \State $\mathbf{z} \gets | \mathbf{I} |$ \Comment get image height and width $\mathbf{z}=[h, w]$
% \State $\bar{\mathbf{I}} \gets ImageResize(\mathbf{I}, [320, 240])$ \Comment resize image to $[320, 240]$ for enhanced segmentation
\If {$\mathcal{M} \in \{{YOLO}, {RF-DETR}\}$}
\State $i \gets ClsID(\tau_p)$ \Comment indices for $\tau_p$ from MS-COCO classes.
\State $\mathbf{M} \gets \mathcal{M}({\mathbf{I}}, i, \lambda)$ \Comment for closed models
\Else 
\State $\mathbf{M} \gets \mathcal{M}({\mathbf{I}}, \tau_p, \tau_n, \lambda)$ \Comment for opened models, {GDSN}
\EndIf
% \State $\mathbf{M} \gets ImageSizeRestore(\bar{\mathbf{M}}, \mathbf{z})$ \Comment resize masks to match the original image dimensions
% \State $\mathbf{B} \gets EstimateBoundingBox(\mathbf{M})$ \Comment generate bounding boxes surrounding $\mathbf{M}$
\newline
\Return $\mathbf{M}$ \Comment returns a list of masks.
\end{algorithmic}
\end{algorithm}

\subsection{Zero-Shot Anomaly Explanation System}
{Our proposed ZS XAI system ($\Omega$) for XAVAD incorporates three XAI components for anomaly categorization (see Section~\ref{sec:zsac_method}), heatmap localization (see Section~\ref{sec:zsam_method}), and anomaly context description (see Section~\ref{sec:zsad_method}).
Figure~\ref{fig:xai_system_diagram} illustrates the system diagrams of the $\Omega$, elaborating major building blocks: 1) XAI guiding initial configuration setup (shown in Fig.~\ref{fig:xai_config_stage_diagram}), and 2) XAI inferencing engine (shown in Fig.~\ref{fig:xai_inference_stage_diagram}). 
The initial setup precomputes shared prompts and embeddings, and the inference engine utilizes these precomputed parameters to generate anomaly XAI on the video data.
}

\begin{figure*}[]
\centering
\begin{subfigure}[]{0.9\textwidth}
\centering
\includegraphics[width=1\textwidth]{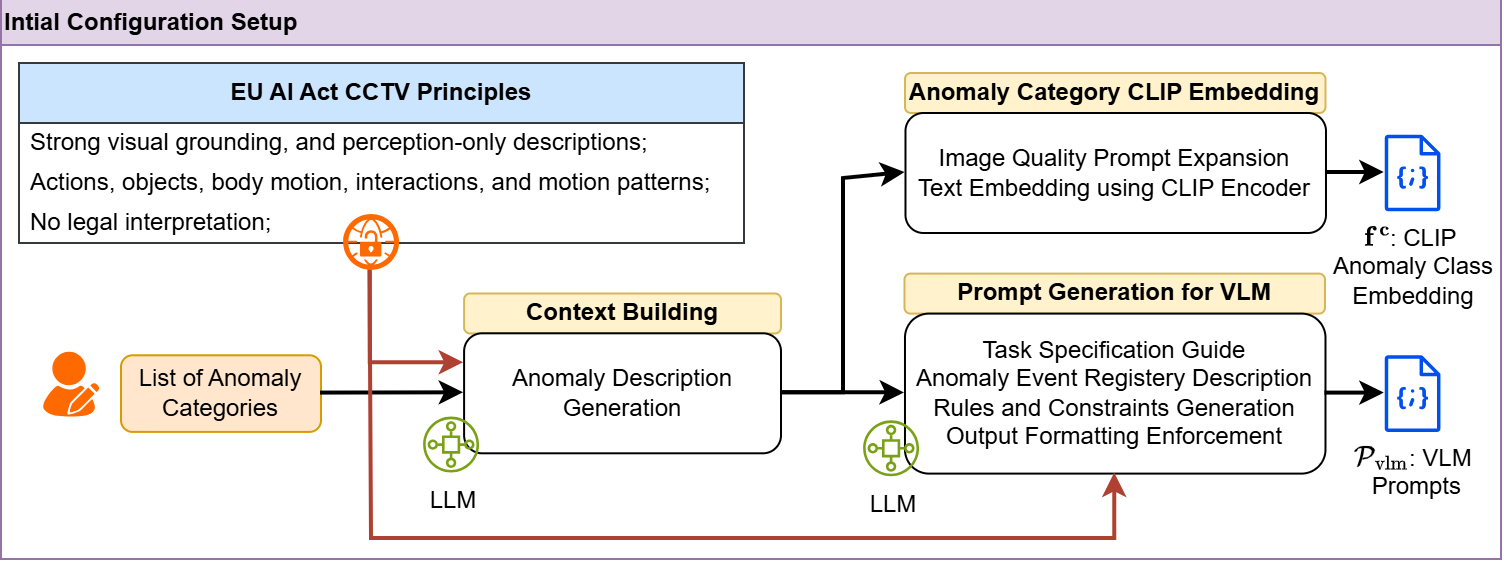} 
\caption{}
\label{fig:xai_config_stage_diagram}
\end{subfigure}
\begin{subfigure}[]{1\textwidth}
\centering
\includegraphics[width=1\textwidth]{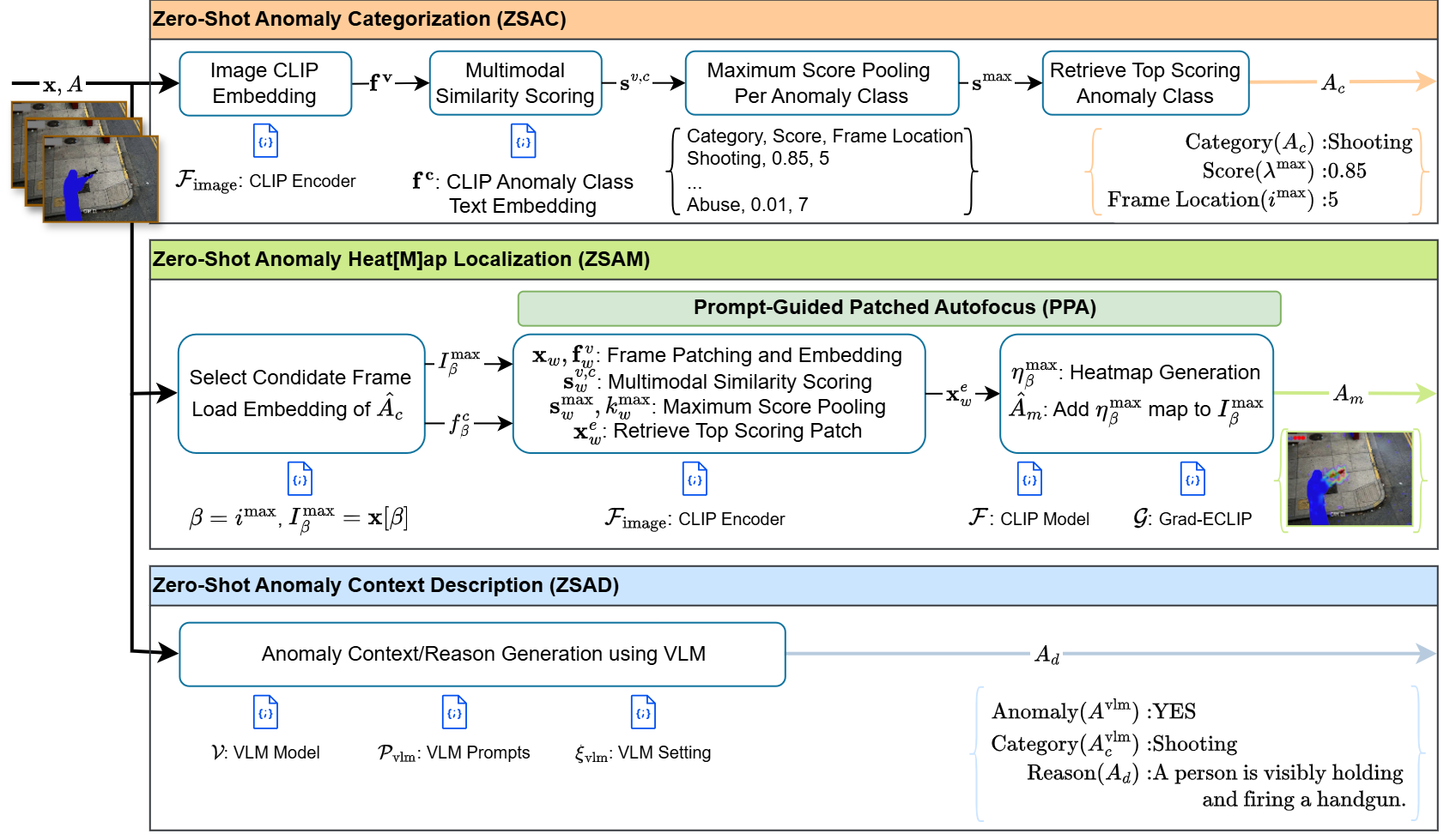} 
\caption{}
\label{fig:xai_inference_stage_diagram}
\end{subfigure}
\caption{\textupdate{System diagrams of the XAI approach of XAVAD: a) one-time guidance configuration setup, and b) XAI inference engine, generating XAI for the alerts streamed from the VAD system. The prompts and embeddings, pre-computed in a), are stored and later employed for the XAI inference. 
}
} 
\label{fig:xai_system_diagram}
\vspace*{-0.5\baselineskip}
\end{figure*}

\subsection{Zero-Shot Anomaly Categorization Approach}
\label{sec:zsac_method}

We employ ZS anomaly categorization (ZSAC) to determine the class type of the anomaly using pre-trained vision-language embedding CLIP models. 
The CLIP models, denoted as $\mathcal{F}_\text{clip}=\{\mathcal{F}_{\text{text}}, \mathcal{F}_\text{image}\}$, enable ZS classification by similarity matching image embeddings with class-specific textual descriptions, without requiring training.
% Thus, the $\Omega$ employs a ZS classifier using a CLIP model on the set of pre-defined anomaly classes $C$ to identify the anomaly category of the given alert.
Despite demonstrated promise in ZS modeling, CLIP's robustness is sensitive to domain change and text formatting~\cite{radford2021learning}. 
Furthermore, the models operate on a single image and require an adaptation mechanism to operate effectively on videos. 
The video-level scores are commonly estimated using temporal aggregation (averaging) over the embeddings or match scores across a sequence of video frames~\cite{rasheed2023fine, aghdam2025actalign}.
Our anomaly ZSAC integrates several methods, as discussed below, to improve ZS performance on video data.

\subsubsection{Anomaly Category CLIP Embedding}
The ZSAC determines the type of anomaly from a set of pre-defined $M$ anomaly categories. 
Each anomaly class category $c \in C$ is defined using text description templates, $\mathcal{T}[c]$, that are encoded by a pretrained CLIP text encoder, $\mathbf{f^c} = \mathcal{F}_{\text{text}}(\mathcal{T})$. These embeddings are typically generated from a single description per class. However, single descriptions are susceptible to noise from inadequate inclusiveness and irrelevant words that impact accuracy. 
We address this challenge by averaging the embeddings from multiple descriptions per class:
\begin{equation}
\label{eq:anml_cat_pool}
\begin{aligned}
f^c_{i} &= \frac{1}{R_i}\sum_{r=1}^{R_i} \mathcal{F}_{\text{text}}(\mathcal{T}_i[r]),\\
% f^c_{i} &= \frac{f^c_{i} }{\left\| f^c_{i} \right\|_2},
\end{aligned}
\end{equation}
where $\mathcal{T}_i$ and $f^c_{i} \in \mathbf{f}^c$ are the set of $R_i$ anomaly descriptions and the CLIP embedding of the class $c_i$, respectively.
The $f^c_{i}$ vector is further normalized by its magnitude, as $f^c_{i} = \frac{f^c_{i} }{\left\| f^c_{i} \right\|_2}$. 

Prior studies utilize definitions and related words provided by an established knowledge base to create anomaly description templates~\cite{pu2024learning, chen2024prompt}. 
Ref.~\cite{pu2024learning} has enhanced VAD using CLIP by employing {ConceptNet}~\cite{speer2017conceptnet} to expand the anomaly categories with related words and aggregate their CLIP embeddings to generate representative embeddings.
However, the keyword-based approach lacks context, as words can have multiple meanings depending on the context. 
We thus prompted LLMs (small models, such as {Llama3.2:1b}~\cite{grattafiori2024llama} and {Gemma3:1b}~\cite{sellergren2025medgemma}, can suffice for this task) to generate relevant textual descriptions for the given set of anomaly types in the crowd monitoring context (see Fig.~\ref{fig:xai_config_stage_diagram}).
LLMs have demonstrated their ability for prompt expansion and action token generation in diverse studies~\cite{wu2024open, wu2024vadclip}.
% Ref.~\cite{aghdam2025actalign} employs LLM to generate ordered sub-action words for ZS video classification in sport domain.
Ref.~\cite{wu2024open} prompts LLMs to obtain action keywords related to normal and anomaly scenarios. Ref.~\cite{wu2024vadclip} expands anomaly class names with tunable descriptions (using training) for CLIP embeddings. 

We provide details of our LLM-assisted CLIP-prompting in Appendix~\ref{sec:zsac_llm_prompt_appendix}. 
We have generated five description templates ($R_i=5$) per anomaly type, following Ref.~\cite{ye2025vera} suggestion that the optimal number of questions for VAD prompting. 
Furthermore, we have extended each description with ten image quality templates to enhance inclusiveness for quality, lighting, resolution, and depth (see Appendix~\ref{sec:zsac_llm_prompt_appendix}). 
% This results in $R_i=50$ text descriptions for each anomaly type, which are eventually aggregated into $M$ embedded features using Eq.~\eqref{eq:anml_cat_pool}.         

The text embeddings for the anomaly classes are pre-computed using Eq.~\eqref{eq:anml_cat_pool} and stored in $\mathbf{f}^c = {f^c_1, \dots, f^c_M}, ~\text{where}~\mathbf{f}^c \in \mathbb{R}^{M \times d}$. The $M$ denotes the number of classes, and $d$ is the vector dimension, e.g., $d=768$ for CLIP-ViT-L/14. 

\subsubsection{Video CLIP Embedding}
For an anonymized video snippet ${\mathbf{x}} \in \mathbb{Z}^{T \times H \times W}$, we generate embeddings $\mathbf{f}^v \in \mathbb{R}^{T \times d}$ for consecutive $T$ frames using CLIP as:
\begin{equation}
\begin{aligned}
f^v_{j} = \mathcal{F}_{\text{image}}({\mathbf{I}}_j), ~\text{for}~{\mathbf{I}}_j \in {\mathbf{x}},~\text{and}~j \in {1,..., T}, 
\end{aligned}
\end{equation}
where $f^c_{i} \in \mathbf{f}^v$ embedding of the $j^\text{th}$ frame.
Previous studies propose aggregating the frame embeddings (using average pooling) for video ZS classification~\cite{rasheed2023fine}. 
However, relevant alerts can span only a few frames and be suppressed by averaging, leading to miscategorization.
We propose extracting the maximum similarity-scoring frame not only to improve classification accuracy but also to localize the alert frame, which will later enable temporal localization during anomaly heatmap generation ($A_m$). 

\subsubsection{Zero-Shot Categorization Scoring}
Our approach involves a three-stage scoring mechanism: 
\begin{itemize}

    \item Calculating semantic matching scores for each frame across all the anomaly classes:
        \begin{equation}
        \begin{aligned}
        \mathbf{s}^{v,c} & = \text{cos}(\mathbf{f}^v,\bar{\mathbf{f}^c}) = \mathbf{f}^v\cdot\bar{\mathbf{f}^c}, %\\
       % ~\text{where} &~\mathbf{f}^v \in \mathbb{R}^{T \times d}, ~\bar{\mathbf{f}^c} \in \mathbb{R}^{d \times M},  ~\mathbf{s}^{v,c} \in \mathbb{R}^{T \times M}, 
        \end{aligned}
        \end{equation}
        where $\mathbf{f}^v \in \mathbb{R}^{T \times d}, ~\bar{\mathbf{f}^c} \in \mathbb{R}^{d \times M},  ~\mathbf{s}^{v,c} \in \mathbb{R}^{T \times M}$. 
        The element $\mathbf{s}^{v,c}_{j} \in \mathbb{R}^{1 \times M}, \text{for}~\mathbf{s}^{v,c}_{j} \in \mathbf{s}^{v,c}$, is cosine similarity scores (matrix product, $\cdot$) for the $j^\text{th}$ frame across the anomaly classes. 
        The bar in $~\bar{\mathbf{f}^c}$ denotes matrix transpose.
    
    \item Pooling maximum score per class across the frames:
        \begin{equation}
        \label{eq:anml_cat_pooling}
        \begin{aligned}
        s^\text{max}_i, j^\text{max}_i = \text{MaxPool}(\mathbf{s}^{v,c}[:,i])~\forall{i} \in {1,..., M},
        \end{aligned}
        \end{equation}
        where $s^\text{max}_i,~\text{and}~j^\text{max}_i$ are the maximum score and localized frame index for the $i^\text{th}$ anomaly class, respectively. 
        
    \item Estimating the final anomaly class with the highest scores across all the classes: 
    \begin{equation}
    \label{eq:anml_cat_final}
        \begin{aligned}
        % \Lambda = \frac{\text{exp}(\lambda_i)}{\sum^{M}_{i=1}\text{exp}(\lambda_i)}, \\
        \Lambda & = \text{softmax}(\{s^\text{max}_i\},~\forall{i}) = \frac{\text{exp}(s^\text{max}_i)}{\sum^{M}_{i=1}\text{exp}(s^\text{max}_i)}, \\
        \lambda^\text{max}, i^\text{max} & = \text{MaxPool}(\Lambda), \\
        \end{aligned}
        \end{equation}
        where $\Lambda = \{\lambda_1, \dots, \lambda_M\}$ is a class probability score vector after softmax normalization, and $\lambda^\text{max}~\text{and}~i^\text{max}$ are the maximum score and its class index, respectively. The predicted categorization results are provided at class index $\beta=i^\text{max}$, with the class label: $A_c = C[\beta]$, score: $S_c = \lambda^\text{max}$, the frame location: $j^\text{max}_{\beta}$, and the localized video frame: ${\mathbf{I}}^\text{max}_\beta ={\mathbf{X}}[j^\text{max}_{\beta}]$.
        
\end{itemize}

% Let $T$ be the number of frames in the video snippet, $M$ the number of anomaly classes, the CLIP embeddings $\mathbf{f}^v \in \mathbb{R}^{T \times d}$ for the consecutive $T$ video frames and $\mathbf{f}^c\in \mathbb{R}^{M \times d}$ for class $j$. The $d$ is the dimension of the embedding vector space, e.g., $d=512$ for the CLIP-ViT-L/32 employed in this study. 
% The cosine similarity score $s_{i,j}$ between frame $i$ and aspect text $j$ is calculated as 
% \begin{equation}
% s_{i,j} \;=\; \frac{f_i \cdot t_j}{\lVert f_i\rVert \,\lVert t_j\rVert}.  
% \end{equation}
% the CLIP embeddings $f^v_i \in \mathbb{R}^d$ for frame $i$ and $f^c_j \in \mathbb{R}^d$ for class $j$.

Several CLIP models exist, including ResNet and ViT variants. In this study, we primarily adopt the {ViT-L/14} model for its promising accuracy with moderate computational overhead. 

\subsection{Zero-Shot Anomaly Spatial Localization}
\label{sec:zsam_method}

We propose a ZS anomaly map (ZSAM) method for spatial explanation of the VAD. 
We adopt Grad-ECLIP~\cite{chenyang_gradeclip} to highlight the saliency heatmap of the anomaly category predicted by our ZSAC. 
Grad-ECLIP produces high-quality heatmaps that explain matching image-text pairs.
% Chenyang et al.~\cite{chenyang_gradeclip} have demonstrated the superiority of the proposed Grad-ECLIP with comprehensive evaluations, comparing it with the state-of-the-art explanation methods. 
Among alternative approaches, we found Grad-ECLIP effective for interpreting ViT CLIP models by highlighting influential features measured through gradients~\cite{chenyang_gradeclip}.
% The explanation utilizes a feature map from the last layer of the $mathcal{F}_{\text{image}}$.
As the mathematical formulation details of Grad-ECLIP are provided in Ref.~\cite{chenyang_gradeclip}, we thus focus below on our modifications that enable computation efficiency and precise anomaly saliency map generation. 

Generating a heatmap for video requires additional effort, such as 1) identifying the relevant video frame (s) for the target anomaly category, and 2) effectively capturing a small but relevant region of interest, such as weapons, which can be hard to detect, as CCTV videos cover large areas~\cite{bai2024one}.
Heatmap explainers struggle to distinguish small objects like a handgun and a knife from ordinary non-threat personal items after the frame is downsized during CLIP preprocessing. 
% Ref.~\cite{zhou2023zegclip, akyon2022slicing} utilize smaller window patches over the frame to mitigate this challenge in related domains. 
Ref.~\cite{zhou2023zegclip} proposes {ZegCLIP} for ZS segmentation using patch slicing, and Ref.~\cite{akyon2022slicing} presents {SAHI} for handling high-resolution images by slicing into patches and sticking back for object detection and segmentation. 
Nevertheless, the above approaches incorporate components that require training and/or are can be expensive depending on the number of patch slices.
% Moreover, finding the optimal sliding window size can be challenging. The size impacts the accuracy, e.g., the presence of large vs. small objects of interest, as well as the computational overhead. 
 
 We achieve frame localization using the maximum ZSAC scoring frame at $j^\text{max}_{\beta}$, generated from Eq.~\eqref{eq:anml_cat_final}. The assumption is that the video frame with the highest match to the predicted anomaly category $A_c$, i.e., ${\mathbf{I}}^\text{max}_\beta =\mathbf{X}[j^\text{max}_{\beta}]$, presents sufficient visual evidence for the anomaly.
 We propose a two-stage prompt-guided patched autofocus (PPA) localization approach to improve accuracy and computation: 1) A lightweight prompt matching (using CLIP, similar to ZSAC) narrows down the search space to the most relevant patch slice, and 2) a heatmap generator, e.g., Grad-ECLIP, performs ZSAM on the candidate patch to produce the anomaly saliency heatmap.
    % using large and small-sized windows. 
  
Our PPA firstly employs a sliding window $w: \text{size}=[H_w, W_w], ~\text{step}=G_w$ over the video frame ${\mathbf{I}}^\text{max}_\beta$ and generates $N_w$ patches. It then identifies the candidate patch as:
\begin{equation}
\begin{aligned}
{\mathbf{x}}_w &= {\psi}({\mathbf{I}}^\text{max}_\beta, w), \\
 \mathbf{f}_w^v &: {f}_{w,k}^v = \mathcal{F}_{\text{image}}({\mathbf{x}}_w[k]),~\text{for}~k \in {1,..., N_w}, 
\end{aligned}
\end{equation}
where $\psi$ is an image grid slicer function with $w$, and $\mathbf{f}_w^v= \{{f}_{w,1}^v, \dots, {f}_{w,N_w}^v\}$ is the CLIP embedding of the patches.
\begin{equation}
    \label{eq:anml_heatmap_pooling_cat}
    \begin{aligned}
     s^{v,c}_w& = \text{cos}(\mathbf{f}^v_w, \bar{{f}^c_\beta}) = \mathbf{f}^v_w\cdot\bar{{f}^c_\beta},\\
    s^\text{max}_w, k^\text{max}_w & = \text{MaxPool}(s^{v,c}_w),\\
    ~\text{where} &~\mathbf{f}^v_w \in \mathbb{R}^{N_w \times d}, ~\bar{{f}^c_\beta} \in \mathbb{R}^{d \times 1},
    \end{aligned}
\end{equation}
where $s^\text{max}_w,~\text{and}~k^\text{max}_w$ are the score and location of the highest matching candidate patch of ${\mathbf{I}}^\text{max}_\beta$ with $f^c_{\beta}$, respectively. 
The $f^c_{\beta}$ is the text embedding of the target category $A_c$, predicted from the ZSAC. 
We follow a similar approach to Eq.~\eqref{eq:anml_cat_pooling} with pooling executed over the image grid slices with a single anomaly class $A_c$. 
We employ CLIP text-image matching instead of invoking the heatmap generator for patch localization to enhance accuracy and computation. 

% At the second stage, PPA applies a heatmap generation algorithm $\mathcal{G}$, i.e., Grad-ECLIP, for the given CLIP model $\mathcal{F}$ on the $k^\text{max}_w$ image using a small sliding window $w^e: \text{size}=[H^e_w, W^e_w],~\text{step}=G_w$ (slicing into ${\mathbf{x}}_w^e$ patches) for refined anomaly saliency map generation:  
% \begin{equation}
%     \label{eq:anml_heatmap_final}
%     \begin{aligned}
%     {\mathbf{x}}_w^e &= {\psi}({\mathbf{x}}_w[k^\text{max}_w], w^e), \\
%     \eta^\text{max}_\beta & = \text{MaxPool}\left(\mathcal{G}(\mathcal{F}, \bar{{f}^c_\beta}, {\mathbf{x}}_w^e, \xi) \right), \\
%     A_m &= \mathcal{O}({\mathbf{I}}^\text{max}_\beta, \eta^\text{max}_\beta),
%     \end{aligned}
% \end{equation}
% where $\eta^\text{max}_\beta$ is the spatial heatmap score of the anomaly, and $\mathcal{O}$ is the image-blending (overlay) function that highlights the heatmap on top of the video frame. 
% The $\eta^\text{max}_\beta$ is the anomaly saliency heatmap that visually localizes the anomaly type of $A_c$ at the $\beta^\text{th}$ frame of the video ${\mathbf{x}}$.  
% The set $\xi = \{\xi_\text{min}, \xi_\text{max}, \bar{\xi}_\text{min}\}$ comprises hyperparameters we integrated to improve $\mathcal{G}$. 

At the second stage, PPA applies a heatmap extraction algorithm $\mathcal{G}$, i.e., Grad-ECLIP with $\mathcal{F}_\text{clip}$, on the candidate patch $\mathbf{x}^e_w=\mathbf{x}_w[k^\text{max}_w]$ for a refined saliency map:  
\begin{equation}
    \label{eq:anml_heatmap_final}
    \begin{aligned}
    \eta^\text{max}_\beta & = \mathcal{G}(\mathcal{F}_\text{clip}, \bar{{f}^c_\beta}, \mathbf{x}^e_w, \xi), \\
    A_m &= \mathcal{O}({\mathbf{I}}^\text{max}_\beta, \eta^\text{max}_\beta),
    \end{aligned}
\end{equation}
where $\eta^\text{max}_\beta$ is the spatial heatmap score of the anomaly, and $\mathcal{O}$ is the image-blending function that highlights the heatmap on top of the frame. 
The $\eta^\text{max}_\beta$ localizes the saliency of the anomaly type of ${A}_c$ at the $\beta^\text{th}$ frame of the video ${\mathbf{x}}$.  
The set $\xi = \{\xi_\text{min}, \xi_\text{max}, \bar{\xi}_\text{min}\}$ are hyperparameters to tune $\mathcal{G}$. 

Grad-ECLIP converts gradient scores $s_e$ directly into a heatmap using \textit{min-max} scaling, which can lead to poor localization in small-score and outlier-score scenarios. When the maximum score is small, it artificially amplifies (stretches) scores, whereas outlying very high scores shrink other relevant regions.
We found that small peak scores are often associated with insufficient text-image matching, and we clean them using $\xi_\text{min}$.
Thus, we handle the scaling using:

\begin{equation}
    \label{eq:anml_heatmap_final}
    \begin{aligned}
        {s}_\text{ppa} &= \begin{cases} 
                    0, & \text{if } s_e < \xi_\text{min}, \\
                    \xi_\text{max}, & \text{if } s_e > \xi_\text{max}, \\
                \end{cases} \\
        \bar{s}_\text{ppa} &= \text{MinMaxScaler}({s}_\text{ppa}),
    \end{aligned}
\end{equation}
where $\eta^\text{max}_\beta = \bar{s}_\text{ppa}: \bar{s}_\text{ppa} > \bar{\xi}_\text{min}$. The ${s}_\text{ppa}$ and $\bar{s}_\text{ppa}$ are heatmap scores of our PPA before and after scaling, respectively.

\subsection{Zero-Shot Anomaly Context-Description Generation}
\label{sec:zsad_method}

We employ small open-source VLM models to reason about the context of detected anomalies using our ZS anomaly description generation (ZSAD) approach.
Although there are several alternative small VLMs, such as SmolVLM (256M, 512M, and 2.2B)~\cite{marafioti2025smolvlm}, InternVL (4B and 8B)~\cite{wang2025internvl3}, LLava (7B)~\cite{liu2024improved}, Video-LLaMA (2B)~\cite{zhang2025videollama}, Gemma (4B)~\cite{sellergren2025medgemma}, and Qwen-VL (0.8B, 2B, 4B, and 9B)~\cite{hui2024qwen2}, we will primarily employ InternVL models for their speed, accuracy, and capability of following structured output. 
The smaller SmolVLMs perform poorly due to hallucinations and small context windows.
We found that Qwen-VL sometimes struggles to follow the output format, and LLaVA can deliver inadequate performance.

We design structured prompts for VLMs that enforce task specification to the XAI rules, constraints, and output formatting (see Fig.~\ref{fig:xai_config_stage_diagram}). 
We formulate the ZSAD function $\Phi$, operating on the input frames ${\mathbf{x}}$, as:
\begin{equation}
    {A}^\text{vlm}, {A}^\text{vlm}_c, {A}_d = \Phi(\mathcal{V}, \mathcal{P}_\text{vlm}, \xi_\text{vlm}, {\mathbf{x}}),
\end{equation}
where $\mathcal{V}$ is the VLM model, $\mathcal{P}_\text{vlm}$ is VLM XAI prompt, and $\xi_\text{vlm}$ is the inference settings for the $\mathcal{V}$.
Our ZSAD returns the anomaly status flag ${A}^\text{vlm}$ and category ${A}^\text{vlm}_c$ besides the context explanation ${A}_d$. Although we focus on ${A}_d$, having ${A}^\text{vlm}$ and ${A}^\text{vlm}_c$ can provide further elucidation on the VLM reasoning flow. Moreover, they could be utilized to spot inconsistencies compared to the VAD (${A}$) and ZSAC (${A}_c$) results.

The settings of the VLM ($\mathcal{P}_\text{vlm}, \xi_\text{vlm}$) are provided in Appendix~\ref {sec:zsad_appendix}. 
The VLM prompts, along with the generation parameters ($\xi_\text{vlm}$: using low $\text{temperature} \leq 0.2$,  $\text{top}\_k \leq 10$, and $\text{top}\_p \geq 0.9$), are designed to improve visual grounding, limit hallucination, and enforce factual generation that is based on the presented evidence. 
We uniformly sample $N_\text{vlm}$ frames from ${\mathbf{x}}$ to generate the anomaly context.
\section{Results and Discussion}
\label{sec:results_and_discussion}

This section presents performance evaluation results of the XAVAD on several public datasets, including XD Violence~\cite{Wu2020not}, UCF-Crime~\cite{sultani2018real}, Simuletic CCTV~\cite{simuletic_weapon_2025, simuletic_knife_2025}, and HIVAU-70k~\cite{zhang2025holmes}. 
We follow the standard test-set split for the XD Violence, UCF-Crime, and HIVAU-70k, while keeping the entire data for Simuletic due to the small dataset size. 
% We implement and compare the widely utilized privacy-protection techniques, presented in Table~\ref{tbl:anony_methods}, along with our approaches. 
% We will also compare the efficacy of our approach with recent methods.

\subsection{Evaluation Approach and Performance Metrics}

Visual AN can be evaluated in different ways to assess its efficacy in privacy protection and retention of non-sensitive utilities after de-identification~\cite{zhang2024understanding, fioresi2023ted, hukkelaas2023deepprivacy2, asres2025low}.
Since our prior study in~\cite{asres2025low} presents in-depth analyses of the trade-off between privacy protection and VAD, this study focuses on retention quality for VAD-XAI.
Due to the lack of adequate public datasets that incorporate full annotations, we will follow a cross-dataset evaluation for the different XAI modalities~\cite{doshi2023towards, lin2026unified, asres2025low}. 
In addition to the empirical assessment, we will present several qualitative illustrations, comparing XAI before (no-AN) and after applying AN (MASKED-AN). 
To provide decent efficacy and be aligned with previous outcomes in Ref.~\cite{asres2025low, lee2024balancing}, the MASKED-AN primarily employs the {YOLO8-M} model, unless explicitly specified, as in~~\ref{sec:zsam}.

We will evaluate the ZSAC using multi-class classification metrics, i.e., \textit{mean $F_1$-score} (${mF}_1$) on the XD Violence~\cite{Wu2020not} and UCF-Crime~\cite{sultani2018real} datasets. 
The $\text{mF}_1$ is an average score of the \textit{$F_1$-scores} over anomaly class labels. 
% We also provide a \textit{weighted $F_1$-score} (denoted as ${F}_1$) for an aggregate average score after each \textit{$F_1$-score} per class is weighted by its sample size.
Moreover, we will present confusion matrices to provide semantic association insights among the anomaly categories. 

We will utilize Simuletic CCTV weapon and knife datasets~\cite{simuletic_weapon_2025, simuletic_knife_2025} to evaluate heatmaps of ZSAM using object detection metrics, such as \textit{average precision} (\textit{AP}) and \textit{average recall} (\textit{AR}). 
The overlap between ground-truth and predicted maps is measured using \textit{intersection over union} (IoU). 

{We will measure metrics including \textit{BLEU}~\cite{papineni2002bleu}, \textit{ROUGE}~\cite{lin2004rouge}, and \textit{BERTScore}~\cite{zhang2019bertscore} over the HIVAU-70k dataset~\cite{zhang2025holmes} to evaluate the anomaly context of ZSAD.}

We will assess the computational cost in GPU peak memory and processing time across the XAI models and methods.

{Generally, higher scores in the above accuracy metrics represent higher robustness in preserving the data utility quality for VAD XAI. 
In the results tables and figures below, we will use arrows (e.g., $\uparrow$ and $\leftarrow$) for the direction of better scores, and \textbf{bold font} for the best score. 
We will denote $\Delta (\%)$ to indicate relative performance change after applying AN, i.e., $\Delta (\%)~=~\frac{100\times(\text{score[MASKED-AN]}-\text{score[No-AN]})}{\text{score[No-AN]}}$. 

\subsection{Evaluation on Anomaly Type Categorization}
\label{sec:zsac}

We evaluate the ZSAC using large-scale public VAD datasets: the XD Violence~\cite{Wu2020not} and UCF-Crime~\cite{sultani2018real}. 
% The XD Violence dataset, the largest multi-scene weakly labeled VAD dataset to date, comprises a total duration of 217 hours from 4754 videos~\cite{Wu2020not}. The test set includes 800 videos, totaling over 2.3 million frames, and covers 6 anomaly categories related to violence. 
The XD Violence test set includes 800 multi-scene videos, totaling over 2.3 million frames, and covers 6 anomaly categories related to violence~\cite{Wu2020not}.
% The videos were collected from a variety of sources, including different types of cameras, movies, and games, which creates a mix of scenes that increases VAD complexity. 
% The UCF-Crime dataset comprises 128 hours of untrimmed CCTV videos from 1,900 sources~\cite{sultani2018real}. The videos capture 13 crime anomaly categories featuring real-world incidents. The test set comprises 290 videos, totaling over 1 million frames. 
The UCF-Crime test set comprises 13 crime anomaly categories featuring real-world incidents in 290 videos, totaling over 1 million frames~\cite{sultani2018real}. 
We evaluate across the datasets' categories (on videos with anomalies are considered) using CLIP models, such as ViT-B/32, ViT-B/16, ViT-L/14, and ViT-L/1433px.
% Following typical practice in VAD studies, we consider video clips with $T=16$ frames for the ZSAC assessment after the raw videos are split into non-overlapping video snippets. 

Table~\ref{tbl:xai_vad_cat_xd_mf1} presents ZSAC on the XD Violence dataset, where the ViT-L14 models outperform the ViT-B models in mF1 score over 7 categories. Despite the ViT-L/14@336px lead, the gain over the ViT-L/14 is marginal. Among the anomaly text prompting mechanisms, the LLM-Prompt achieves $mF_1@\text{Top}1$ spanning $0.569\text{--}0.689$ and consistently outperforms the Keyword-Prompt with $0.469\text{--}0.516$ and ConceptNet with $0.542\text{--}0.656$ across all CLIP models with No-AN. For the MASKED-AN, the $\Delta mF_1@\text{Top}1$ drops by $7.28\%\text{--}12.74\%$, $3.45\%\text{--}5.69\%$, and $5.29\%\text{--}10.19\%$, respectively. 
% The worst degradation $\Delta mF_1@\text{Top}1:~5.10\%\text{--}12.74\%$ is observed on the ViT-B/16, while the ViT-L/14 maintains the least drop within $3.45\%\text{--}7.28\%$. 
The ViT-L/14 holds the smallest drop of $3.45\%\text{--}7.28\%$.
% Figure~\ref{fig:xai_vad_cat_xd_mf1} depicts the overall performance comparison across CLIP models with and without AN. 
% The majority of the performance drop comes from the confusion between \textit``Abuse"} vs. \textit{``Fighting"}, and \textit{``Abuse"} vs. \textit{``Shooting"} (as shown in Fig.~\ref{fig:xai_vad_cat_xd_cm} in Appendix~\ref{sec:zsac_appendix}). This is not unexpected, as the categories are closely related, and some videos contain multiple anomaly scenes in the XD dataset. This has been demonstrated by the substantial increase in accuracy when considering the next top categories (at $\text{Top}2$ and $\text{Top}3$) for both No-AN and MASKED-AN. 
% % The AN trade-off improves to $\%$ he ViT-L/14 demonstrates that 

Table~\ref{tbl:xai_vad_cat_ucf_mf1} presents the performance on the UCF-Crime dataset over 14 categories. Due to the low video quality of the dataset and a higher number of categories, we analyze the performance at $\text{Top}1$, $\text{Top}3$, and $\text{Top}5$. 
ViT-L14 models still outperform the ViT-B models in the mF1 score. The LLM-Prompt generally outperforms the Keyword-Prompt and ConceptNet-Prompt, but with a limited gain. Nevertheless, the LLM achieves stable performance across all the models in both No-AN and MASKED-AN, unlike the others, which exhibit dips on the ViT-B/16 model. 
For example, the $\Delta mF_1@\text{Top}3$ after AN falls in the range of $3.51\%\text{--}9.17\%$, $3\%\text{--}20.84\%$, and $6.17\%\text{--}15.90\%$ for the LLM, Keyword, and ConceptNet prompts, respectively. 
% The ConceptNet-Prompt performs the worst, while the Keyword-Prompt achieves competitive accuracy with the LLM, particularly on the larger ViT-L14 models. 
We provide further results on weighted F1 scores, considering sample size per anomaly class, in Appendix~\ref{sec:zsac_appendix} (see Table~\ref{tbl:xai_vad_cat_xd_f1} and Table~\ref{tbl:xai_vad_cat_ucf_f1}).

\begin{table*}[]
\centering
\caption{Performance of ZSAC on the XD Violence Dataset~\cite{Wu2020not}.}
\resizebox{1\textwidth}{!}{
\begin{tabular}{lcccccccccc}
\hline
\multirow{2}{*}{CLIP Model} & \multirow{2}{*}{Prompt Mode} & \multicolumn{3}{c}{${mF}_1$@Top$1\uparrow$}                       & \multicolumn{3}{c}{${mF}_1$@Top$2\uparrow$}                       & \multicolumn{3}{c}{${mF}_1$@Top$3\uparrow$}                       \\  \cline{3-11} 
                             &                               & No-AN          & MASKED-AN      & $\Delta (\%)$  & No-AN          & MASKED-AN      & $\Delta (\%)$  & No-AN          & MASKED-AN      & $\Delta (\%)$  \\ \hline

{ViT-B/32}                          & Keyword-Prompt                                    & 0.469                     & 0.442                         & -5.69\%                            & 0.597                     & 0.561                         & -6.03\%                            & 0.696                     & 0.669                         & -3.91\%                            \\
                                                 & ConceptNet-Prompt~\cite{pu2024learning}                                 & 0.542                     & 0.501                         & -7.66\%                            & 0.733                     & 0.662                         & -9.75\%                            & 0.817                     & 0.741                         & -9.29\%                            \\
                                                 & LLM-Prompt                                        & 0.569                     & 0.516                         & -9.25\%                            & 0.739                     & 0.666                         & -9.89\%                            & 0.826                     & 0.763                         & -7.63\%                            \\
{ViT-B/16}                          & Keyword-Prompt                                    & 0.474                     & 0.450                         & -5.10\%                            & 0.593                     & 0.575                         & -2.95\%                            & 0.668                     & 0.672                         & 0.55\%                             \\
                                                 & ConceptNet-Prompt~\cite{pu2024learning}                                 & 0.577                     & 0.518                         & -10.19\%                           & 0.713                     & 0.626                         & -12.21\%                           & 0.805                     & 0.699                         & -13.12\%                           \\
                                                 & LLM-Prompt                                        & 0.625                     & 0.545                         & -12.74\%                           & 0.774                     & 0.699                         & -9.68\%                            & 0.855                     & 0.802                         & -6.23\%                            \\

{ViT-L/14}                          & Keyword-Prompt                                    & 0.516                     & 0.498                         & \textbf{-3.45\%}                   & 0.628                     & 0.620                         & \textbf{-1.25\%}                   & 0.697                     & 0.710                         & \textbf{1.87\%}                    \\
                                                 & ConceptNet-Prompt~\cite{pu2024learning}                                 & 0.645                     & 0.611                         & -5.29\%                            & 0.778                     & 0.750                         & -3.70\%                            & 0.853                     & 0.839                         & -1.66\%                            \\
                                                 & LLM-Prompt                                        & 0.683                     & \textbf{0.633}                & -7.28\%                            & 0.811                     & 0.754                         & -7.04\%                            & 0.874                     & 0.825                         & -5.64\%                            \\
{ViT-L/14@336px}                     & Keyword-Prompt                                    & 0.511                     & 0.491                         & -4.02\%                            & 0.620                     & 0.611                         & -1.46\%                            & 0.687                     & 0.694                         & 1.05\%                             \\
                                                 & ConceptNet-Prompt~\cite{pu2024learning}                                 & 0.656                     & 0.619                         & -5.57\%                            & 0.775                     & \textbf{0.755}                & -2.51\%                            & 0.858                     & \textbf{0.841}                & -1.98\%                            \\
                                                 & LLM-Prompt                                        & \textbf{0.689}            & 0.630                         & -8.54\%                            & \textbf{0.813}            & 0.752                         & -7.43\%                            & \textbf{0.875}            & 0.823                         & -5.92\%                           
 \\ \hline 

\end{tabular}
}
\noindent\footnotesize{\par  
% The $\Delta (\%) = \frac{100 \times (\text{score[MASKED-AN]} - \text{score[No-AN]})}{\text{score[No-AN]}}$.
We use $T=16$ consecutive video frame snippets.
\par}
\label{tbl:xai_vad_cat_xd_mf1}
\vspace*{-0.5\baselineskip}
\end{table*}

\begin{table*}[]
\caption{Performance of ZSAC on the UCF-Crime Dataset~\cite{sultani2018real}.}
\resizebox{1\textwidth}{!}{
\begin{tabular}{lcccccccccc}
\hline
\multirow{2}{*}{CLIP Model} & \multirow{2}{*}{Prompt Mode} & \multicolumn{3}{c}{${mF}_1$@Top$1\uparrow$}                       & \multicolumn{3}{c}{${mF}_1$@Top$3\uparrow$}                       & \multicolumn{3}{c}{${mF}_1$@Top$5\uparrow$}                       \\  \cline{3-11}                              
&                               & No-AN          & MASKED-AN      & $\Delta (\%)$  & No-AN          & MASKED-AN      & $\Delta (\%)$  & No-AN          & MASKED-AN      & $\Delta (\%)$  \\ \hline
{ViT-B/32}                          & Keyword-Prompt                                    & 0.242                     & 0.216                         & -10.68\%                           & 0.423                     & 0.410                         & \textbf{-3.00}                   & 0.599                     & 0.594                         & \textbf{-0.81\%}                   \\
                                                 & ConceptNet-Prompt~\cite{pu2024learning}                                 & 0.244                     & 0.237                         & \textbf{-2.85}                   & 0.419                     & 0.394                         & -6.17\%                            & 0.639                     & 0.610                         & -4.56\%                            \\
                                                 & LLM-Prompt (Ours)                                       & 0.258                     & 0.235                         & -8.90\%                            & 0.476                     & 0.432                         & -9.17\%                            & 0.595                     & 0.582                         & -2.06\%                            \\
{ViT-B/16}                          & Keyword-Prompt                                    & 0.236                     & 0.228                         & -3.40\%                            & 0.485                     & 0.442                         & -8.97                            & 0.627                     & 0.611                         & -2.59\%                            \\
                                                 & ConceptNet-Prompt~\cite{pu2024learning}                                 & 0.237                     & 0.211                         & -10.95\%                           & 0.418                     & 0.376                         & -9.96                            & 0.568                     & 0.544                         & -4.18\%                            \\
                                                 & LLM-Prompt (Ours)                                       & 0.241                     & 0.223                         & -7.66\%                            & 0.523                     & 0.504                         & -3.51\%                            & 0.727                     & \textbf{0.691}                & -4.91\%                            \\

{ViT-L/14}                          & Keyword-Prompt                                    & \textbf{0.316}            & 0.247                         & -22.06\%                           & \textbf{0.562}            & 0.445                         & -20.84\%                           & 0.728                     & 0.673                         & -7.56\%                            \\
                                                 & ConceptNet-Prompt~\cite{pu2024learning}                                 & 0.247                     & 0.235                         & -4.83\%                            & 0.471                     & 0.396                         & -15.90\%                           & 0.662                     & 0.647                         & -2.26\%                            \\
                                                 & LLM-Prompt (Ours)                                       & 0.301                     & 0.275                         & -8.74\%                            & 0.558                     & \textbf{0.515}                & -7.80\%                            & 0.733                     & 0.671                         & -8.40\%                            \\
{ViT-L/14@336px}                     & Keyword-Prompt                                    & 0.292                     & 0.248                         & -14.96\%                           & 0.540                     & 0.429                         & -20.56\%                           & \textbf{0.764}            & 0.685                         & -10.42\%                           \\
                                                 & ConceptNet-Prompt~\cite{pu2024learning}                                 & 0.262                     & 0.238                         & -9.17\%                            & 0.454                     & 0.383                         & -15.52\%                           & 0.661                     & 0.638                         & -3.44\%                            \\
                                                 & LLM-Prompt (Ours)                                       & 0.299                     & \textbf{0.276}                & -7.48\%                            & 0.519                     & 0.493                         & -4.95\%                            & 0.715                     & 0.632                         & -11.68\%                          
                             \\ \hline
\end{tabular}
}
\label{tbl:xai_vad_cat_ucf_mf1}
\vspace*{-0.5\baselineskip}
\end{table*}

% \begin{figure*}[!t]
% \centering
% \begin{subfigure}[]{1\linewidth}
% \centering
% \includegraphics[width=1\linewidth]{images/xai_clip_ad_xd_category_clf_mf1_compare_lineplot_all_clipmodel_all_topk.jpg} 
% \end{subfigure}
% \caption{Performance of the ZSAC on the XD Violence Dataset~\cite{Wu2020not}. The model complexity of the CLIP models increases from left to right.}
% \label{fig:xai_vad_cat_xd_mf1}
% \vspace*{-0.5\baselineskip}
% \end{figure*}

% \begin{figure*}[!t]
% \centering
% \begin{subfigure}[]{1\linewidth}
% \centering
% \includegraphics[width=1\linewidth]{images/xai_clip_ad_ucf_category_clf_mf1_compare_lineplot_all_clipmodel_all_topk.jpg} 
% \end{subfigure}
% \caption{Performance of the ZSAC on the UCF-Crime Dataset~\cite{sultani2018real}. The model complexity of the CLIP models increases from left to right.}
% \label{fig:xai_vad_cat_ucf_mf1}
% \vspace*{-0.5\baselineskip}
% \end{figure*}

\subsubsection{Impact of Temporal Score Pooling}

Table~\ref{tbl:xai_vad_cat_xd_ucf_mf1_ablation} presents an ablation study on our semantic score pooling methods for adopting CLIP from images to videos using Eq.~\eqref{eq:anml_cat_pooling}, and summarizes the relative gain of the LLM-Prompt over the Keyword-Prompt. While the gain of max-pooling over average-pooling over a sequence of video frames remains limited ($\leq 3\%$), the leverage over the baseline increases substantially with LLM-assisted prompting. The mF1 increases by $33\%$ and $28\%$ before AN at $\text{Top}1$ and $\text{Top}2$, respectively, and $28\%$ and $22\%$ after AN on the XD dataset. Similarly, the gain on the UCF-Crime dataset is $-4\%$ and $1\%$ before AN, and $13\%$ and $18\%$ after AN, at $\text{Top}1$ and $\text{Top}3$, respectively. 

\begin{table*}[]
\centering
\caption{Comparison of Pooling Methods: ZSAC using the ViT-L/14 CLIP Model.}
\resizebox{0.95\textwidth}{!}{
\begin{tabular}{lccccccc}
\hline
\multirow{3}{*}{Prompt Mode} & \multirow{3}{*}{Pooling Method} & No-AN                & MASKED-AN             & No-AN                & MASKED-AN             & No-AN               & MASKED-AN              \\ \cline{3-8}
                                
 &  & \multicolumn{6}{c}{XD Violence Dataset~\cite{Wu2020not}}      \\ \cline{3-8} 
    &  & \multicolumn{2}{c}{${mF}_1$@Top$1\uparrow$}                     & \multicolumn{2}{c}{${mF}_1$@Top$2\uparrow$}                     & \multicolumn{2}{c}{${mF}_1$@Top$3\uparrow$}                     \\ \hline
         
Keyword-Prompt               & Average Pooling                 & 0.515 (--)                & 0.496 (--)                    & 0.633 (--)                & 0.616 (--)                    & 0.705 (--)                & 0.701 (--)                    \\
                             & Maximum Pooling                     & 0.516 (0\%)                & 0.498 (0\%)                    & 0.628 (-1\%)              & 0.620 (1\%)                   & 0.697 (-1\%)              & 0.710 (1\%)                   \\
{LLM-Prompt (Ours)}  & Average Pooling                 & 0.673 (31\%)              & 0.621 (25\%)                  & 0.798 (26\%)              & 0.750 (22\%)                  & 0.859 (22\%)              & 0.818 (17\%)                  \\
                             & Maximum Pooling                     & \textbf{0.683 (33\%)}     & \textbf{0.633 (28\%)}         & \textbf{0.811 (28\%)}     & \textbf{0.754 (22\%) }                 & \textbf{0.874 (24\%)}     & \textbf{0.825 (18\%)}      \\  \hline
   &  &         \multicolumn{6}{c}{UCF-Crime Dataset~\cite{sultani2018real}} \\ \cline{3-8} 
   &  & \multicolumn{2}{c}{${mF}_1$@Top$1\uparrow$}                     & \multicolumn{2}{c}{${mF}_1$@Top$3\uparrow$}                     & \multicolumn{2}{c}{${mF}_1$@Top$5\uparrow$}                     \\ \hline 

{Keyword-Prompt} & Average Pooling                 & 0.312 (--)           & 0.244 (--)            & 0.551 (--)           & 0.435 (--)            & \textbf{0.736 (--)} & 0.671 (--)             \\
                                & Maximum Pooling                     & \textbf{0.316 (1\%)} & 0.247 (1\%)           & \textbf{0.562 (2\%)} & 0.445 (2\%)           & 0.728 (-1\%)        & \textbf{0.673 (0\%)} \\
{LLM-Prompt (Ours)}     & Average Pooling                 & 0.298 (-4\%)         & 0.269 (10\%)          & 0.555 (1\%)          & 0.496 (14\%)          & 0.714 (-3\%)        & 0.661 (-1\%)           \\
                                & Maximum Pooling                     & 0.301 (-4\%)         & \textbf{0.275 (13\%)} & 0.558 (1\%)          & \textbf{0.515 (18\%)} & 0.733 (0\%)          & 0.671 (0\%)         
                                      \\ \hline
\end{tabular}
}
\noindent\footnotesize{\par  
The relative score is calculated with respect to the baseline: \textit{Keyword-Prompt} with \textit{average pooling} in Eq.~\eqref{eq:anml_cat_pooling}.
\par}
\label{tbl:xai_vad_cat_xd_ucf_mf1_ablation}
\vspace*{-0.5\baselineskip}
\end{table*}

\subsection{Evaluation on Anomaly Visual Localization}
\label{sec:zsam}

Since the ViT-L14 CLIP model with LLM prompting achieves the best ZSAC performance in most use cases and has $4 \times$ lower complexity than ViT-L/14@336px, we employ it for our ZSAM study. 
We investigate three major approaches for ZS heatmap generation: 1) Grad-ECLIP~\cite{chenyang_gradeclip}, 2) Grad-ECLIP with thresholding in Eq.~\eqref{eq:anml_heatmap_final} (Grad-ECLIP + Norm), and 3) Grad-ECLIP with PPA (PPA + Grad-ECLIP + Norm). 
The PPA approach uses a sliding window $w=[448, 448]$, twice the CLIP's base input resolution following~\cite{gu2025univad}, with step $G_w=336$, and empirically determined thresholds $\xi_\text{min}=0.05, \xi_\text{max}=0.50,~\text{and}~\bar{\xi}_\text{min}=0.30$, unless explicitly stated. 
% We have primarily employed the YOLOv8m-seg model (denoted as YOLO8-M hereafter) for AN, which offers a balance between segmentation accuracy and computation overhead. 
% Moreover, we have experimented with larger (X-versions) and recent models, such as YOLO11, YOLO26, and RF-DETR~\cite{chakrabarty2026yolo26, robinson2025rf}, to evaluate their effectiveness on relevant utility retention for ZSAM after AN. Furthermore, open-vocabulary GroundedSAM2 models~\cite{ren2024grounded} are employed to assess accuracy in challenging scenarios, such as object-of-interest occlusion due to visual overlap with a person's body.

We utilize Simuletic CCTV datasets~\cite{simuletic_weapon_2025, simuletic_knife_2025} to evaluate the heatmap localization for ZSAM efficacy. The datasets contain AI-generated CCTV images across different scenes, including weapon- and knife-related threats, with varying profiles. 
The datasets include annotated bounding boxes for the target anomalies, such as guns and knives. 
We convert the RGB heatmap of our XAI map to grayscale, apply a threshold $\bar{\xi}_\text{bb}$ to obtain bounding boxes for the maximum-area detected objects, and evaluate detection performance. 
% We use \textit{intersection over union} (IoU) between the ground-truth and predicted boxes to evaluate accuracy using popular object detection metrics---average precision (AP) and average recall (AR). 
We found that the ground-truth annotation boxes cover larger areas, such as the part of the weapon held by the hand, whereas the predicted ZSAM maps are localized to the objects. 
For a fair comparison, we have thus used a $\text{IoU}@[10\%,50\%]$ overlap percentage, ranging from $10\%$ to $50\%$ in $5\%$ increments. 
Furthermore, we evaluate the localization across different anomaly sizes. We break down the performance with size-based analysis: ${small}=[0, 32]$, ${medium}=[32, 96]$, and ${large}=[96,-]$ use cases.

We have employed two high-resolution CCTV datasets from Simuletic: 1) Simuletic CCTV Weapon dataset for gun threat detection, which contains 141 scenes with an image size of $1358 \times 924$, and 2) Simuletic CCTV Knife dataset for knife detection, which contains 114 scenes with an image size of $1358 \times 916$. 
The dataset demonstrates realistic scenes with diverse CCTV view scenarios in depth, view angle, lighting conditions, and location. 
The scenes with active threats are selected from the datasets for evaluation. 
We employ the ViT-L/14 CLIP model with LLM-prompted embedding for gun- and knife-holding for the ZSAM.

Table~\ref{tbl:xai_vad_heatmap_simuletic_weapon} provides the ZSAM performance on the Simuletic CCTV Weapon dataset, where our auto-focus approach significantly outperforms the baseline Grad-ECLIP method. The $\text{AP}$ increases from $0.307$ to $0.569$, and the $\text{AR}$ from $0.473$ to $0.715$. The $\text{AP}$ of the PPA increases from $0.18$ to above $0.90$ when the IoU reduces from $50\%$ to $10\%$, indicating that the ZSAM captures the anomaly region with high accuracy, but its bounding box IoU can remain smaller. This is not unexpected for a box-based evaluation, as a small mismatch in the boxes can degrade the IoU substantially. 
% Moreover, the Simuletic datasets involve annotations with extended boxes beyond the target, including holding hands, and sometimes across occlusions, whereas the ZSAM boxes are localized in high-saliency regions (as shown in Fig.~\ref{fig:xai_vad_heatmap_simuletic_weapon_knife_map_bboxes_mismatch} in Appendix~\ref{sec:zsam_appendix}). 
In the size-based analysis, the Grad-ECLIP struggles with small objects with $\text{AP}_\text{small}=0.0$, whereas our PPA demonstrates robustness to varying sizes, i.e., $\text{AP}_\text{small}=0.710$ while simultaneously maintaining higher accuracy for the medium and large sizes.

Our PPA approach achieves a more substantial improvement on the Simuletic CCTV Knife dataset (as presented in Table~\ref{tbl:xai_vad_heatmap_simuletic_knife}).  
The $\text{AP}$ increases from $0.098$ to $0.589$, leveraged with our PPA. 
When the IoU reduces from $50\%$ to $10\%$, the $\text{AP}$ increases from $0.088$ to $0.952$. 
The higher $\text{IoU}@50\%$ score is because the knife dataset mainly contains medium- and large-sized objects, which are detected with high accuracy at $0.579$ and $0.684$, respectively (see Table~\ref{tbl:xai_vad_heatmap_simuletic_knife}).

The MASKED-AN degrades the AP by $43\%$ and $36\%$ for the weapon and knife datasets, respectively. The main drop emanates from the higher IoU, i.e., at $\text{IoU}@50\%$ with $51\%$ and $75\%$, whereas the $\text{IoU}@10\%$ stays around $27\%$ and $23\%$, respectively. 
Other possible reasons are:
\begin{itemize}
    \item \textbf{Poor segmentation for overlapped objects:} Unknown objects, including weapons, are segmented poorly using pre-trained YOLO when they visually overlap the person's body (as shown in Fig.~\ref{fig:xai_vad_heatmap_simuletic_weapon_knife_occluded_map} in Appendix~\ref{sec:zsam_appendix}).
    \item \textbf{Imprecise segmentation mask boundaries:} As person figures get smaller in the far view of CCTV, the mask pixels become less precise in isolating hand-held objects. 
    \item \textbf{Impact of mask color:} Since most weapons, such as guns and rifles, have a predominantly darker color, the black AN mask can impact their ZSAM saliency.  
\end{itemize}

To address the above challenges, we conduct further experiments using more robust and heavier segmentation models and configuration settings (as discussed in Section~\ref{sec:zsam_seg_improve_overlap}).

% weapon-gun
\begin{table*}[]
\centering
\caption{Performance of ZSAM on the Simuletic CCTV Weapon Dataset~\cite{simuletic_weapon_2025}. The MASKED-AN uses the YOLO8-M Model.}
\resizebox{1\textwidth}{!}{
\begin{tabular}{lccccccccccccc}
\hline
XAI MAP Method                                                     &               & $AP$$\uparrow$              & $AP[10\%]$$\uparrow$           & $AP[50\%]$$\uparrow$           & $AR$$\uparrow$              & $AR[10\%]$$\uparrow$           & $AR[50\%]$$\uparrow$           & $AP_{\text{small}}$$\uparrow$       & $AP_{\text{medium}}$$\uparrow$      & $AP_{\text{large}}$$\uparrow$       & $AR_{\text{small}}$$\uparrow$       & $AR_{\text{medium}}$$\uparrow$      & $AR_{\text{large}}$$\uparrow$       \\ \hline
% Grad-ECLIP                                               & No-AN         & 0.307          & 0.669          & 0.090          & 0.473          & 0.785          & 0.215          & 0.000          & 0.305          & 0.383          & 0.000          & 0.374          & 0.688          \\
%  Grad-ECLIP + Norm (Ours)                                      & No-AN         & 0.311          & 0.600          & 0.077          & 0.479          & 0.723          & 0.223          & 0.000          & 0.308          & 0.408          & 0.000          & 0.402          & 0.662          \\
% {PPA + Grad-ECLIP + Norm (Ours)} & No-AN         & \textbf{0.569} & \textbf{0.901} & \textbf{0.180} & \textbf{0.715} & \textbf{0.938} & \textbf{0.392} & \textbf{0.710} & \textbf{0.541} & \textbf{0.655} & \textbf{0.759} & \textbf{0.712} & \textbf{0.715} \\
% {PPA + Grad-ECLIP + Norm (Ours)}                                                                   & MASKED-AN     & 0.318          & 0.656          & 0.088          & 0.460          & 0.754          & 0.208          & 0.159          & 0.374          & 0.277          & 0.352          & 0.528          & 0.366          \\
% {PPA + Grad-ECLIP + Norm (Ours)}                                                                   & $\Delta (\%)$ & -25\%          & -24\%          & -9\%           & -25\%          & -18\%          & -18\%          & -55\%          & -17\%          & -38\%          & -41\%          & -18\%          & -35\%       \\ \hline   
Grad-ECLIP~\cite{chenyang_gradeclip}                             & No-AN         & 0.307 & 0.669 & 0.090 & 0.473 & 0.785 & 0.215 & 0.000     & 0.305      & 0.383     & 0.000     & 0.374      & 0.688     \\
Grad-ECLIP + Norm (Ours)                     & No-AN         & 0.311 & 0.600 & 0.077 & 0.479 & 0.723 & 0.223 & 0.000     & 0.308      & 0.408     & 0.000     & 0.402      & 0.662     \\
PPA + Grad-ECLIP + Norm (Ours) & No-AN         & \textbf{0.569} & \textbf{0.901} & \textbf{0.180} & \textbf{0.715} & \textbf{0.938} & \textbf{0.392} & \textbf{0.710}     & \textbf{0.541}      & \textbf{0.655}     & \textbf{0.759}     & \textbf{0.712}      & \textbf{0.715}     \\
PPA + Grad-ECLIP + Norm (Ours)              & MASKED-AN     & 0.318 & 0.656 & 0.088 & 0.460 & 0.754 & 0.208 & 0.159     & 0.374      & 0.277     & 0.352     & 0.528      & 0.366     \\
                                                  & $\Delta (\%)$ & -44\% & -27\% & -51\% & -36\% & -20\% & -47\% & -78\%     & -31\%      & -58\%     & -54\%     & -26\%      & -49\%      \\ \hline   

\end{tabular}
}
\noindent\footnotesize{\par  
The $AP:AP_{\text{IoU}@[10\%,50\%]}, AP[x\%]:AP_{\text{IoU}@[x\%]}, AR:AR_{\text{IoU}@[10\%,50\%]},~\text{and}~AR[x\%]:AR_{\text{IoU}@[x\%]}$. 
% The $\Delta (\%)$ is relative performance after applying AN and calculated as $\Delta (\%) = \frac{\text{score[MASKED-AN]} - \text{score[No-AN]}}{\text{score[No-AN]}}$. 
Thresholds: $\xi_\text{min}=0.05, \xi_\text{max}=0.50, \bar{\xi}_\text{min}=0.30,~\text{and}~\bar{\xi}_\text{bb}=0.5$.
\par}
\label{tbl:xai_vad_heatmap_simuletic_weapon}
\vspace*{-0.5\baselineskip}
\end{table*}

% knife
\begin{table*}[]
\centering
\caption{Performance of ZSAM on the Simuletic CCTV Knife Dataset~\cite{simuletic_knife_2025}. The MASKED-AN uses the YOLO8-M Model.}
\resizebox{1\textwidth}{!}{
\begin{tabular}{lccccccccccccc}
\hline
XAI MAP Method                                                     &               & $AP$$\uparrow$              & $AP[10\%]$$\uparrow$           & $AP[50\%]$$\uparrow$           & $AR$$\uparrow$              & $AR[10\%]$$\uparrow$           & $AR[50\%]$$\uparrow$           & $AP_{\text{small}}$$\uparrow$       & $AP_{\text{medium}}$$\uparrow$      & $AP_{\text{large}}$$\uparrow$       & $AR_{\text{small}}$$\uparrow$       & $AR_{\text{medium}}$$\uparrow$      & $AR_{\text{large}}$$\uparrow$       \\ \hline
Grad-ECLIP~\cite{chenyang_gradeclip}                             & No-AN         & 0.098          & 0.245          & 0.015          & 0.185          & 0.354          & 0.061          & --        & 0.113          & 0.142          & --        & 0.145          & 0.278          \\
Grad-ECLIP + Norm (Ours)                     & No-AN         & 0.136          & 0.299          & 0.021          & 0.208          & 0.404          & 0.071          & --        & 0.167          & 0.147          & --        & 0.224          & 0.170          \\
PPA + Grad-ECLIP + Norm (Ours) & No-AN         & \textbf{0.589} & \textbf{0.952} & \textbf{0.088} & \textbf{0.697} & \textbf{0.970} & \textbf{0.253} & --        & \textbf{0.579} & \textbf{0.684} & --        & \textbf{0.702} & \textbf{0.685} \\
PPA + Grad-ECLIP + Norm (Ours)              & MASKED-AN     & 0.375          & 0.733          & 0.022          & 0.476          & 0.758          & 0.111          & --        & 0.349          & 0.505          & --        & 0.443          & 0.552          \\
                                                  & $\Delta (\%)$ & -36\%          & -23\%          & -75\%          & -32\%          & -22\%          & -56\%          & --        & -40\%          & -26\%          & --        & -37\%          & -19\%    \\ \hline   
\end{tabular}
}
\noindent\footnotesize{\par  
% The $\Delta (\%)$ is relative performance after applying AN and calculated as $\Delta (\%) = \frac{\text{score[MASKED-AN]} - \text{score[No-AN]}}{\text{score[No-AN]}}$. 
The $AP:AP_{\text{IoU}@[10\%,50\%]}, AP[x\%]:AP_{\text{IoU}@[x\%]}, AR:AR_{\text{IoU}@[10\%,50\%]},~\text{and}~AR[x\%]:AR_{\text{IoU}@[x\%]}$. Thresholds:  $\xi_\text{min}=0.05, \xi_\text{max}=0.50, \bar{\xi}_\text{min}=0.30,~\text{and}~\bar{\xi}_\text{bb}=0.5$.
\par}
\label{tbl:xai_vad_heatmap_simuletic_knife}
\vspace*{-0.5\baselineskip}
\end{table*}

\subsubsection{Robustness of Masking Models on Overlapped Objects}
\label{sec:zsam_seg_improve_overlap}

The ZSAM results have demonstrated that YOLO8-M AN masking has a substantial impact on performance due to imprecise and imperfect segmentation regions. Thus, we experiment with large segmentation models ($M \rightarrow X$), and recent improved models from YOLO11, YOLO26, and RF-DETR~\cite{chakrabarty2026yolo26, robinson2025rf}. The closed-vocabulary models of YOLO and RF-DETR are trained on MS-COCO datasets and can mask weapon objects during AN. We employ our GDSN, based on open-vocabulary GroundedSAM2~\cite{ren2024grounded}, to retain objects of interest when overlapped by the person's body.

Using more robust YOLO models on the weapon (gun) datasets reduces the MASKED-AN trade-off from $44\%$ to $35\%\text{--}42\%$ in $\text{AP}$ and $36\%$ to $26\%\text{--}33\%$ in $\text{AR}$ (see Table~\ref{tbl:xai_vad_heatmap_simuletic_weapon_multi_mask_methods}). The RF-DETR models perform with leaner segmentation, with an AP drop rate of $32\%\text{--}33\%$. 
Nevertheless, both YOLO and RF-DETR are still unable to segment overlapped objects effectively. 
Our GDSN-L (L denotes large of SAM2.1-Large~\cite{ravi2025sam}) improves ZSAM by limiting AN impact to $24\%$ and $19\%$ for the AP and AR, respectively. The leverage of our GDSN emanates from its robustness in segmenting weapons when overlapping with the person's body, achieved through the negative segmentation discussed in Section~\ref{sec:seg_method} (as shown in  Fig.~\ref{fig:xai_vad_heatmap_simuletic_weapon_knife_occluded_map} in Appendix~\ref{sec:zsam_appendix}). 

The YOLO and RF-DETR models reduce the AP drop rate from the baseline of $36\%$ to $30\%$ and $22\%$, respectively, on the knife dataset (see Table~\ref{tbl:xai_vad_heatmap_simuletic_knife_multi_mask_methods}). 
The MASKED-AN with our GDSN maintains the ZSAM accuracy robustly, with an AP drop rate of $1\%$, demonstrating the importance of solving overlap during the segmentation mask generation. 
% \textupdate{The approach even improves the accuracy at $\text{IoU}@50\%$ by $36\%$, which can be attributed to the denoising effect of masking aiding the ZSAM to localize better. }
The enhanced score for the knife over the gun can be attributed to variations in object size, overlap possibility, and object color.
Guns can cover a large area and are susceptible to occlusion, which impacts higher IoU, as demonstrated by the comparable $\text{IoU}@10\%$ trade-off with $6\%$ and $13\%$ compared to $\text{IoU}@50\%$. 
The possibility of overlap on the person's body—guns can be held closer to the body while walking or running, whereas knives are often held to the side with a lower chance of overlap on the person's body.
For object color attributes, guns often come in black, while knives have a shiny silver color, and the difference in performance is not unexpected, as CLIP is known for its good color perception~\cite{chenyang_gradeclip}.
In conclusion, the ZSAM's performance on MASKED-AN using GDSN is promising. Particularly, the lower trade-offs at $\text{IoU}@10\%$ indicate ZSAM's capability in localizing the anomaly region, although the heatmap might miss part of the region. 
\begin{table*}[]
\centering
\caption{Performance of ZSAM with PPA + Grad-ECLIP + Norm (Ours) with Different Segmentation Methods on the Simuletic CCTV Weapon Dataset~\cite{simuletic_weapon_2025}. The YOLO models employ SAHI~\cite{akyon2022slicing} with a window size of $[640 \times 640]$ to handle high-resolution images.}
\resizebox{1\textwidth}{!}{
\begin{tabular}{lcccccccccccc}
\hline
AN Method (Seg. Method)    & $AP$$\uparrow$      & $\Delta(\%)$$\uparrow$    & $AP[10\%]$$\uparrow$     & $\Delta(\%)$$\uparrow$  & $AP[50\%]$$\uparrow$   & $\Delta(\%)$$\uparrow$ & $AR$$\uparrow$    & $\Delta(\%)$$\uparrow$    & $AR[10\%]$$\uparrow$   & $\Delta(\%)$$\uparrow$  & $AR[50\%]$$\uparrow$   & $\Delta(\%)$$\uparrow$  \\ \hline
No-AN (--)                       & 0.569          & --             & 0.901          & --             & 0.180          & --             & 0.715          & --             & 0.938          & --            & 0.392          & --             \\
MASKED-AN (YOLO8-M)       & 0.318          & -44\%          & 0.656          & -27\%          & 0.088          & -51\%          & 0.460          & -36\%          & 0.754          & -20\%         & 0.208          & -47\%          \\
MASKED-AN (YOLO8-X)       & 0.328          & -42\%          & 0.639          & -29\%          & 0.095          & -47\%          & 0.478          & -33\%          & 0.746          & -20\%         & 0.223          & -43\%          \\
MASKED-AN (YOLO11-M)       & 0.340          & -40\%          & 0.624          & -31\%          & 0.126          & -30\%          & 0.491          & -31\%          & 0.731          & -22\%         & 0.262          & -33\%          \\
MASKED-AN (YOLO11-X)       & 0.370          & -35\%          & 0.647          & -28\%          & 0.105          & -42\%          & 0.529          & -26\%          & 0.769          & -18\%         & 0.246          & -37\%          \\
MASKED-AN (YOLO26-M)       & 0.366          & -36\%          & 0.653          & -28\%          & 0.087          & -52\%          & 0.516          & -28\%          & 0.769          & -18\%         & 0.238          & -39\%          \\
MASKED-AN (YOLO26-X)       & 0.330          & -42\%          & 0.634          & -30\%          & 0.076          & -58\%          & 0.506          & -29\%          & 0.762          & -19\%         & 0.238          & -39\%          \\
MASKED-AN (RF-DETR-M)   & 0.389          & -32\%          & 0.733          & -19\%          & 0.120          & -33\%          & 0.518          & -28\%          & 0.808          & -14\%         & 0.231          & -41\%          \\
MASKED-AN (RF-DETR-M)    & 0.384          & -33\%          & 0.704          & -22\%          & 0.125          & -31\%          & 0.520          & -27\%          & 0.777          & -17\%         & 0.254          & -35\%          \\
MASKED-AN (GDSN-L (Ours)) & \textbf{0.430} & \textbf{-24\%} & \textbf{0.786} & \textbf{-13\%} & \textbf{0.130} & \textbf{-28\%} & \textbf{0.580} & \textbf{-19\%} & \textbf{0.862} & \textbf{-8\%} & \textbf{0.292} & \textbf{-26\%} \\ \hline 
\end{tabular}
}
\noindent\footnotesize{\par  
% The $\Delta (\%)$ is relative performance after applying AN and calculated as $\Delta (\%) = \frac{\text{score[MASKED-AN]} - \text{score[No-AN]}}{\text{score[No-AN]}}$.  
The $AP:AP_{\text{IoU}@[10\%,50\%]}, AP[x\%]:AP_{\text{IoU}@[x\%]}, AR:AR_{\text{IoU}@[10\%,50\%]},~\text{and}~AR[x\%]:AR_{\text{IoU}@[x\%]}$. 
\newline Thresholds:  $\xi_\text{min}=0.05, \xi_\text{max}=0.50, \bar{\xi}_\text{min}=0.30,~\text{and}~\bar{\xi}_\text{bb}=0.5$.
\par}
\label{tbl:xai_vad_heatmap_simuletic_weapon_multi_mask_methods}
\vspace*{-0.5\baselineskip}
\end{table*}

% knife-multi-mask-models
\begin{table*}[]
\centering
\caption{Performance of ZSAM with PPA + Grad-ECLIP + Norm (Ours) with Different Segmentation Methods on the Simuletic CCTV Knife Dataset~\cite{simuletic_weapon_2025}. The YOLO models employ SAHI~\cite{akyon2022slicing} with a window size of $[640 \times 640]$ to handle high-resolution images.}
\resizebox{1\textwidth}{!}{
\begin{tabular}{lcccccccccccc}
\hline
AN Method (Seg. Method)    & $AP$$\uparrow$      & $\Delta(\%)$$\uparrow$    & $AP[10\%]$$\uparrow$     & $\Delta(\%)$$\uparrow$  & $AP[50\%]$$\uparrow$   & $\Delta(\%)$$\uparrow$ & $AR$$\uparrow$     & $\Delta(\%)$$\uparrow$    & $AR[10\%]$$\uparrow$   & $\Delta(\%)$$\uparrow$  & $AR[50\%]$$\uparrow$   & $\Delta(\%)$$\uparrow$  \\ \hline
No-AN (--)                         & 0.589          & --            & 0.952          & --            & 0.088          & --            & 0.697          & --            & 0.970          & --            & 0.253          & --            \\
MASKED-AN (YOLO8-M)       & 0.375          & -36\%         & 0.733          & -23\%         & 0.022          & -75\%         & 0.476          & -32\%         & 0.758          & -22\%         & 0.111          & -56\%         \\
MASKED-AN (YOLO8-X)       & 0.367          & -38\%         & 0.705          & -26\%         & 0.017          & -81\%         & 0.473          & -32\%         & 0.747          & -23\%         & 0.111          & -56\%         \\
MASKED-AN (YOLO11-M)       & 0.358          & -39\%         & 0.671          & -30\%         & 0.024          & -73\%         & 0.468          & -33\%         & 0.727          & -25\%         & 0.131          & -48\%         \\
MASKED-AN (YOLO11-X)       & 0.410          & -30\%         & 0.786          & -17\%         & 0.018          & -80\%         & 0.523          & -25\%         & 0.818          & -16\%         & 0.101          & -60\%         \\
MASKED-AN (YOLO26-M)       & 0.375          & -36\%         & 0.742          & -22\%         & 0.023          & -74\%         & 0.496          & -29\%         & 0.788          & -19\%         & 0.121          & -52\%         \\
MASKED-AN (YOLO26-X)       & 0.358          & -39\%         & 0.730          & -23\%         & 0.030          & -66\%         & 0.481          & -31\%         & 0.778          & -20\%         & 0.141          & -44\%         \\
MASKED-AN (RF-DETR-M)   & 0.458          & -22\%         & 0.800          & -16\%         & 0.080          & -9\%          & 0.575          & -18\%         & 0.828          & -15\%         & 0.222          & -12\%         \\
MASKED-AN (RF-DETR-M)    & 0.430          & -27\%         & 0.769          & -19\%         & 0.065          & -26\%         & 0.532          & -24\%         & 0.798          & -18\%         & 0.182          & -28\%         \\
MASKED-AN (GDSN-L (Ours)) & \textbf{0.582} & \textbf{-1\%} & \textbf{0.891} & \textbf{-6\%} & \textbf{0.120} & \textbf{36\%} & \textbf{0.677} & \textbf{-3\%} & \textbf{0.909} & \textbf{-6\%} & \textbf{0.273} & \textbf{8\%} 
 \\ \hline 
\end{tabular}
}
\noindent\footnotesize{\par  
% The $\Delta (\%)$ is relative performance after applying AN and calculated as $\Delta (\%) = \frac{\text{score[MASKED-AN]} - \text{score[No-AN]}}{\text{score[No-AN]}}$. 
% The $AP:AP_{\text{IoU}@[10\%,50\%]}, AP[x\%]:AP_{\text{IoU}@[x\%]}, AR:AR_{\text{IoU}@[10\%,50\%]},~\text{and}~AR[x\%]:AR_{\text{IoU}@[x\%]}$. Thresholds:  $\xi_\text{min}=0.05, \xi_\text{max}=0.50, \bar{\xi}_\text{min}=0.30,~\text{and}~\bar{\xi}_\text{bb}=0.5$.
\par}
\label{tbl:xai_vad_heatmap_simuletic_knife_multi_mask_methods}
\vspace*{-0.5\baselineskip}
\end{table*}

% \begin{figure*}[!t]
% \centering
% \begin{subfigure}[]{0.75\linewidth}
% \centering
% \includegraphics[width=1\linewidth]{images/simuletic_data/xai_clip_ad_gunholding_map_ap_ar.jpg} 
% \label{fig:xai_vad_heatmap_simuletic_weapon_multi_mask_methods_lineplot}
% \caption{}
% \end{subfigure}
% \begin{subfigure}[]{0.75\linewidth}
% \centering
% \includegraphics[width=1\linewidth]{images/simuletic_data/xai_clip_ad_knifeholding_map_ap_ar.jpg} 
% \label{fig:xai_vad_heatmap_simuletic_knife_multi_mask_methodss_lineplot}
% \caption{}
% \end{subfigure}
% \caption{Performance of the ZSAM using different heatmap-to-bounding-box thresholds $\bar{\xi}_\text{bb}$ on Simuletic CCTV a) Weapon~\cite{simuletic_weapon_2025} and b) Knife datasets~\cite{simuletic_knife_2025}. 
% The plots demonstrate the superiority of our Grad-ECLIP with PPA (PPA + Grad-ECLIP + Norm) and GDSN approaches. 
% Optimal performance at $\bar{\xi}_\text{bb}\leq 0.5$.
% }
% \label{fig:xai_vad_heatmap_simuletic_weapon_knife_multi_mask_methodss_lineplot}
% \vspace*{-0.5\baselineskip}
% \end{figure*}

% \begin{figure*}[!t]
% \centering
% \begin{subfigure}[]{1\linewidth}
% \centering
% \includegraphics[width=1\linewidth]{images/xai_clip_ad_ucf_category_clf_mf1_compare_lineplot_all_clipmodel_all_topk.jpg} 
% \end{subfigure}
% \caption{Performance of the ZSAC on the UCF-Crime Dataset~\cite{sultani2018real}. The model complexity of the CLIP models increases from left to right.}
% \label{fig:xai_vad_cat_ucf_mf1}
% \vspace*{-0.5\baselineskip}
% \end{figure*}

\subsection{Evaluation on Anomaly Context Generation}

{Zhang et. al.~\cite{zhang2025holmes} have released the HIVAU-70k dataset that contains curated anomaly descriptions for the XD Violence and UCF-Crime VAD datasets. The test set includes 3,300 samples at clip, event, and video granularities.
The clips are random-length samples from each event. 
The events correspond to temporally continuous anomaly states in the videos; non-continuous
anomalous states are considered separate events. Our study employs the event granularity to focus the caption on single non-interrupted anomaly scenes.
}

{Table~\ref{tbl:xai_vad_vlm_xd_ucf} presents the accuracy of the ZSAD using lightweight VLM models on the HIVAU-70k dataset.  
The ZSAD system generates a short, concise characterization of the anomaly, while the ground-truth annotations consist of longer captions of the videos. 
We found that BLEU and ROUGE scores are less effective since they penalize short responses due to lower recall. 
Text summarization metrics, such as BERTScore, are relatively robust for evaluating the ZSAD, as they measure accuracy using semantic similarity (capturing synonyms but still affected by text length) between the ground truth and the predicted anomaly captions.
The InternVL3.5 models achieve the best score with BERTScore $F_1$ reaching $0.422$, while the SmoVLM2 is the least performing.
The performance remains promising after MASKED-AN, with a drop of $5.70\%$--$5.97\%$, and $2.41\%$--$6.71\%$, for the XD and UCF-Crimes videos, respectively, for the InternVL3.5 models.
Furthermore, the ground truth annotations contain specific personal attributes, such as ``man'' and ``woman'', and colors of clothes and vehicles, which have also lowered the score for MASKED-AN, as these visual attributes are removed from the video by the AN. 
Overall, our ZSAD has improved performance, even with smaller models, compared to the benchmark ZS VLM scores in Ref.~\cite{zhang2025holmes}, which demonstrates the merit of curated VAD-XAI prompting.
}

\begin{table*}[]
\centering
\caption{{Performance of ZSAD on the HIVAU-70k-Event Dataset~\cite{zhang2025holmes}. BLEU refers to the cumulative values from BLEU-(1...4), and ROUGE is the cumulative of ROUGE-(1,2, and L). The BERTScore utilizes RoBERTa-base embedder, and is rescaled as in Ref.~\cite{zhang2019bertscore}.}}
\resizebox{1\textwidth}{!}{
\begin{tabular}{lccccccccccccccc}
\hline
\multirow{3}{*}{VLM Model (Params)} & \multicolumn{3}{c}{BLUE$\uparrow$}                         & \multicolumn{3}{c}{ROUGE\_L$\uparrow$}                     & \multicolumn{3}{c}{ROUGE$\uparrow$}                        & \multicolumn{3}{c}{BERTScore ($P$)$\uparrow$}                  & \multicolumn{3}{c}{BERTScore (${F}_1$)$\uparrow$}                 \\ \cline{2-16}
                                    & No-AN          & MASKED-AN      & $\Delta (\%)$  & No-AN          & MASKED-AN      & $\Delta (\%)$  & No-AN          & MASKED-AN      & $\Delta (\%)$  & No-AN          & MASKED-AN      & $\Delta (\%)$  & No-AN          & MASKED-AN      & $\Delta (\%)$  \\ \cline{2-16}
                                    & \multicolumn{15}{c}{HIVAU-70k [XD Violence + UCF-Crime] Dataset}                          \\ \hline
Video-ChatGPT (7B)~\cite{maaz2024video}*                 & 0.068          & --             & --             & 0.048          & --             & --             & --             & --             & --             & --             & --             & --             & --             & --             & --             \\
Video-LLaMA (7B)~\cite{zhang2023video}*                     & 0.079          & --             & --             & 0.067          & --             & --             & --             & --             & --             & --             & --             & --             & --             & --             & --             \\
Video-LLaVA (7B)~\cite{lin2024video}*                     & 0.046          & --             & --             & 0.023          & --             & --             & --             & --             & --             & --             & --             & --             & --             & --             & --             \\
LLaVA-Next-Video (7B)~\cite{zhang2024video}*                & 0.091          & --             & --             & 0.080          & --             & --             & --             & --             & --             & --             & --             & --             & --             & --             & --             \\
QwenVL2 (7B)~\cite{wang2024qwen2}*                         & 0.082          & --             & --             & 0.081          & --             & --             & --             & --             & --             & --             & --             & --             & --             & --             & --             \\
InternVL2 (8B)~\cite{chen2024far}*                       & 0.101          & --             & --             & 0.102          & --             & --             & --             & --             & --             & --             & --             & --             & --             & --             & --             \\ \hline 
                                    & \multicolumn{15}{c}{HIVAU-70k [XD Violence] Dataset}                                        \\ \hline 
SmolVLM2 (2.2B)~\cite{marafioti2025smolvlm}+              & 0.185          & 0.145 & -22.03\%         & 0.157          & 0.145          & -7.19\%          & 0.340          & 0.308          & -9.29\%          & 0.385          & 0.349          & -9.26\%          & 0.306          & 0.277          & -9.30\%          \\
Gemma3 (4B)~\cite{sellergren2025medgemma}+                        & 0.136          & 0.119          & -12.68\%         & 0.201          & 0.180          & -10.45\%         & 0.301          & 0.290          & \textbf{-3.68\%} & 0.440          & 0.420          & \textbf{-4.63\%} & 0.346          & 0.331          & \textbf{-4.33\%} \\
InternVL3.5 (4B)~\cite{wang2025internvl3}+                  & 0.226          & 0.189          & -16.23\%         & \textbf{0.210} & \textbf{0.193} & -8.27\%          & 0.432          & 0.378          & -12.35\%         & 0.513          & \textbf{0.488} & -4.80\%          & 0.410          & 0.387          & -5.70\%          \\
InternVL3.5 (8B)~\cite{wang2025internvl3}+                   & \textbf{0.258} & \textbf{0.232} & \textbf{-9.98\%} & 0.142          & 0.137          & \textbf{-3.26\%} & \textbf{0.460} & \textbf{0.411} & -10.60\%         & \textbf{0.522} & 0.486          & -6.85\%          & \textbf{0.422} & \textbf{0.397} & -5.97\%          \\ \hline 
                                    & \multicolumn{15}{c}{HIVAU-70k [UCF-Crime] Dataset}                              \\ \hline 
SmolVLM2 (2.2B)~\cite{marafioti2025smolvlm}+                    & 0.182          & 0.201          & 10.05\%          & 0.158          & 0.159          & 0.49\%           & 0.322          & 0.337          & \textbf{4.58\%}  & 0.406          & 0.404          & \textbf{-0.59\%} & 0.312          & 0.308          & \textbf{-1.48\%} \\
Gemma3 (4B)~\cite{sellergren2025medgemma}+                        & 0.135          & 0.152          & \textbf{12.07\%} & 0.163          & 0.152          & -6.59\%          & 0.268          & 0.271          & 1.02\%           & 0.412          & 0.404          & -1.86\%          & 0.327          & 0.322          & -1.64\%          \\
InternVL3.5 (4B)~\cite{wang2025internvl3}+                   & 0.183          & 0.164          & -10.17\%         & \textbf{0.179} & \textbf{0.174} & -3.30 \%         & 0.331          & 0.301          & -9.01\%          & 0.464          & 0.431          & -7.01\%          & 0.370          & 0.345          & -6.71\%          \\
InternVL3.5 (8B)~\cite{wang2025internvl3}+                   & \textbf{0.245} & \textbf{0.256} & 4.49\%           & 0.136          & 0.138          & \textbf{1.34\%}  & \textbf{0.367} & \textbf{0.376} & 2.39\%           & \textbf{0.473} & \textbf{0.444} & -6.09\%          & \textbf{0.377} & \textbf{0.368} & -2.41\%       
 \\ \hline 
\end{tabular}
}
\noindent\footnotesize{\par We use $N_\text{vlm}=16, {temperature}=0.2, {top\_p}=0.90, ~\text{and}~{top\_k}=10$. 
The ``+'' denotes using our structured VLM prompting. The ``*'' denotes benchmark ZS results reported in Ref.~\cite{zhang2025holmes}. 
\par}
\label{tbl:xai_vad_vlm_xd_ucf}
\vspace*{-0.5\baselineskip}
\end{table*}

\subsection{\textupdate{Computational Cost Analysis}}
\label{sec:cc_appendix}

\textupdate{We have analyzed the processing GPU time and memory cost of several methods employed in our ZS XAVAD pipeline (see Table~\ref{tbl:xai_cc}). We measured the cost on a system with 32 AMD EPYC Milan processors, 64~GB of RAM, and a 40~GB Nvidia A100D-40C.
Our PPA achieves $4\times$ memory efficiency and $30\%$ faster processing time while improving ZSAM AP accuracy by $4\text{--}6 \times$ compared to Grad-ECLIP. In contrast, simple patching with Grad-ECLIP increases the cost by $1.8 \times$ and $6.8 \times$, respectively.
The speed varies slightly for the ZSAC CLIP models, except for the ViT-L/14@336px, which is $2\text{--}3 \times$ slower.
YOLO is the most memory-efficient, and RF-DETR is the fastest among the AN mask generation models. 
Despite substantial accuracy gains of the GDSN in ZSAM, it has a relatively higher memory cost (due to large SAM models), while the difference in speed compared to YOLO-M and -X remains marginal.
The computational complexity of VLM models depends on several factors, such as model size, active parameters, context window size, and KV cache optimization. 
The InternVL3.5 models maintain fast inference and promising ZSAD accuracy even with the larger model sizes.
}

\begin{table}[]
\centering
\caption{\textupdate{Computational Cost Analysis on AN and XAI Inference. We employed parallel multi-core processing for Multi-Frame and Patch Inference. The N, S, M, L, and X denote tiny, small, medium, large, and extra-large models. }}
\resizebox{1\columnwidth}{!}{
\begin{tabular}{lcc}
\hline
    Methods                                   & Peak GPU Memory (MB)$\downarrow$                                        & GPU Time (s)$\downarrow$                                                       \\ \hline
% AN Method (Seg. Model)                & \multicolumn{2}{c}{Simuletic CCTV Datasets, Avg. Image Size=$[920 \times 1358]$}                                \\ \hline
\multirow{2}{*}{AN Method (Seg. Model)} & \multicolumn{2}{c}{Simuletic CCTV Datasets} \\
 & \multicolumn{2}{c}{Avg. Image Size=$[920 \times 1358]$} \\\hline
YOLO8-M*                           & \textbf{195}                                              & 0.095                                                              \\
YOLO8-M                            & 231                                                       & 0.310                                                              \\
YOLO8-X                            & 471                                                       & 0.315                                                              \\
YOLO11-M                            & 244                                                       & 0.335                                                              \\
YOLO11-X                            & 455                                                       & 0.370                                                              \\
YOLO26-M                            & 263                                                       & 0.350                                                              \\
YOLO26-X                            & 479                                                       & 0.370                                                              \\
RF-DETR-M                        & 1498                                                      & 0.110                                                              \\
RF-DETR-L                         & 1504                                                      & \textbf{0.080}                                                     \\
GDSN-N (Ours)                     & 2000                                                      & 0.290                                                              \\
GDSN-S (Ours)                    & 2028                                                      & 0.305                                                              \\
GDSN-L (Ours)                    & 2708                                                      & 0.395                                                              \\\hline
% ZSAC: CLIP Model (Pool)         & \multicolumn{2}{c}{XD and UCF-Crime Datasets, Avg. Frame Size=$[320 \times 240]$, $T=16$}                   \\\hline
\multirow{2}{*}{ZSAC: CLIP Model (Pooling Method)} & \multicolumn{2}{c}{XD Violence and UCF-Crime Datasets} \\
 & \multicolumn{2}{c}{Avg. Frame Size=$[320 \times 240]$, $T=16$} \\\hline
ViT-B/32 (Average)                     & \textbf{384}                                              & \textbf{0.060}                                                     \\
ViT-B/32 (Maximum) (Ours)                        & \textbf{384}                                              & \textbf{0.060}                                                     \\
ViT-B/16 (Average)                     & 437                                                       & 0.060                                                              \\
ViT-B/16 (Maximum) (Ours)                        & 437                                                       & 0.065                                                              \\
ViT-L/14 (Average)                     & 1050                                                      & 0.070                                                              \\
ViT-L/14 (Maximum) (Ours)                        & 1050                                                      & 0.065                                                              \\
ViT-L/14@336px (Average)               & 1227                                                      & 0.200                                                              \\
ViT-L/14@336px (Maximum) (Ours)         & 1227                                                      & 0.130                                                              \\\hline
% ZSAM: MAP Method                       & \multicolumn{2}{c}{Simuletic CCTV Datasets, Avg. Image Size=$[920 \times 1358]$}                                \\\hline
\multirow{2}{*}{ZSAM: MAP Method} & \multicolumn{2}{c}{Simuletic CCTV Datasets} \\
 & \multicolumn{2}{c}{Avg. Image Size=$[920 \times 1358]$} \\\hline
Grad-ECLIP                             & 7989                                                      & 0.180                                      \\
Grad-ECLIP + Norm (Ours)                     & 7989                                                      & 0.195                                                              \\
Patching + Grad-ECLIP (Ours)                  & 14241                                                     & 1.240                                                              \\
Patching + Grad-ECLIP + Norm (Ours)           & 14241                                                     & 1.245                                                              \\
PPA + Grad-ECLIP (Ours)                      & \textbf{2026}                                             & \textbf{0.120}                                                     \\
PPA + Grad-ECLIP + Norm (Ours)                & \textbf{2026}                                             & {0.125}                                                     \\\hline
% {ZSAD: VLM Model} & \multicolumn{2}{c}{HIVAU-70k Dataset, Avg. Frame Size={$[320 \times 240]$}, $N_\text{vlm}=16$} \\\hline
\multirow{2}{*}{ZSAD: VLM Model} & \multicolumn{2}{c}{HIVAU-70k Dataset} \\
 & \multicolumn{2}{c}{Avg. Frame Size={$[320 \times 240]$}, $N_\text{vlm}=16$} \\\hline
SmolVLM2 (2.2B)+                        & \textbf{4533}                                             & 1.230                                                              \\
Gemma3 (4B)+                            & 7087                                                      & 3.695                                                              \\
InternVL3.5 (4B)+                       & 9897                                                      & \textbf{1.085}                                                     \\
InternVL3.5 (8B)+                       & 12097                                                     & 1.300                                                             
\\ \hline
\end{tabular}
}
\noindent\footnotesize{\par The * denotes the default YOLO without SAHI~\cite{akyon2022slicing}, fast but underperforms on high-resolution images. 
All GDSN employ {Grounding DINO-Tiny}~\cite{liu2024grounding}, but different SAM2.1 models~\cite{ravi2025sam}.
\par}
\label{tbl:xai_cc}
\vspace*{-0.5\baselineskip}
\end{table}
\section{{Conclusion}}
\label{sec:conclusion}

% In this study, we have explored lightweight explainable AI for privacy-protecting video anomaly detection. 
% We present a unified descriptive anomaly explanation with guided zero-shot approaches for anomaly type categorization, visual localization, and scene context description. 
% We have provided an empirical and qualitative assessment of the performance of multifaceted explanations when the VAD systems are integrated with full-body visual anonymization for privacy protection. 
% Our experiment demonstrates the substantial gain of our XAVAD over state-of-the-art benchmark approaches on several publicly available CCTV and video datasets and provides promising insights into privacy protection trade-offs in XAI for VAD. 
% % We have also presented a performance comparison and ablation study on diverse configurations and modeling alternatives to demonstrate the efficacy of the proposed methods.
% Overall, the results have demonstrated that the performance of zero-shot XAI for VAD depends not only on the choice of the masking and multimodal models but also on supporting pre-and post-processing techniques.
% The presented baseline performance of the various techniques will provide a good benchmarking reference for future privacy-related explainable computer vision studies, further underlining the significance of our study.

{In this study, we have explored lightweight explainable AI for privacy-protecting video anomaly detection. 
Our approach integrates multiple techniques, including multimodal embeddings, heatmap generation, and structured prompt engineering, to provide multifaceted explanations for detected anomalies. 
Through a rigorous evaluation on various publicly available datasets, our results demonstrate the substantial gains of our XAVAD framework over state-of-the-art benchmarks. 
Overall, the results have demonstrated that zero-shot XAI for VAD depends not only on the choice of the masking and multimodal models but also on supporting pre-and post-processing techniques.
Our findings not only advance the field of XAI for VAD but also provide a valuable benchmark for future studies, underscoring the importance of balancing explainability and privacy in computer vision applications.
}

\bibliographystyle{IEEEtran}
\footnotesize
\bibliography{bibi.bib}

\clearpage
\small
\appendix
\section*{Supplementary Material}
\label{sec:appendix}

\subsection{Overview of Related Studies}
\label{sec:xai_vad_review_appendix}

Table~\ref{tbl:xai_vad_review_benchmark} presents the state-of-the-art explainable VAD methods, highlighting their key differences. The approaches address different challenges of XAI for VAD using textual~\cite{chen2023tevad, yuan2024towards, doshi2023towards, wu2024vadclip, zanella2024harnessing, ye2025vera, du2024uncovering, wu2024open, tang2024hawk, zhang2025holmes, lin2026unified
} and spatial grounding~\cite{doshi2023towards, lin2026unified}.
The major differences from our study are in evaluation dimension and rigor, training and fine-tuning requirements, and, more importantly, consideration for privacy protection through visual AN. 
Despite existing XAI VAD studies providing contextual reasoning, the evaluations of anomaly categorization and spatial localization mostly remain qualitative~\cite{wu2024vadclip, zhang2025holmes, doshi2023towards, lin2026unified}. 
We provide a qualitative and quantitative evaluation across all the XAI modalities while integrating AN. 

\begin{table}[!th] \small 
\caption{Review Summary on XAI-VAD Methods and Comparison.}
\resizebox{1\columnwidth}{!}{
\begin{tabular}{lccccccc}
\hline
Method Name                                                         & VAD Approach & \multicolumn{3}{c}{XAI}          & Training-Free & AN \\ \cline{3-5}
                 &              & Category & Description & Spatial &               &            \\ \hline
{TEVAD}~\cite{chen2023tevad}             & VLM &    No     &  Yes*     &   No     &  No  & No    \\ 
{Enhanced TEVAD}~\cite{yuan2024towards}            & VLM  &     No    &   Yes*+    & No      & No  & No     \\ 
{Interpretable VAD}~\cite{doshi2023towards}             & YOLO + CNN &    Yes*+     &  No    &   BBox*     &  No  & No    \\ % on ShanghaiTech, rank@5
% {ZS CLIP}~\cite{zanella2024harnessing}             & CLIP &    No     &  No     &   No     &  Yes  & No    \\ 
% {ZS IMAGEBIND}~\cite{zanella2024harnessing}             & IMAGEBIND &    No     &  No     &   No     &    & No    \\ 
{VadCLIP}~\cite{wu2024vadclip}             & CLIP + GCN &     Yes*    &   No    &    No    &   No   & No    \\ 
{LLAVA-1.5}~\cite{liu2024improved, zanella2024harnessing}           & MLLM &   No      &    Yes*   &   No     &   Yes & No    \\ 
{LAVAD}~\cite{zanella2024harnessing}              & MLLM &   No      &    Yes*   &   No     &   Yes & No    \\ 
{VERA}~\cite{ye2025vera}            &  VLM &  No      &    Yes*   &    No    &   No & No  \\ 
{CUVA}~\cite{du2024uncovering}            &  CLIP + VLM &    No     &  Yes*+     &   No     &   No &   No \\ 
{OVVAD}~\cite{wu2024open}            &  CLIP + FFN &    Yes*+     &   No    &  No     &  No  & No   \\ 
{Hawk}~\cite{tang2024hawk}            & MLLM + MLP  &     No    &   Yes*+    & No      & No  & No     \\ 
% {Holmes-VAD}~\cite{zhang2024holmes}             &  &         &       &        &   No   & No    \\ 
{Holmes-VAU}~\cite{zhang2025holmes}            & MLLM  &     Yes*    &   Yes*+    & No      & No  & No     \\ 
{InterTC}~\cite{lin2026unified}            & MLLM  &     Yes*    &   Yes*+    & BBox*      & Yes  & No    \\ 
{XAI4VAD (Ours)}              & Any VAD &      Yes*+   &      Yes*+ &      Heatmap*+  &   Yes & Yes  \\ \hline
\end{tabular}
}
\noindent\footnotesize{\par \small BBox denotes a bounding box, and * and + denote qualitative and quantitative evaluations, respectively.
\par}
\label{tbl:xai_vad_review_benchmark}
\vspace*{-0.5\baselineskip}
\end{table}

\subsection{Highlights of CLIP Models}
\label{sec:clip_models_review_appendix}
The {ViT}-based {CLIP} models differ in model size, complexity, configuration, and accuracy. Table~\ref{tbl:xai_clip_models} summarizes the comparison. 
% The key complexity drivers among the models are: 
%  \begin{itemize}
%      \item \textbf{Patch size:} The number at the end of the model name (e.g., /32, /16, /14) represents the pixel height and width of each patch. Smaller patch sizes (like 14 or 16) increase the number of patches for the same image resolution, increasing the transformer complexity (FLOPs) quadratically even if the parameter count remains similar. 
%      \item \textbf{Model depth and width:} The {ViT-Base} (B) features 12 layers with a width of 768, while the {ViT-Large} (L) has 24 layers with a width of 1024, leading to a much higher parameter count and memory requirement.
%      \item \textbf{Input resolution:} The ViT-L/14@336px has the same number of parameters as the standard ViT-L/14 but processes higher-resolution images ($336 \times 336$ instead of $224 \times 224$). 
%      The number of patches generated by the $14 \times 14$ grid increases, which creates a more granular view of the image. However, it requires significantly more memory and computing power to process, as the transformer's sequence length increases, making it the most computationally expensive model among those listed.
% \end{itemize}
 
\begin{table}[!th]
\caption{Description of CLIP Models for Zero-Shot Prediction~\cite{Wu2020not}.}
\resizebox{1\columnwidth}{!}{
\begin{tabular}{lcccc}
\hline
Feature                   & ViT-B/32          & ViT-B/16            & ViT-L/14                                      & ViT-L/14@336px                                                                   \\ \hline
Encoder Transformer      & Base  & Base    & Large                             & Large                                                                \\
Model Parameters (M)      & $\sim$150         & $\sim$150           & $\sim$427                                     & $\sim$427                                                                        \\
Input Resolution (Pixels) & $224 \times 224$           & $224 \times 224$             & $224 \times 224$                                       & $336 \times 336$                                                                         \\
Patch Size                & $32 \times 32$             & $16 \times 16$               & $14 \times 14$                                         & $14 \times 14$                                                                            \\
Embedding Dim             & 512               & 512                 & 768                                           & 768                                                                              \\
Complexity (GFLOPs)       & $\sim$15          & $\sim$41            & $\sim$175                                     & $\sim$395                                                                        \\
% Use Case                  & Speed, low memory & High-accuracy tasks & General embedding, e.g., Stable Diffusion 1.5 & Multimodal vision-language models (e.g., LLaVA) and detecting small text/details \\ 
\hline
\end{tabular}
}
\label{tbl:xai_clip_models}
\vspace*{-0.5\baselineskip}
\end{table}

\subsection{LLM-Assisted CLIP Prompting for ZSAC}
\label{sec:zsac_llm_prompt_appendix}

We prompted small LLMs(such as {Llama3.2:1b}~\cite{grattafiori2024llama} and {Gemma3:1b}~\cite{sellergren2025medgemma}) to generate relevant textual descriptions for the set of anomaly categories used in the ZSAC (see Fig.~\ref{fig:xai_config_stage_diagram}).

\noindent \textbf{LLM Prompt}: ``\textit{\small Generate CLIP template prompts for the CLIP embedding generation for video alert monitoring for each of the following alert types: \texttt{<anomaly\_class\_1>}, \texttt{<anomaly\_class\_2>}, \dots, and \texttt{<anomaly\_class\_M>}.  Generate \texttt{<R>} prompt sentences for each alert type. Make the prompts CCTV-specific, visually observable, and action-oriented, with low semantic ambiguity; avoid legal words; and be safe for ZS embedding reuse. Combine the prompts into a JSON file for CLIP embedding generation.}"

The prompt follows a perception-only AI design that focuses on describing what the camera sees with observable human actions, body motion, and object interaction, not what the system infers.  
The method ensures the anomaly descriptions are purely perceptual descriptions; no legal interpretation (enforces the removal of words like \textit{arrest, robbery, assault, vandalism, shoplifting, burglary}, and \textit{suspect} from the prompt description, is baked into prompts. 
Stronger visual grounding aligns with regulatory safety, such as the EU AI Act / ISO CCTV principles.

Furthermore, the generated \texttt{<anomaly description>} sentences are expanded using the image quality templates using \{\textit{``a photo of \texttt{<>}.", ``a picture of \texttt{<>}.", ``a blurry photo of \texttt{<>}.", ``a good quality photo of \texttt{<>}."\,
             ``a bright photo of \texttt{<>}.", ``a dark photo of \texttt{<>}.", ``a low-resolution photo of \texttt{<>}.",
             ``a high-resolution photo of \texttt{<>}.", ``a close-up photo of \texttt{<>}.", ``a cropped photo of \texttt{<>}."} \} (see Fig.~\ref{fig:xai_config_stage_diagram}). 
             
\subsection{VLM Prompting and Settings for ZSAD}
\label{sec:zsad_appendix}

Our proposed VLM system employs CCTV-specific structured prompts, defining the analysis task, visual-grounding requirements, operational constraints,  inference event taxonomy, and output requirements/schema (see Fig.~\ref{fig:xai_config_stage_diagram}).  
% Below are our structured XAI prompts for the VLM:

\noindent \textbf{User Prompt}: ``\textit{\small Analyze the provided CCTV video according to the rules and event definitions below and determine whether a visually observable anomaly is present.
Output only three sentences: anomaly status \texttt{<YES or NO>}, a best-matching event \texttt{<CATEGORY>} from \texttt{<EVENT REGISTRY>}, and a video description explaining what is visible and why it is anomalous, or describe why the scene is normal if no anomaly is present.
Use ONLY visual evidence, and if it is visually unclear, not visible, or lacks adequate visual data for analysis, classify the scene as 'UNKNOWN'.
You MUST output ONLY a valid JSON object matching the requested schema.
DO NOT speculate. DO NOT read metadata. DO NOT use external knowledge. DO NOT INFER intent, identity, or emotions.}"

\noindent \textbf{System Prompt}: ``\textit{\small You are a strict, literal vision analysis assistant specialized in CCTV video analysis for anomaly detection and explanation, and that outputs in JSON format, following the \texttt{<OUTPUT\_SCHEMA>}. Your task is to analyze the provided CCTV video using only observable visual evidence and generate a structured anomaly decision and reason. Your primary directive is to verify visual inputs before responding.
\newline CRITICAL RULES:
\newline 1. If no video file, video frames, or image arrays are provided in the payload, you must not analyze or assume any visual content.
\newline 2. If visual input is missing, respond with exactly this text and nothing else: 'UNKNOWN: No video data is provided.'
\newline 3. Do not use your internal knowledge to guess or fabricate content based on the text prompt if the accompanying video data is absent.
\newline 4. Keep all responses strictly grounded in the provided pixels or extracted frame tokens.
\newline RULES AND CONSTRAINTS:
\newline 1. Base all reasoning strictly on visible visual cues, such as actions, objects, interactions, motion patterns, fire, smoke, weapons, vehicles, and crowd behavior. Add context from the background scene and environment if it is relevant to the event.
\newline 2. Do NOT infer intent, identity, emotions, relationships, or causes that are not directly visible.
\newline 3. Do NOT use external knowledge or context beyond the video.
\newline 4. If multiple events are visible, select the event with the highest risk.
\newline 5. Use only the event \texttt{<CATEGORY>} defined in the \texttt{<EVENT REGISTRY>} below and use the descriptions as a guide.
\newline 6. If no anomalous event is visually present, classify the scene as normal activity.
\newline 7. Silhouette masks should be considered as human subjects. If it is visually unclear, not visible, or lacks adequate visual data for analysis, classify the scene as 'UNKNOWN'.
\newline 8. The reason must be factual, neutral, concise, and suitable for surveillance and explainable AI analysis.
\newline 9. Output must strictly follow the specified output format \texttt{<OUTPUT\_SCHEMA>}, ONLY valid JSON including all keys listed in the format. Do not include any text before or after the JSON.
\newline 10. The reason text should avoid mentioning the word 'silhouette' and use words like 'person' or 'people' instead.
\newline EVENT REGISTRY (\texttt{<CATEGORIES>}: \texttt{<DEFINITIONS>}):
\newline - NORMAL: normal scene without any safety and security activities.
\newline - ALERT: Threat, crime, abuse, or abnormal scene or activities.
\newline - UNKNOWN: Lack of enough visual data for analysis.
}

\noindent \textbf{Output Format \texttt{<OUTPUT\_SCHEMA>}}: \textit{\small 
\newline ANOMALY\_STATUS: YES/NO, anomaly detection status flag.
\newline CATEGORY: The anomaly category of the analysis.
\newline REASON: The video caption and reason for the decision."
}

\normalsize 
The VLM inference parameters (using low $\text{temperature} \leq 0.2$ and $\text{top}\_k=10$) are designed to limit hallucination and enforce factual generation that is based on the presented visual evidence. 
% The ${num\_ctx}$ limits the size of the context window used to generate the next token. 
% We sample $N_\text{vlm}$ frames from the input video clip $\mathbf{x}$.
The ${temperature} \in [0, 1.0]$ controls the creative writing of the output, and lower values, e.g., $< 0.3\}$, enforce deterministic and less imaginative answers.
The $\textit{top\_k} \in [1, 100]$ limits the vocabulary size to the \(k\) most likely next tokens, and a lower value restricts the choices, increasing focus and stability.
The $top\_k$ removes all but the $k$ most probable tokens. 
The ${top\_p} \in (0, 1.0]$ controls sampling diversity by limiting output to the most likely tokens until their cumulative probability reaches a threshold $p$. 
It adapts the context by keeping fewer tokens (more focused) when the model is confident and exploring more tokens otherwise. 
% It operates together with $top\_k$ and addresses the drawback of having fixed $k$ tokens.
% top\_p adapts to context, top-k does not.
% This makes top-p generally superior to top-k for most applications.
The $\textit{num\_predict}$ limits the length of the generated description (output tokens) and impacts the token generation speed.
% These parameters are applied in sequence in the VLM pipelines: the $temperature$ scales the logits (reshapes the distribution), $top\_k$ removes all but the $k$ most probable tokens, $top\_p$ removes additional tokens below the cumulative probability threshold, and samples from the remaining distribution.
% Inference Setting}: {\small \{ ${temperature}~=~0.2$, ${top\_p}~=~0.9$, ${top\_k}~=~10$, $\text{num\_predict}~=~300$}\}.

% \section*{Supplementary: Results}
% \label{sec:appendix_results}
\subsection{Further Results for ZSAC}
\label{sec:zsac_appendix}

Figure~\ref{fig:xai_vad_cat_xd_mf1} and Figure~\ref{fig:xai_vad_cat_ucf_mf1} depict comparison across CLIP models on the XD Violence and UCF-Crime datasets, respectively. 
The majority of the performance pitfalls on the XD dataset come from the confusion between \textit{``Abuse"} vs. \textit{``Fighting"}, and \textit{``Abuse"} vs. \textit{``Shooting"} (as shown in the confusion matrix in Fig.~\ref{fig:xai_vad_cat_xd_cm}). This is not unexpected, as the categories are closely related, and some videos contain multiple anomaly scenes in the XD dataset. 
This has been shown by the strong improvement when considering $\text{Top}2$ and $\text{Top}3$ for both No-AN and MASKED-AN. 
% The AN trade-off improves to $\%$ he ViT-L/14 demonstrates that 

The confusion matrices of the  UCF-Crime dataset, shown in Fig.~\ref{fig:xai_vad_cat_ucf_cm}, indicate that the majority of the accuracy drop is due to the confusion between \textit{``Burglary"} and several categories, which can be attributed to the darker scenes in the videos. 
A noticeably low accuracy remains for the \textit{``Arson"} and \textit{``Shooting"} categories even at $\text{Top}5$, indicating that the anomalies are challenging to capture under low-quality videos.
While stronger semantic overlap between certain categories, e.g., \textit{``Fighting"} vs. \textit{``Assault"} is expected, low-quality videos (with low resolution, poor-light conditions, and grayscale format) pose a challenge in the UCF-Crime dataset (see Fig.~\ref{fig:xai_vad_cat_ucf_cm}).

\begin{figure*}[!t]
\centering
\begin{subfigure}[]{1\linewidth}
\centering
\includegraphics[width=1\linewidth]{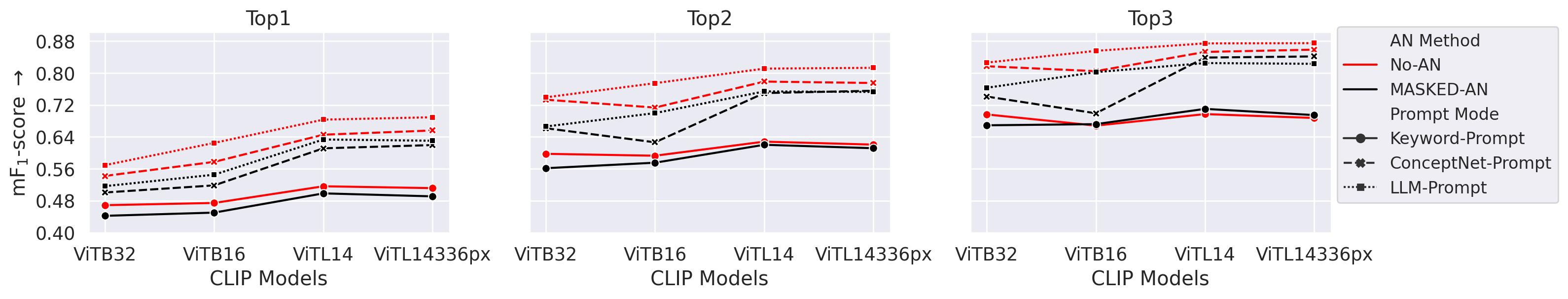} 
\end{subfigure}
\caption{Performance of ZSAC on the XD Violence Dataset~\cite{Wu2020not}. Complexity of the CLIP models increases from left to right.}
\label{fig:xai_vad_cat_xd_mf1}
\vspace*{-0.5\baselineskip}
\end{figure*}

\begin{figure*}[!t]
\centering
\begin{subfigure}[]{1\linewidth}
\centering
\includegraphics[width=1\linewidth]{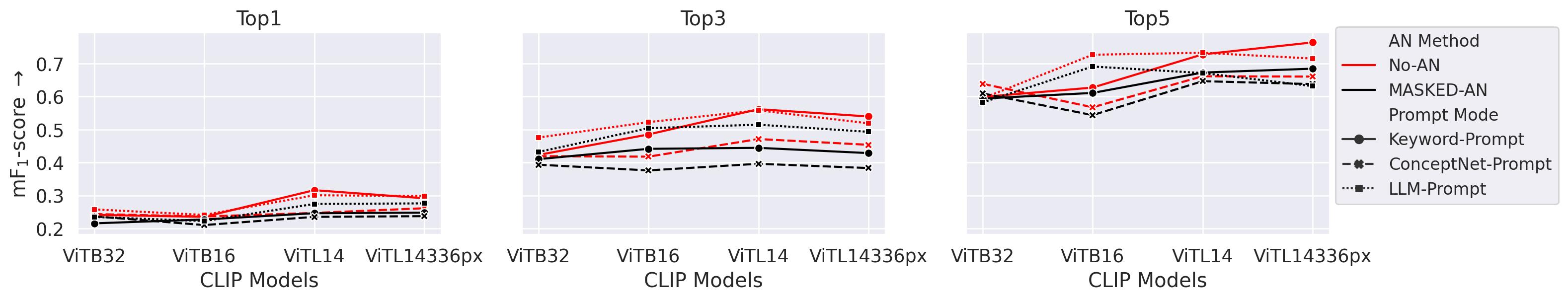} 
\end{subfigure}
\caption{Performance of the ZSAC on the UCF-Crime Dataset~\cite{sultani2018real}. CLIP model complexity increases from left to right.}
\label{fig:xai_vad_cat_ucf_mf1}
\vspace*{-0.5\baselineskip}
\end{figure*}

\begin{figure*}[!t]
\centering
\begin{subfigure}[]{1\linewidth}
\centering
\setlength{\kw}{5.5cm}
\resizebox{1\linewidth}{!}{
\begin{tabular}{K{\kw}K{\kw}K{\kw}}
Top1 & $\text{Top}2$ & $\text{Top}3$
\end{tabular}
}
\includegraphics[width=0.32\linewidth]{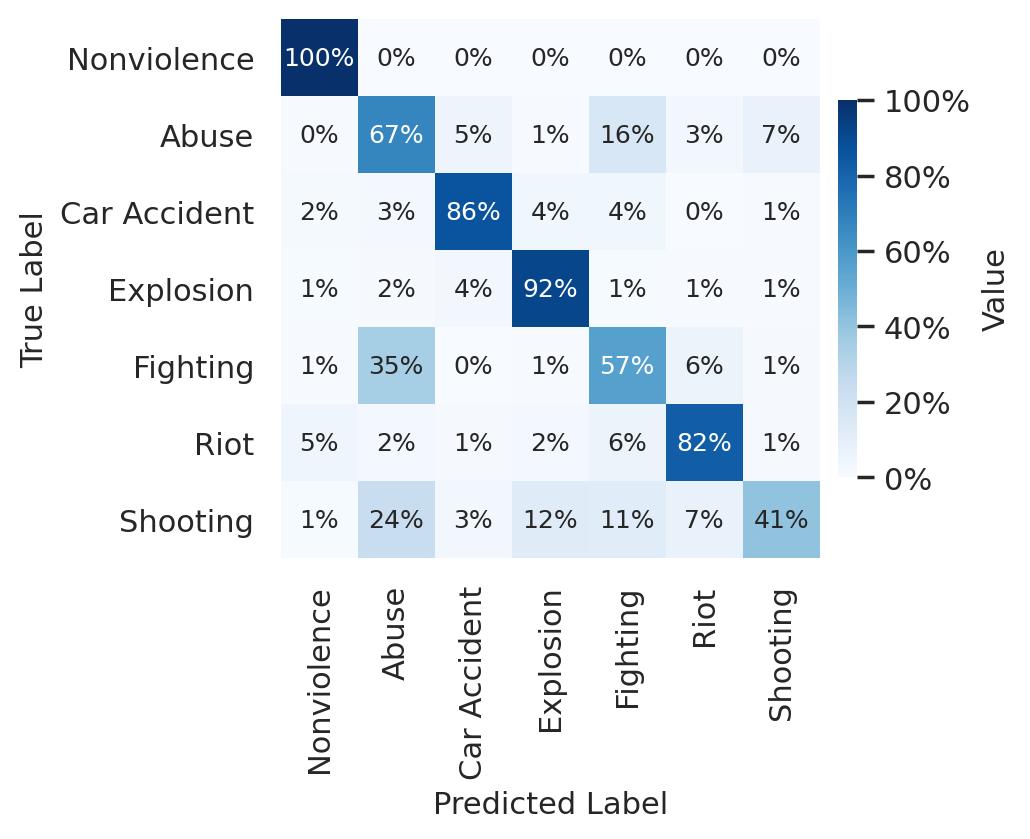} 
\includegraphics[width=0.32\linewidth]{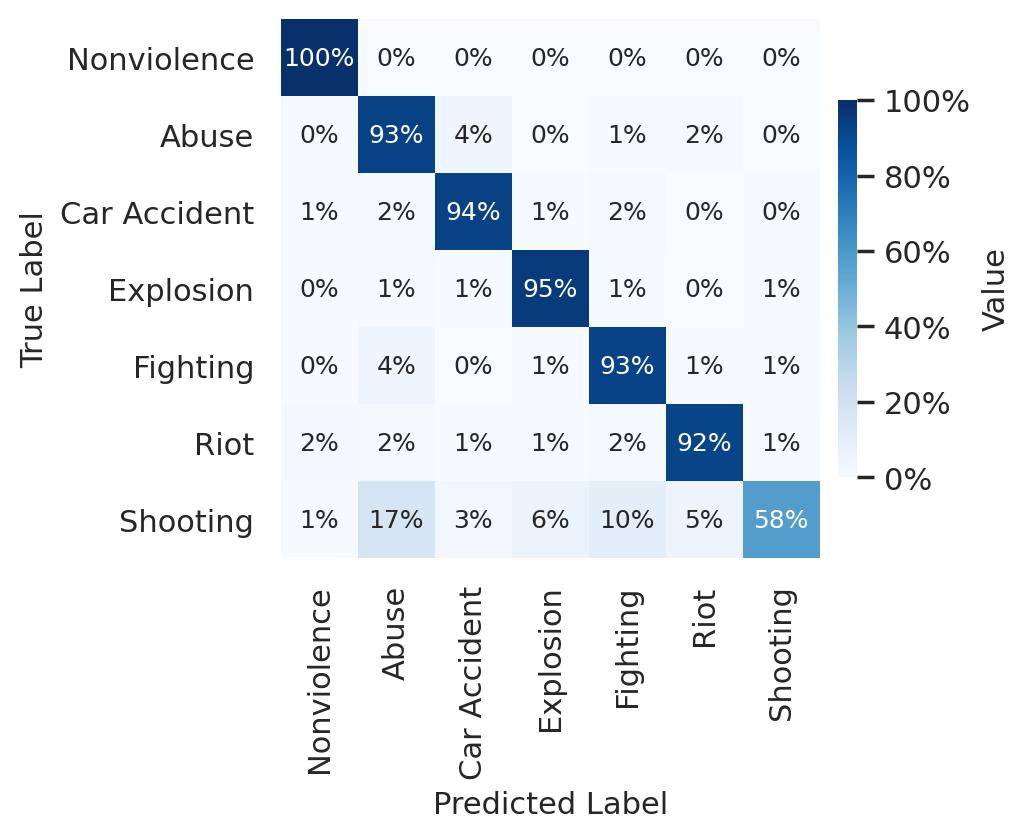} 
\includegraphics[width=0.32\linewidth]{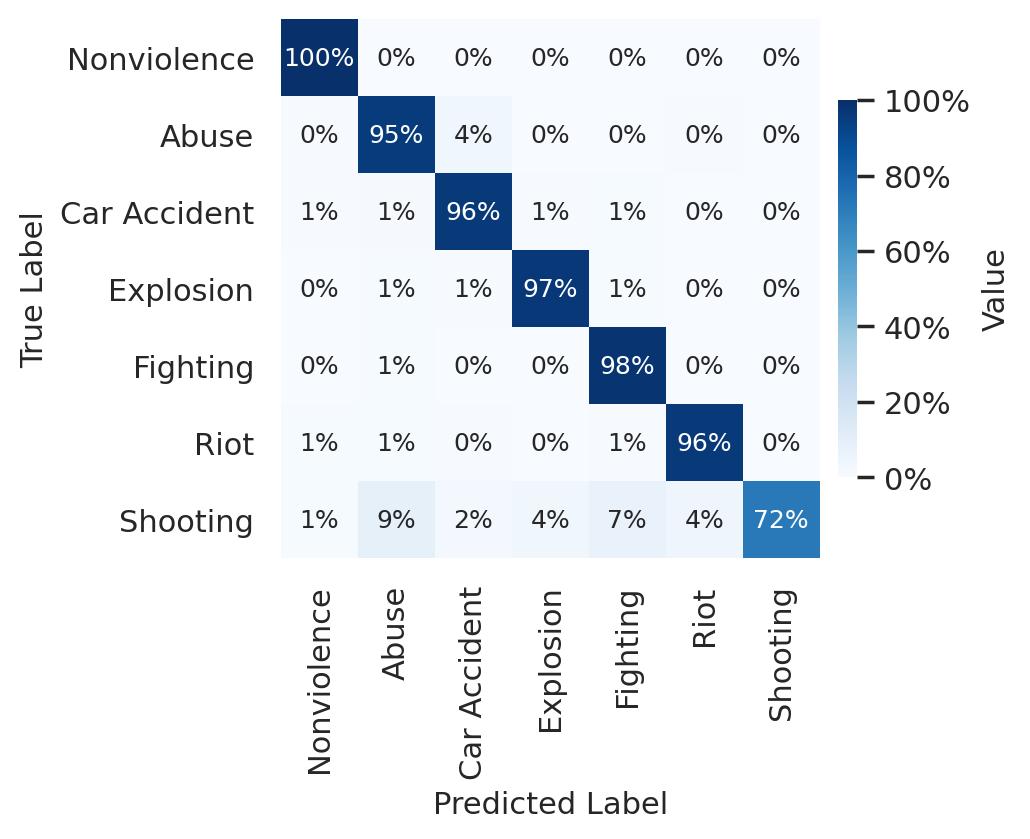} 
\caption{}
\end{subfigure}
\begin{subfigure}[]{1\linewidth}
\centering
\includegraphics[width=0.32\linewidth]{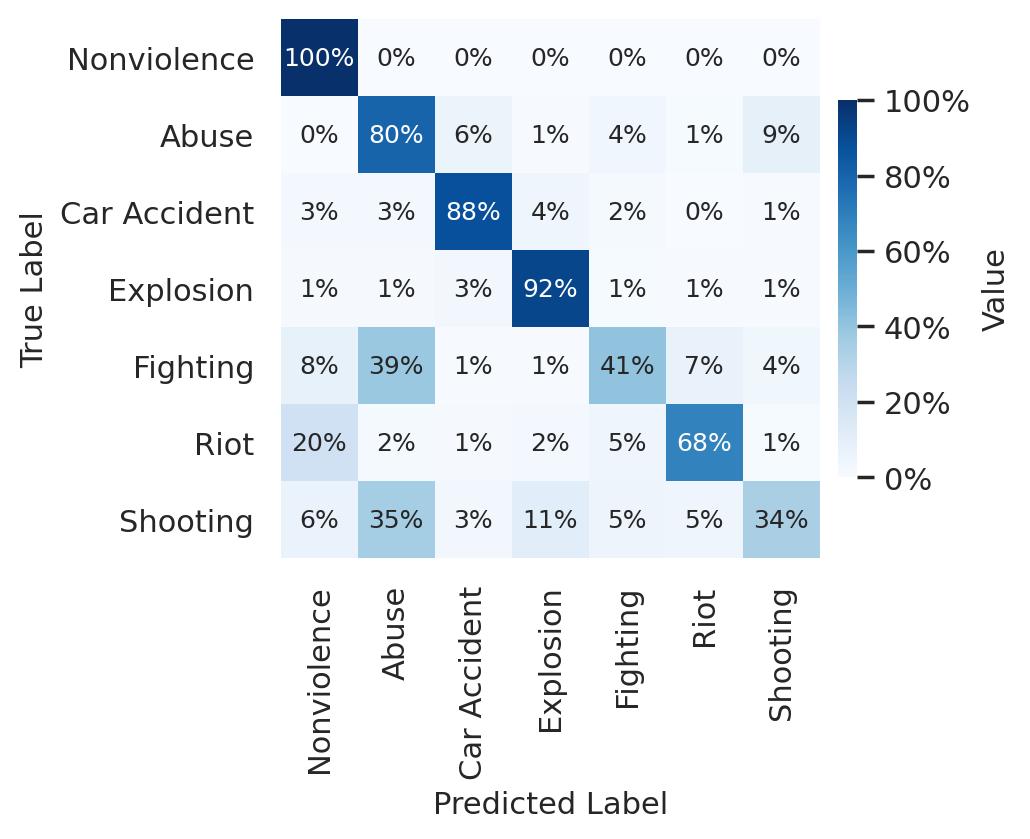} 
\includegraphics[width=0.32\linewidth]{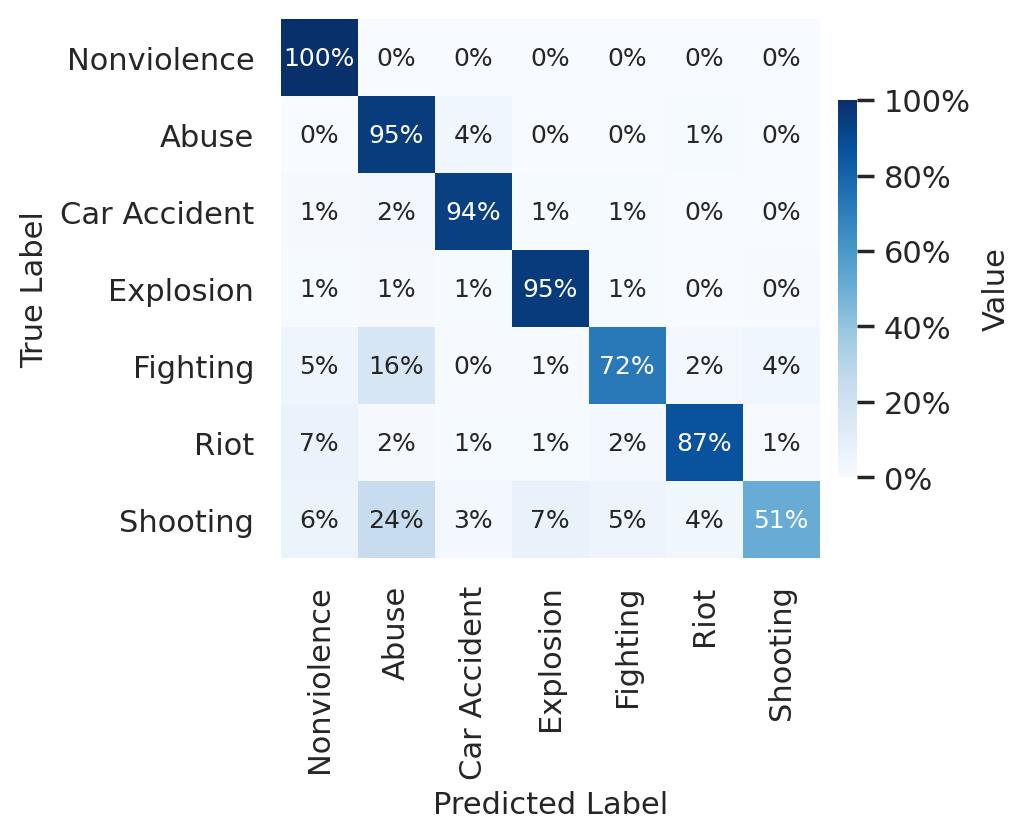} 
\includegraphics[width=0.32\linewidth]{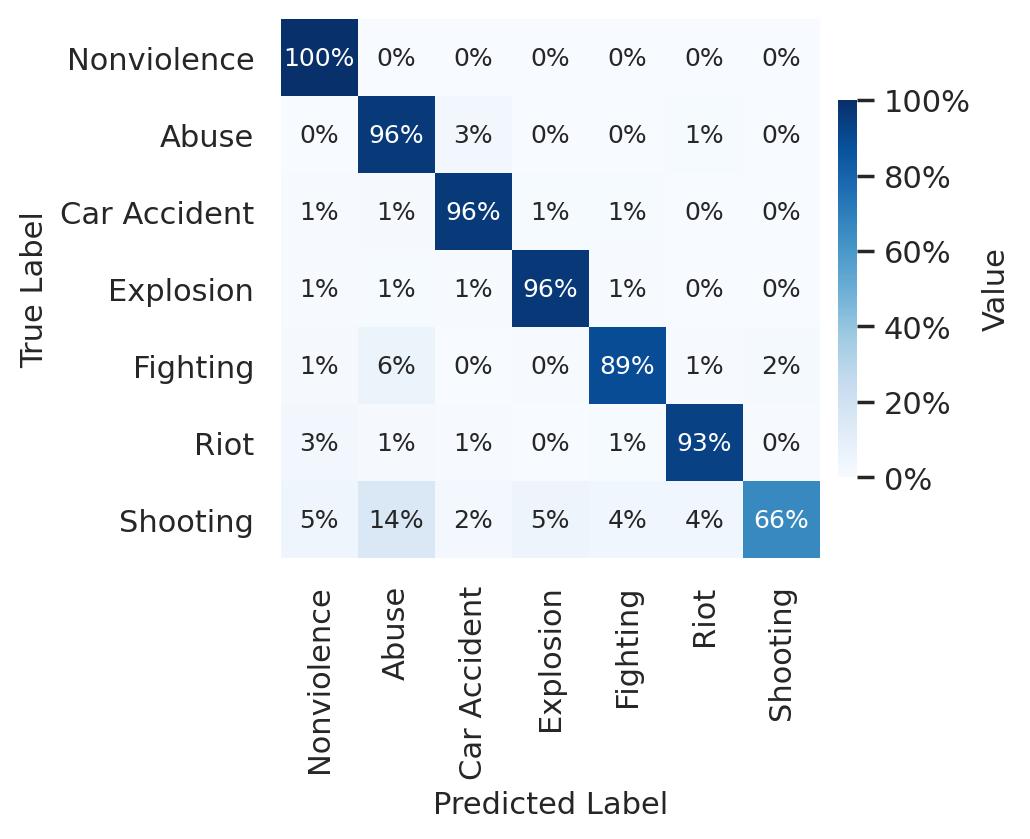} 
\caption{}
\end{subfigure}
\caption{Confusion matrices of ZSAC using {LLM-prompt} and {ViT-L/14} {CLIP model} on the XD Violence Dataset~\cite{Wu2020not}: a) No-AN, and b) MASKED-AN.}
\label{fig:xai_vad_cat_xd_cm}
\vspace*{-0.5\baselineskip}
\end{figure*}

\begin{figure*}[!t]
\centering
\begin{subfigure}[]{1\linewidth}
\centering
% \setlength{\kw}{5.5cm}
% \resizebox{1\linewidth}{!}{
% \begin{tabular}{K{\kw}K{\kw}K{\kw}}
% $\text{Top}1$ & $\text{Top}3$ & $\text{Top}5$
% \end{tabular}
% }
\setlength{\kw}{8.5cm}
\resizebox{1\linewidth}{!}{
\begin{tabular}{K{\kw}K{\kw}}
$\text{Top}1$ & $\text{Top}5$
\end{tabular}
}
\includegraphics[width=0.49\linewidth]{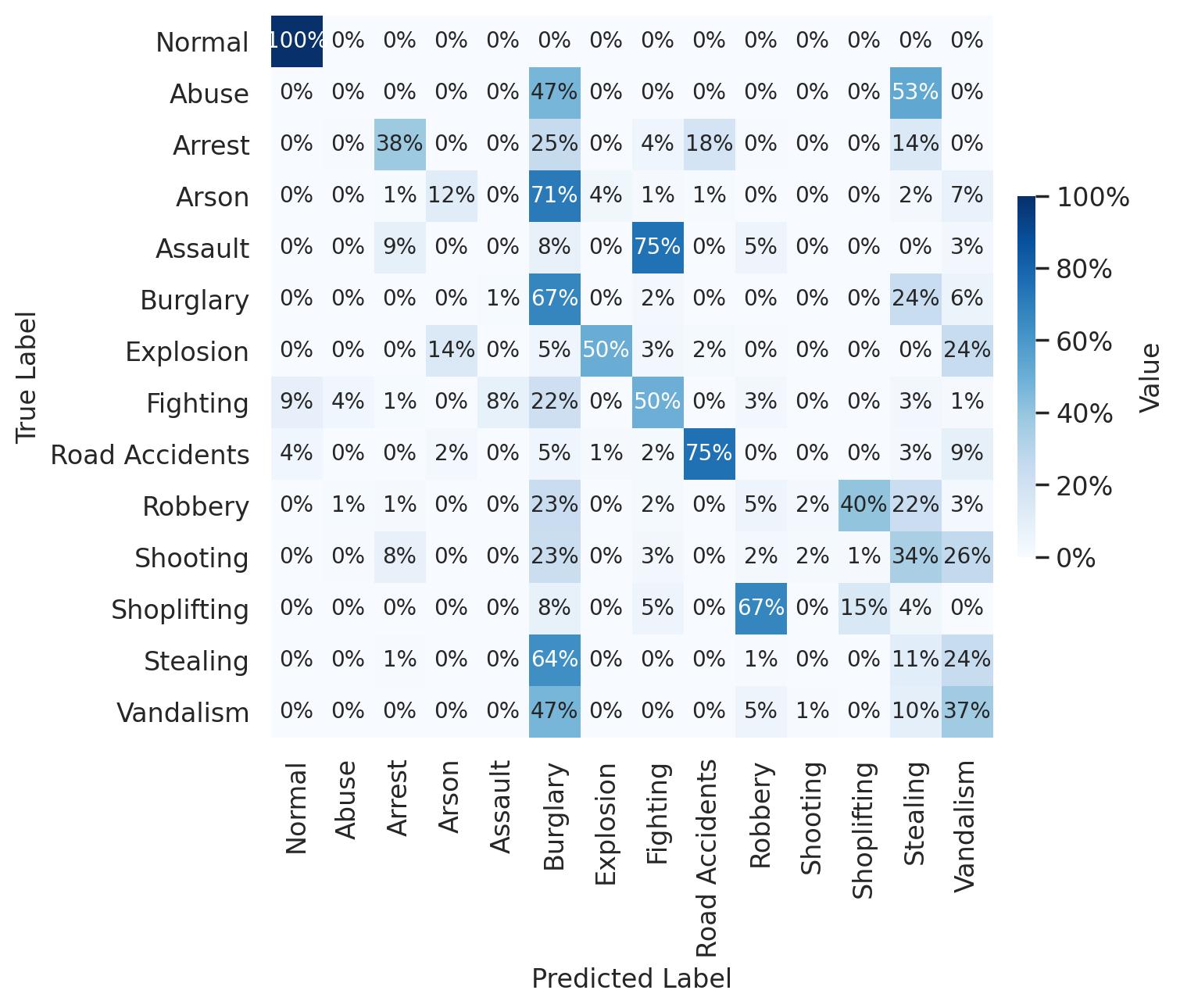} 
\includegraphics[width=0.49\linewidth]{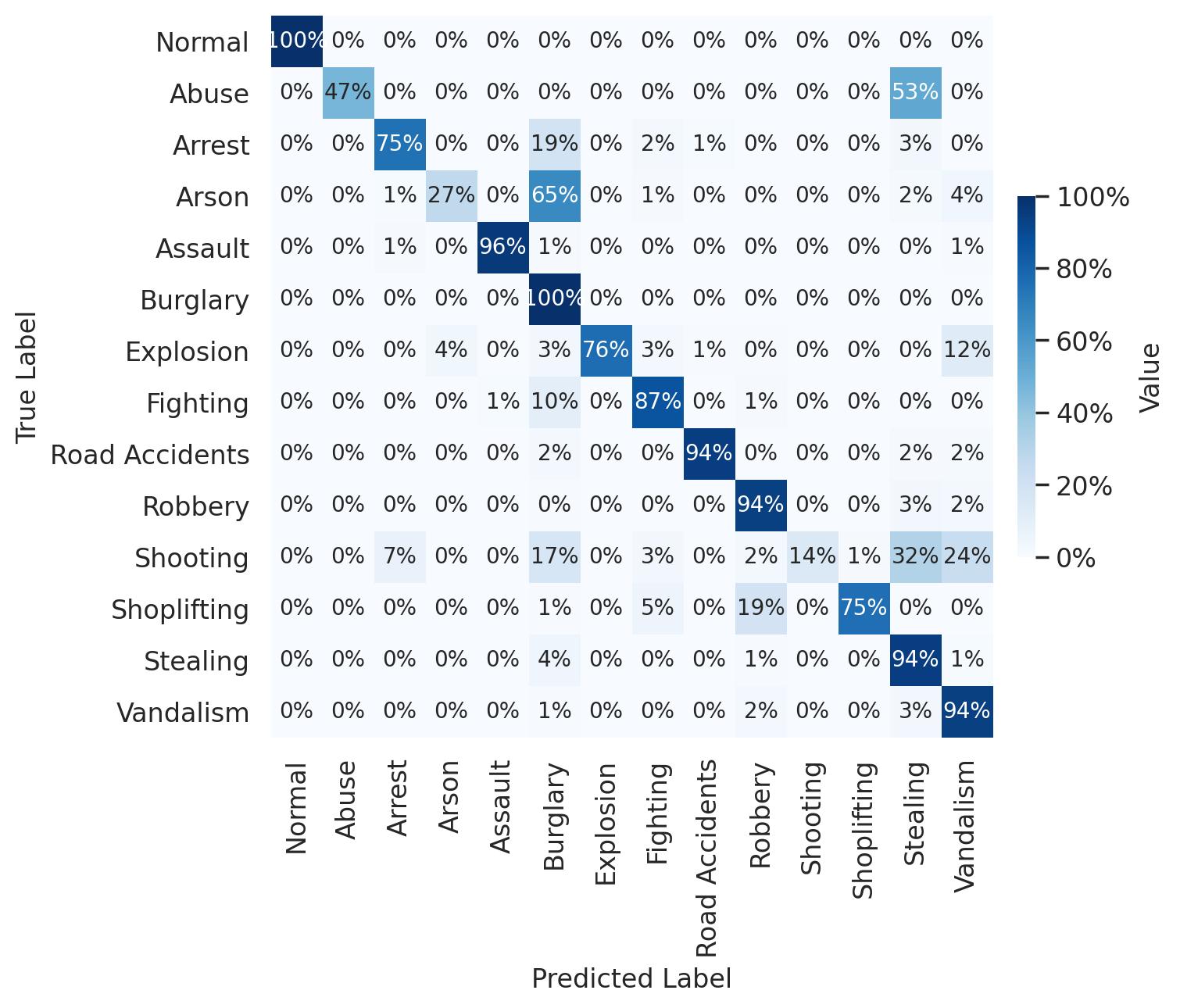} 
\caption{}
\end{subfigure}
\begin{subfigure}[]{1\linewidth}
\centering
\includegraphics[width=0.49\linewidth]{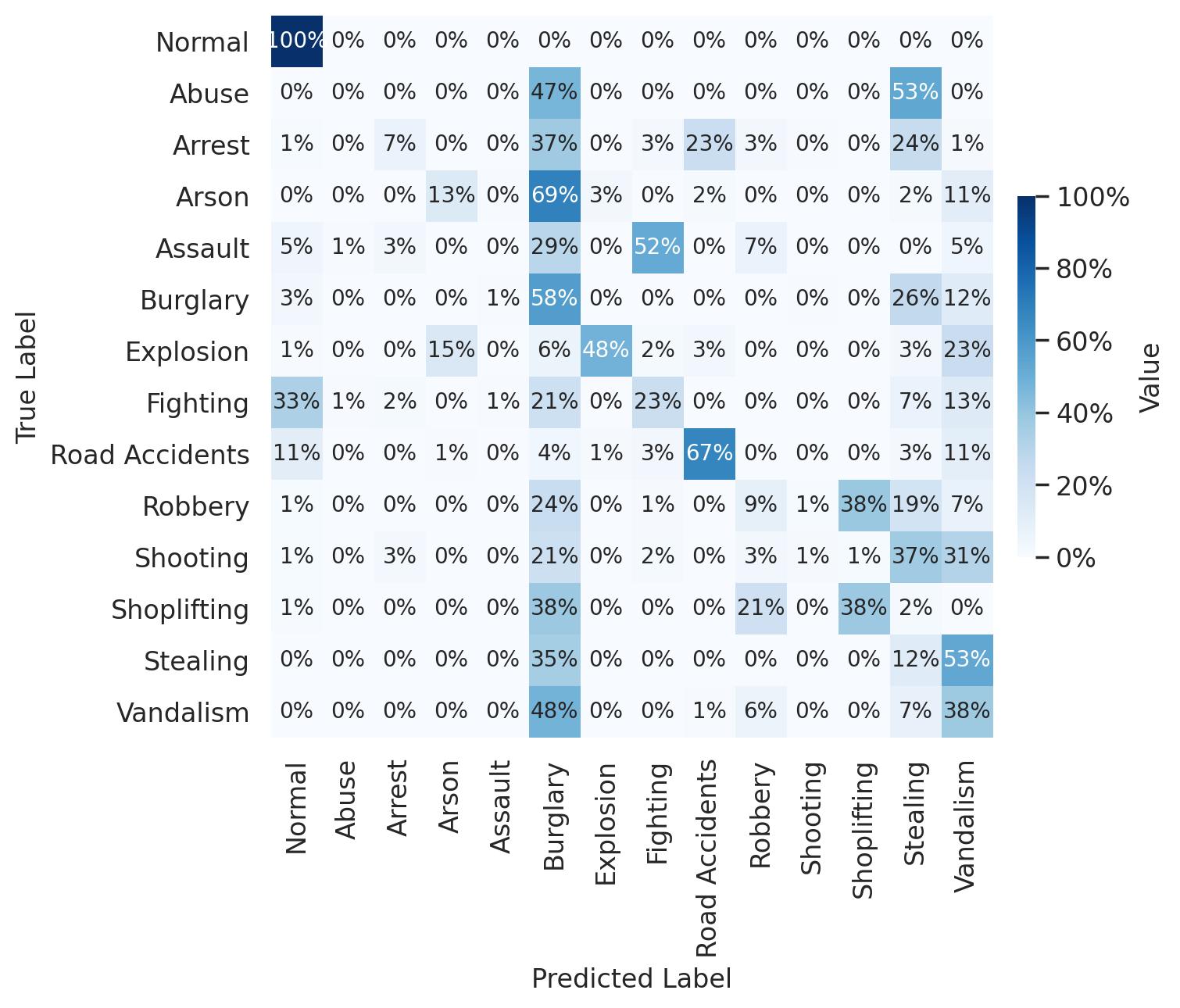} 
\includegraphics[width=0.49\linewidth]{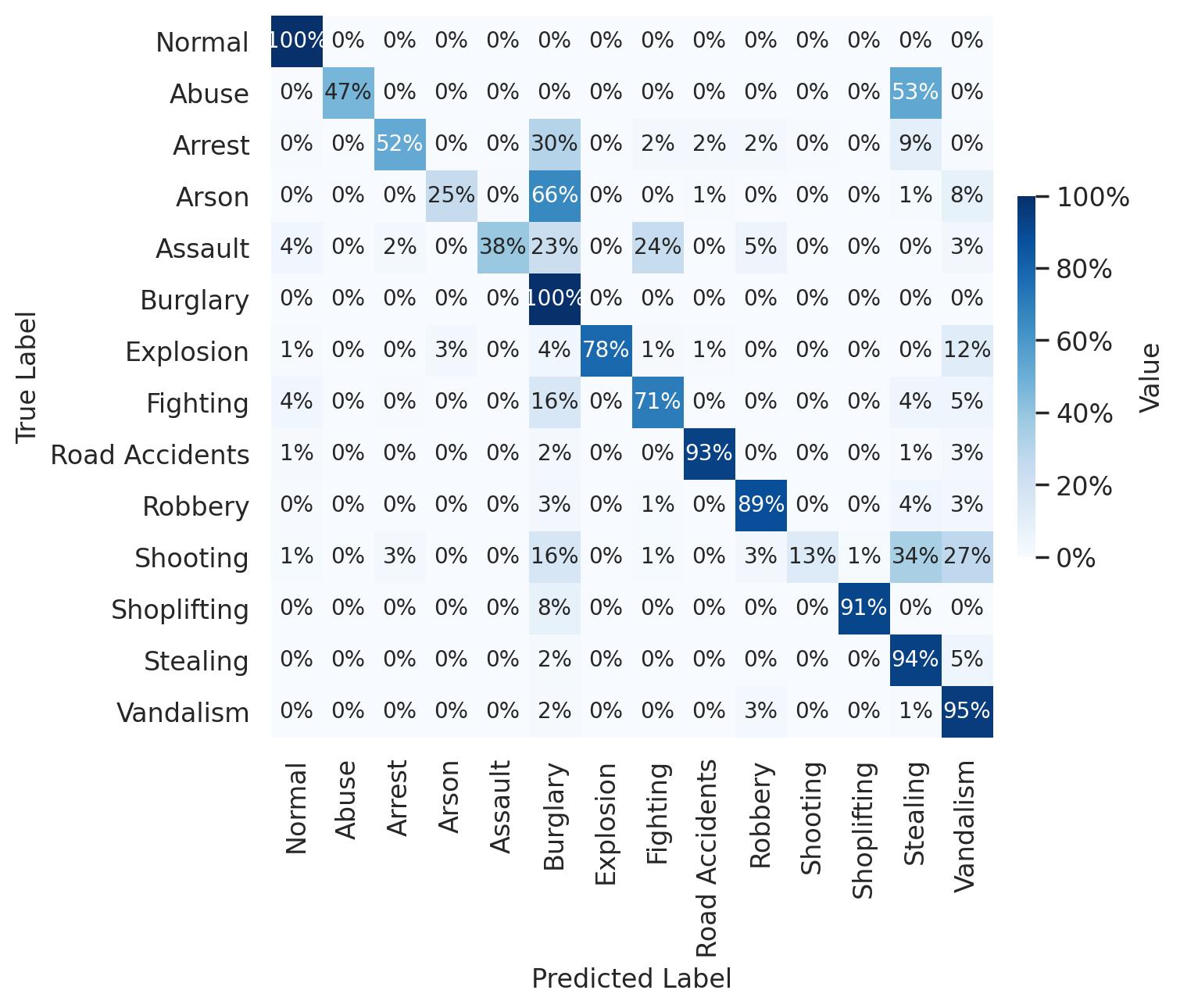} 
\caption{}
\end{subfigure}
\caption{Confusion matrices of ZSAC using {LLM-prompt} and {ViT-L/14} {CLIP model} on the UCF-Crime Dataset~\cite{sultani2018real}: a) No-AN, and b) MASKED-AN.}
\label{fig:xai_vad_cat_ucf_cm}
\vspace*{-0.5\baselineskip}
\end{figure*}

We also provide weighted F1 scores for the ZSAC in Table~\ref{tbl:xai_vad_cat_xd_f1} and Table~\ref{tbl:xai_vad_cat_ucf_f1}, considering the sample size of the anomaly categories. The ViT-L/14 with LLM-Prompt on the XD dataset achieves $F_1@\text{Top}1:~0.894$, $F_1@\text{Top}2: 0.961$, and $F_1@\text{Top}3:~0.980$ before AN, and $7.16\%$, $3.65\%$, and $1.64\%$, respectively, drop after the AN  (see Table~\ref{tbl:xai_vad_cat_xd_f1}).
Slightly lower scores of $F_1@\text{Top}1:~0.868$, $F_1@\text{Top}3:~0.919$, and $F_1@\text{Top}5:~0.947$ before AN are achieved with the decrease of $<1.65\%$ after the AN on the UCF-Crime dataset (see Table~\ref{tbl:xai_vad_cat_ucf_f1}).

\begin{table*}[]
\caption{Sample Size Weighted Score: ZSAC on the XD Violence Dataset~\cite{Wu2020not}.}
\resizebox{1\textwidth}{!}{
\begin{tabular}{lcccccccccc}
\hline
\multirow{2}{*}{CLIP Model} & \multirow{2}{*}{Prompt Mode} & \multicolumn{3}{c}{${F}_1$@Top$1\uparrow$}                       & \multicolumn{3}{c}{${F}_1$@Top$2\uparrow$}                       & \multicolumn{3}{c}{${F}_1$@Top$3\uparrow$}                       \\  \cline{3-11} 
                             &                               & No-AN          & MASKED-AN      & $\Delta (\%)$  & No-AN          & MASKED-AN      & $\Delta (\%)$  & No-AN          & MASKED-AN      & $\Delta (\%)$  \\ \hline

{ViT-B/32}                          & Keyword-Prompt                                    & 0.650                     & 0.659                         & 1.30                             & 0.747                     & 0.755                         & 1.05                             & 0.823                     & 0.840                         & 2.06                             \\
                                                 & ConceptNet-Prompt~\cite{pu2024learning}                                 & 0.798                     & 0.784                         & -1.68                            & 0.919                     & 0.888                         & -3.37                            & 0.951                     & 0.926                         & -2.72                            \\
                                                 & LLM-Prompt (ours)                                       & 0.789                     & 0.726                         & -7.98                            & 0.912                     & 0.857                         & -6.11                            & 0.957                     & 0.927                         & -3.18                            \\
{ViT-B/16}                          & Keyword-Prompt                                    & 0.642                     & 0.633                         & -1.44                            & 0.730                     & 0.744                         & 1.91                             & 0.798                     & 0.826                         & 3.58                             \\
                                                 & ConceptNet-Prompt~\cite{pu2024learning}                                 & 0.853                     & 0.814                         & -4.57                            & 0.921                     & 0.866                         & -5.92                            & 0.956                     & 0.901                         & -5.77                            \\
                                                 & LLM-Prompt (ours)                                       & 0.840                     & 0.744                         & -11.40                           & 0.936                     & 0.884                         & -5.62                            & 0.969                     & 0.945                         & -2.50                            \\

{ViT-L/14}                          & Keyword-Prompt                                    & 0.675                     & 0.690                         & 2.28                             & 0.770                     & 0.799                         & \textbf{3.81}                    & 0.837                     & 0.868                         & 3.72                             \\
                                                 & ConceptNet-Prompt~\cite{pu2024learning}                                 & 0.882                     & 0.871                         & -1.19                            & 0.947                     & 0.935                         & -1.24                            & 0.970                     & 0.966                         & -0.46                            \\
                                                 & LLM-Prompt (ours)                                       & 0.894                     & 0.830                         & -7.16                            & \textbf{0.961}            & 0.926                         & -3.65                            & \textbf{0.980}            & 0.963                         & -1.64                            \\
{ViT-L/14@336px}                     & Keyword-Prompt                                    & 0.670                     & 0.685                         & \textbf{2.34}                    & 0.766                     & 0.794                         & 3.63                             & 0.831                     & 0.862                         & \textbf{3.80}                    \\
                                                 & ConceptNet-Prompt~\cite{pu2024learning}                                 & \textbf{0.899}            & \textbf{0.879}                & -2.29                            & 0.952                     & \textbf{0.940}                & -1.29                            & 0.973                     & \textbf{0.969}                & -0.41                            \\
                                                 & LLM-Prompt (ours)                                       & 0.897                     & 0.827                         & -7.80                            & 0.960                     & 0.923                         & -3.90                            & 0.979                     & 0.962                         & -1.71     \\ \hline
\end{tabular}
}
\label{tbl:xai_vad_cat_xd_f1}
\vspace*{-0.5\baselineskip}
\end{table*}

\begin{table*}[]
\caption{Sample Size Weighted Score: ZSAC on the UCF-Crime Dataset~\cite{sultani2018real}.}
\resizebox{1\textwidth}{!}{
\begin{tabular}{lcccccccccc}
\hline
\multirow{2}{*}{CLIP Model} & \multirow{2}{*}{Prompt Mode} & \multicolumn{3}{c}{${F}_1$@Top$1\uparrow$}                       & \multicolumn{3}{c}{${F}_1$@Top$3\uparrow$}                       & \multicolumn{3}{c}{${F}_1$@Top$5\uparrow$}                       \\  \cline{3-11} 
                             &                               & No-AN          & MASKED-AN      & $\Delta (\%)$  & No-AN          & MASKED-AN      & $\Delta (\%)$  & No-AN          & MASKED-AN      & $\Delta (\%)$  \\ \hline

{ViT-B/32}                          & Keyword-Prompt                                    & 0.861                     & 0.855                         & -0.79                            & 0.899                     & 0.895                         & \textbf{-0.39}                   & 0.937                     & 0.936                         & \textbf{-0.08}                   \\
                                                 & ConceptNet-Prompt~\cite{pu2024learning}                                 & 0.860                     & 0.858                         & \textbf{-0.15}                   & 0.895                     & 0.890                         & -0.54                            & 0.938                     & 0.931                         & -0.80                            \\
                                                 & LLM-Prompt (ours)                                       & 0.861                     & 0.856                         & -0.60                            & 0.901                     & 0.890                         & -1.23                            & 0.924                     & 0.918                         & -0.67                            \\
{ViT-B/16}                          & Keyword-Prompt                                    & 0.858                     & 0.855                         & -0.38                            & 0.910                     & 0.900                         & -1.11                            & 0.942                     & 0.940                         & -0.20                            \\
                                                 & ConceptNet-Prompt~\cite{pu2024learning}                                 & 0.855                     & 0.850                         & -0.57                            & 0.894                     & 0.885                         & -0.99                            & 0.925                     & 0.921                         & -0.43                            \\
                                                 & LLM-Prompt (ours)                                       & 0.855                     & 0.852                         & -0.37                            & 0.906                     & 0.900                         & -0.63                            & 0.938                     & 0.929                         & -1.03                            \\

{ViT-L/14}                          & Keyword-Prompt                                    & \textbf{0.876}            & 0.860                         & -1.85                            & \textbf{0.928}            & 0.903                         & -2.64                            & \textbf{0.963}            & \textbf{0.948}                & -1.58                            \\
                                                 & ConceptNet-Prompt~\cite{pu2024learning}                                 & 0.859                     & 0.857                         & -0.19                            & 0.908                     & 0.892                         & -1.76                            & 0.939                     & 0.936                         & -0.38                            \\
                                                 & LLM-Prompt (ours)                                       & 0.868                     & \textbf{0.860}                & -0.97                            & 0.919                     & \textbf{0.904}                & -1.64                            & 0.947                     & 0.933                         & -1.54                            \\
{ViT-L/14@336px}                     & Keyword-Prompt                                    & 0.871                     & \textbf{0.860}                & -1.32                            & 0.921                     & 0.899                         & -2.34                            & 0.953                     & 0.939                         & -1.48                            \\
                                                 & ConceptNet-Prompt~\cite{pu2024learning}                                 & 0.860                     & 0.856                         & -0.46                            & 0.902                     & 0.886                         & -1.76                            & 0.943                     & 0.935                         & -0.79                            \\
                                                 & LLM-Prompt (ours)                                       & 0.867                     & 0.859                         & -0.96                            & 0.912                     & 0.902                         & -1.09                            & 0.945                     & 0.927                         & -1.98                           \\ \hline
\end{tabular}
}
\label{tbl:xai_vad_cat_ucf_f1}
\vspace*{-0.5\baselineskip}
\end{table*}

\subsection{Further Results for ZSAM}
\label{sec:zsam_appendix}

Figure~\ref{fig:xai_vad_heatmap_simuletic_weapon_knife_multi_mask_methodss_lineplot} depicts performance across different XAI heatmap and AN masking methods, demonstrating the superiority of our PPA and GDSN approaches across different settings.
% We recommend considering the sequence of video frames (tracking-based) instead of a single image and a human-in-the-loop to improve the performance further after AN.

Figure~\ref{fig:xai_vad_heatmap_simuletic_weapon_knife_all_map} portrays ZSAM on sample images, visually comparing the XAI maps in localizing alert weapons. 
Grad-ECLIP struggles to localize distant or small threats, e.g., Scene6\_4.png and Knife\_scenario\_98.png. 
The heatmap improves with score normalization, while map score remains sensitive to object size. The map score can become miniature and sometimes miss completely (e.g., Scene6\_4.png). 
Grad-ECLIP with our PPA substantially enhances localization with sanitized heatmaps across varying object sizes and preserves its robustness after AN.

Figure~\ref{fig:xai_vad_heatmap_simuletic_weapon_knife_occluded_map} illustrates weapon occlusion after AN with different person segmentation models. Our improved GDSN detects occluded weapons that overlap the person's body, unlike YOLO and RF-DETR models.

Some of the pitfalls of ZSAM on the Simuletic datasets can be attributed to annotation inaccuracy and occlusions. The Simuletic datasets include annotations with extended boxes beyond the target, including holding hands and across occlusions, whereas the ZSAM boxes are localized in high-saliency regions (as shown in Fig.~\ref{fig:xai_vad_heatmap_simuletic_weapon_knife_map_bboxes_mismatch}). 

\begin{figure*}[!t]
\centering
% \begin{subfigure}[]{0.75\linewidth}
% \centering
% \includegraphics[width=1\linewidth]{images/simuletic_data/xai_clip_ad_gunholding_map_ap_ar.jpg} 
% \label{fig:xai_vad_heatmap_simuletic_weapon_multi_mask_methods_lineplot}
% \caption{}
% \end{subfigure}
% \begin{subfigure}[]{0.75\linewidth}
% \centering
% \includegraphics[width=1\linewidth]{images/simuletic_data/xai_clip_ad_knifeholding_map_ap_ar.jpg} 
% \label{fig:xai_vad_heatmap_simuletic_knife_multi_mask_methodss_lineplot}
% \caption{}
% \end{subfigure}

\begin{subfigure}[]{0.38\linewidth}
\centering
\includegraphics[width=1\linewidth]{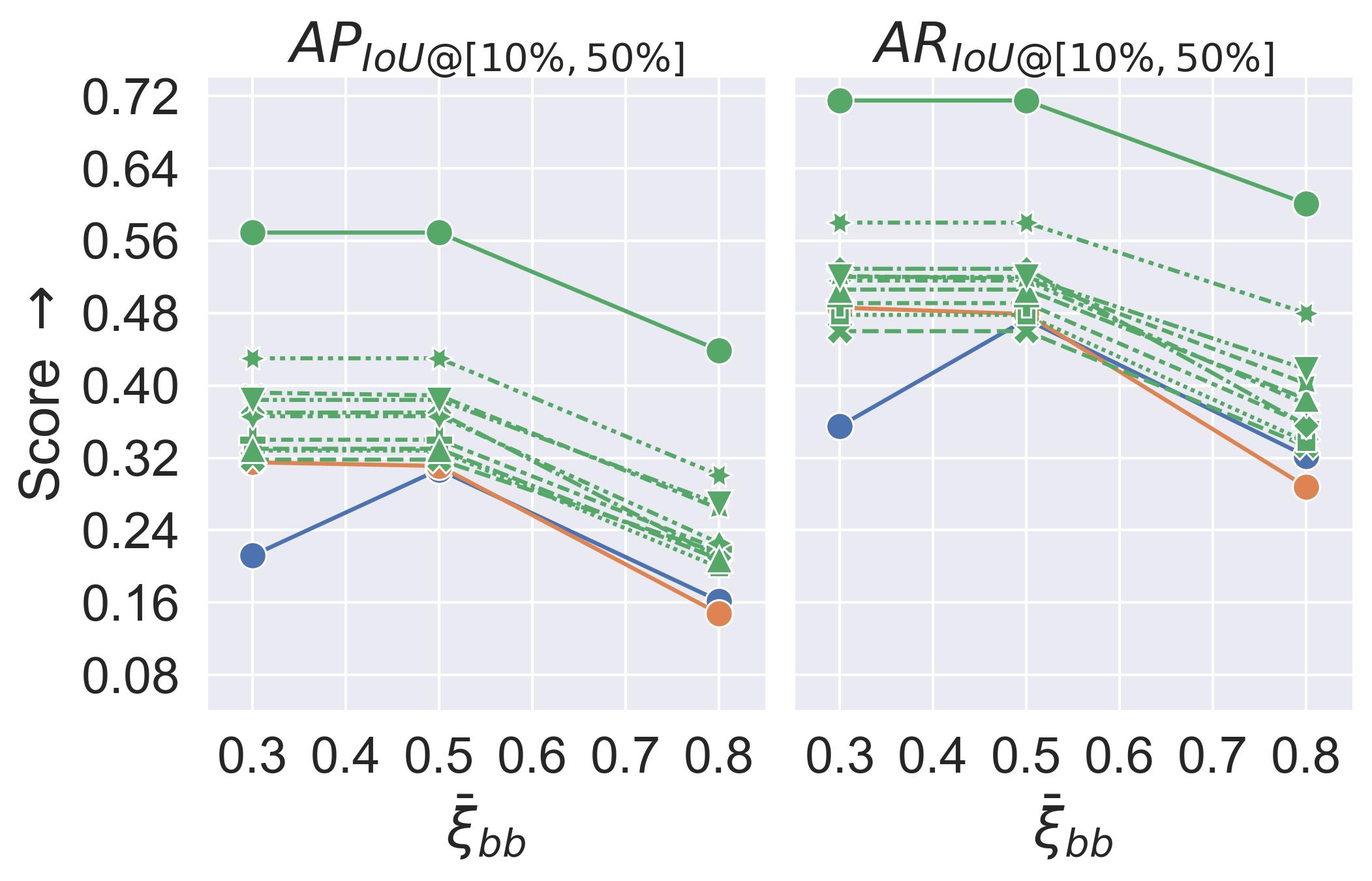} 
\label{fig:xai_vad_heatmap_simuletic_weapon_multi_mask_methods_lineplot}
\caption{}
\end{subfigure}
\begin{subfigure}[]{0.61\linewidth}
\centering
\includegraphics[width=1\linewidth]{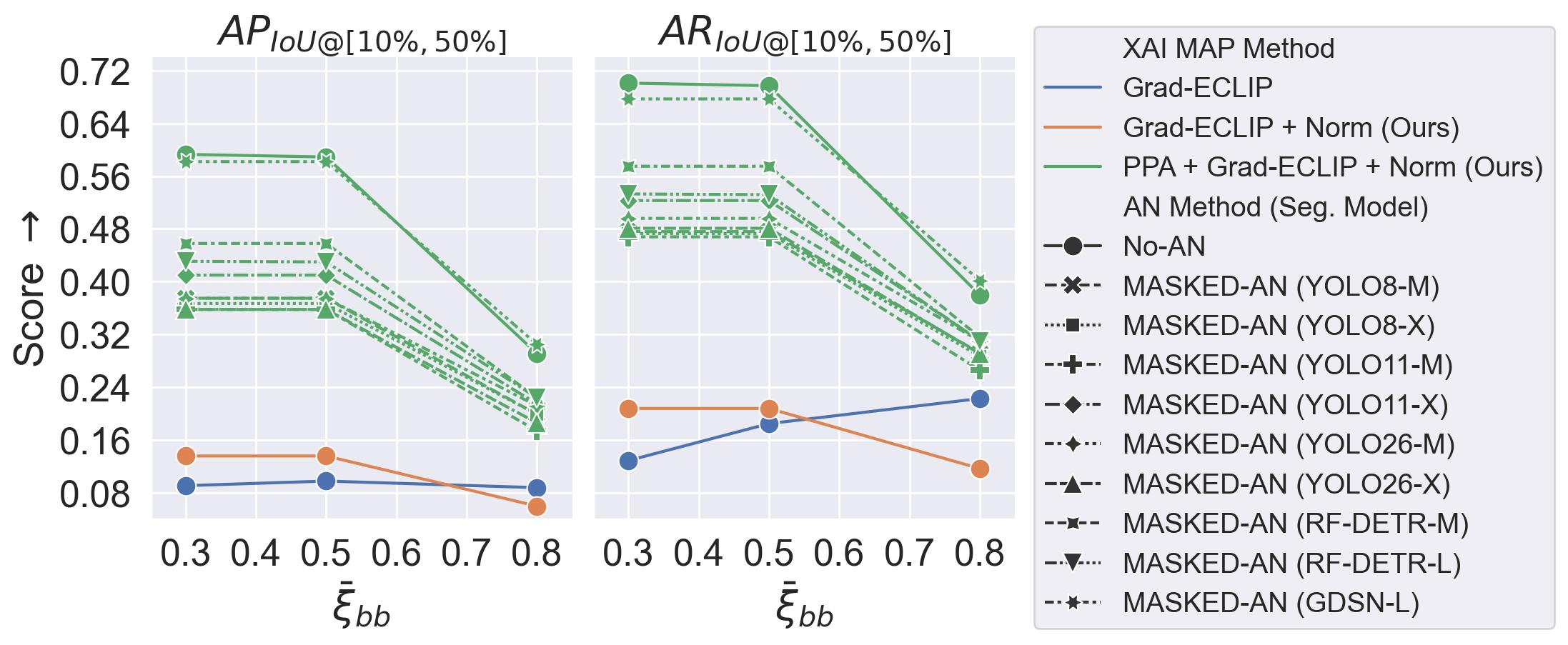} 
\label{fig:xai_vad_heatmap_simuletic_knife_multi_mask_methodss_lineplot}
\caption{}
\end{subfigure}
\caption{\textupdate{Performance of ZSAM using different heatmap-to-bounding-box thresholds $\bar{\xi}_\text{bb}$ on Simuletic CCTV a) Weapon~\cite{simuletic_weapon_2025} and b) Knife datasets~\cite{simuletic_knife_2025}. 
The plots demonstrate the superiority of our Grad-ECLIP with PPA (PPA + Grad-ECLIP + Norm) and GDSN approaches. 
Optimal performance at $\bar{\xi}_\text{bb} = 0.5$.
}}
\label{fig:xai_vad_heatmap_simuletic_weapon_knife_multi_mask_methodss_lineplot}
\vspace*{-0.5\baselineskip}
\end{figure*}

\begin{figure*}[hb] \small
\centering
\setlength{\kw}{2.0cm}
\begin{subfigure}[]{1\textwidth}
\centering
% \begin{tabular}{m{1.5cm}K{\kw}K{\kw}K{\kw}K{\kw}K{\kw}K{\kw}}
\begin{tabular}{@{}K{0.7cm}K{2.5cm}@{}K{2.5cm}K{2.5cm}@{}K{2.5cm}K{2.5cm}@{}K{2.5cm}}
 & \multicolumn{2}{c}{Grad-ECLIP~\cite{chenyang_gradeclip}} & \multicolumn{2}{c}{Grad-ECLIP + Norm (Ours)} & \multicolumn{2}{c}{PPA + Grad-ECLIP + Norm (Ours)} \\
 & No-AN        & MASKED-AN       & No-AN              & MASKED-AN              & No-AN                     & MASKED-AN  \\
% \end{tabular}
\small \rotatebox[origin=l]{90}{\parbox{2.3cm}{\centering Scene5\_7.png}} & \includegraphics[width=1\linewidth]{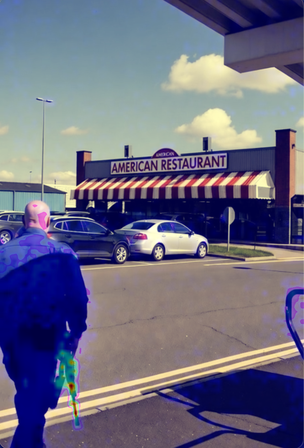} & 
\includegraphics[width=1\linewidth]{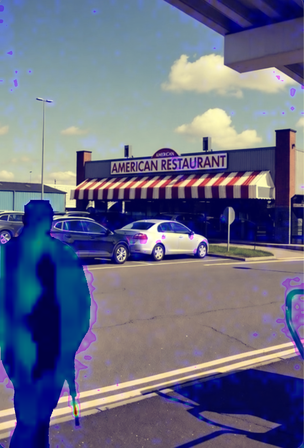} & 
\includegraphics[width=1\linewidth]{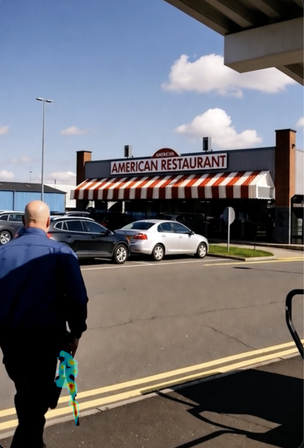} & \includegraphics[width=1\linewidth]{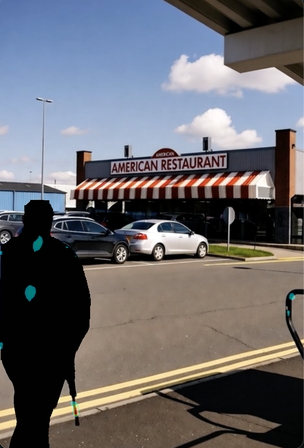} & 
\includegraphics[width=1\linewidth]
{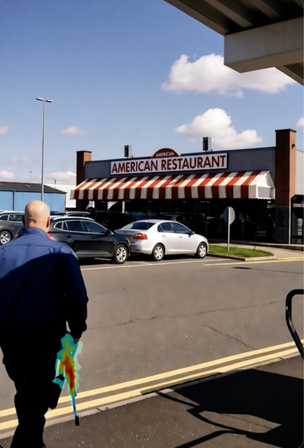} & 
\includegraphics[width=1\linewidth]{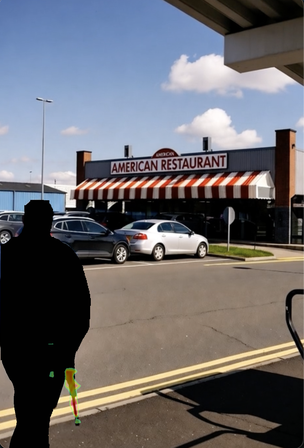} 
\\ 
\small \rotatebox[origin=l]{90}{\parbox{2.3cm}{\centering Scene6\_4.png}} & \includegraphics[width=1\linewidth]{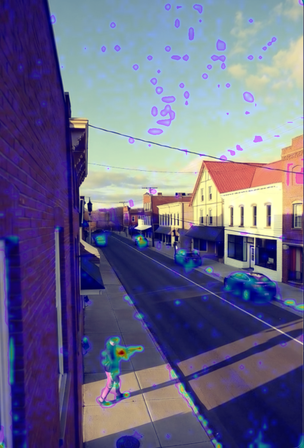} & 
\includegraphics[width=1\linewidth]{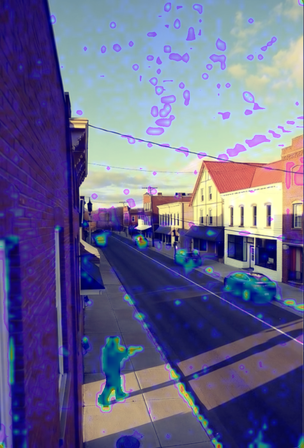} & 
\includegraphics[width=1\linewidth]{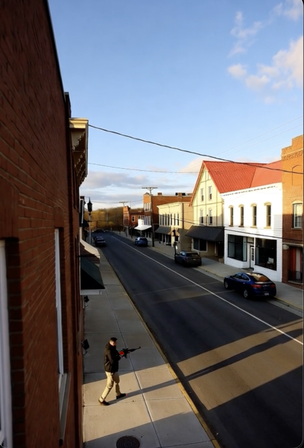} & \includegraphics[width=1\linewidth]{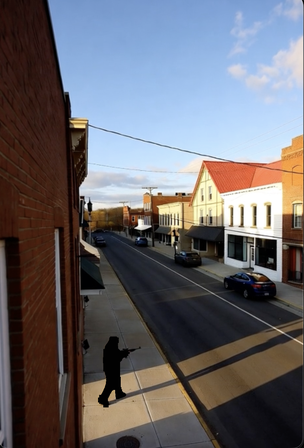} & 
\includegraphics[width=1\linewidth]
{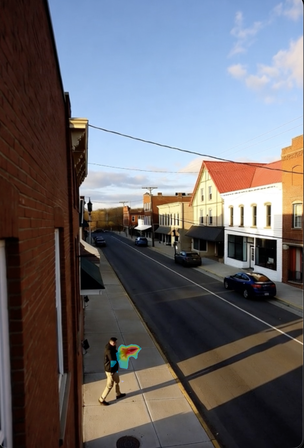} & 
\includegraphics[width=1\linewidth]{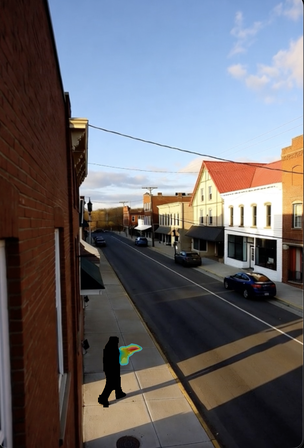} 
\\ 

\small \rotatebox[origin=l]{90}{\parbox{2.3cm}{\centering Knife\_scenario\_98.png}} & \includegraphics[width=1\linewidth]{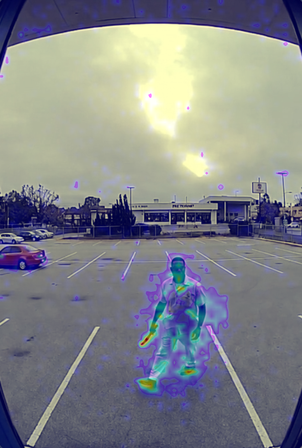} & 
\includegraphics[width=1\linewidth]{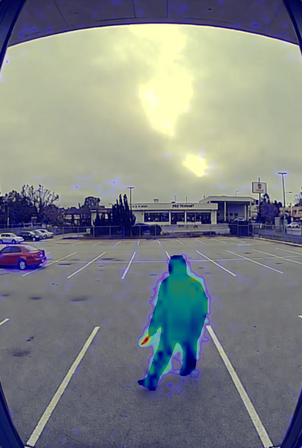} & 
\includegraphics[width=1\linewidth]{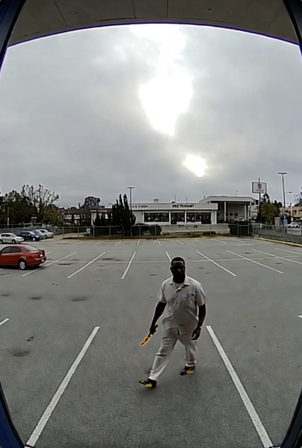} & \includegraphics[width=1\linewidth]{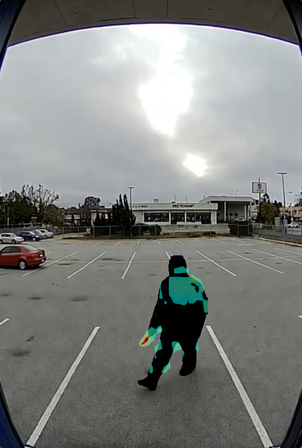} & 
\includegraphics[width=1\linewidth]
{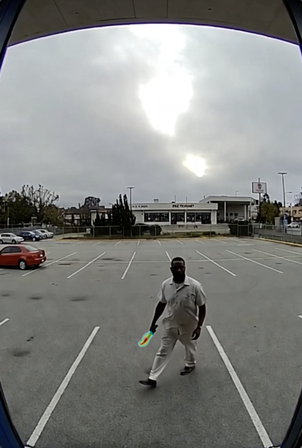} & 
\includegraphics[width=1\linewidth]{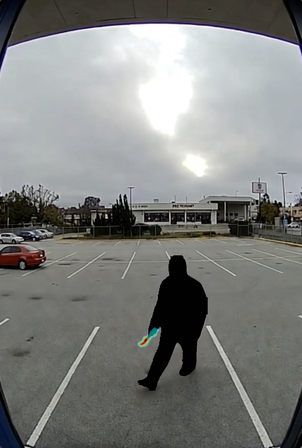} 
\\ 
\small \rotatebox[origin=l]{90}{\parbox{2.3cm}{\centering Knife\_scenario\_31.png}} & \includegraphics[width=1\linewidth]{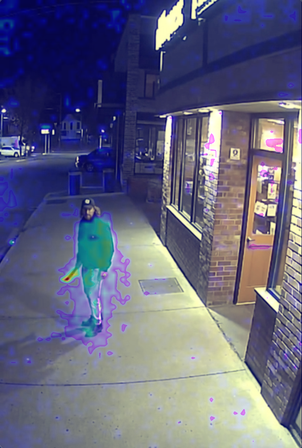} & 
\includegraphics[width=1\linewidth]{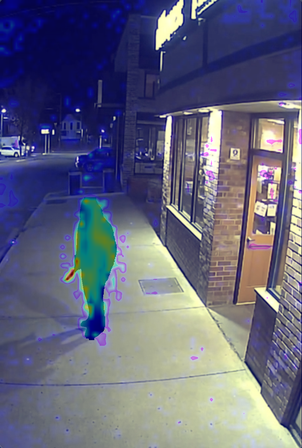} & 
\includegraphics[width=1\linewidth]{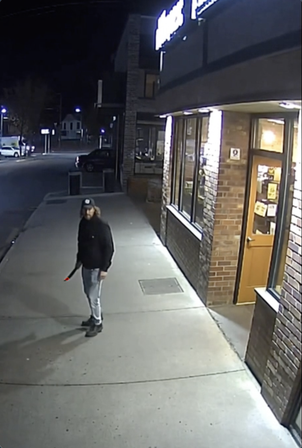} & \includegraphics[width=1\linewidth]{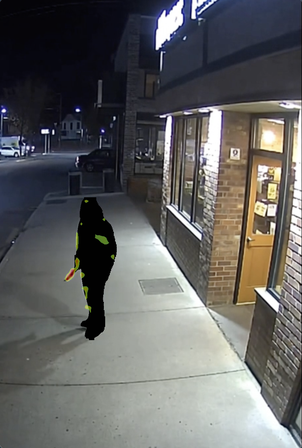} & 
\includegraphics[width=1\linewidth]
{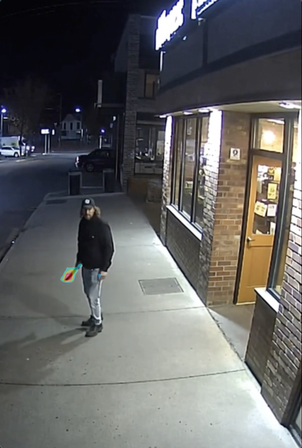} & 
\includegraphics[width=1\linewidth]{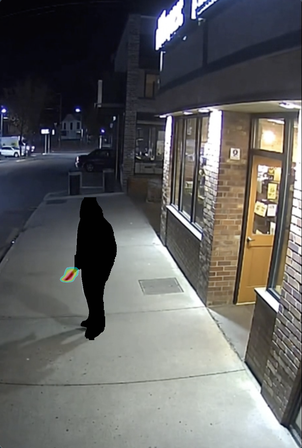} 
\\ 
 \end{tabular}
\vspace{-0.1cm}
\end{subfigure}
\caption{Performance of the ZSAM on the Simuletic CCTV Weapon and Knife datasets~\cite{simuletic_weapon_2025, simuletic_knife_2025} using the ViT-L/14 CLIP model with LLM-enhanced prompt embedding for gun holding and knife holding, respectively. The segmentation model for the MASKED-AN is YOLO8-M.
% Thresholds: $\xi_\text{min}=0.05, \xi_\text{max}=0.50, \bar{\xi}_\text{min}=0.30$.
}
\label{fig:xai_vad_heatmap_simuletic_weapon_knife_all_map}
\vspace*{-0.5\baselineskip}
\end{figure*}

% weapon-gun- and knife ulti-mask-models
\begin{figure*}[hb] \small
\centering
\setlength{\kw}{2.0cm}
\begin{subfigure}[]{1\textwidth}
\centering
\small
% \begin{tabular}{@{}K{0.7cm}K{2.0cm}@{}K{2.0cm}@{}K{2.0cm}@{}K{2.0cm}@{}K{2.0cm}@{}K{2.0cm}@{}K{2.0cm}@{}K{2.0cm}@{}K{2.0cm}@{}K{2.0cm}@{}}
 \begin{tabular}{@{}K{0.05cm}K{1.75cm}@{}K{1.75cm}@{}K{1.75cm}@{}K{1.75cm}@{}K{1.75cm}@{}K{1.75cm}@{}K{1.75cm}@{}K{1.75cm}@{}K{1.75cm}@{}K{1.75cm}@{}}
& \multicolumn{1}{c}{\small No-AN} & \multicolumn{9}{c}{\small MASKED-AN} \\
& -- & YOLO8-M & YOLO8-X & YOLO11-M & YOLO11-X& YOLO26-M & YOLO26-X & RF-DETR-M & RF-DETR-L & GDSN (Ours) \\
% trim left bottom right top
\small \rotatebox[origin=l]{90}{\parbox{2.3cm}{\centering Scene5\_4.png}} & \includegraphics[trim={3.5cm 1cm 4cm 6cm}, clip, width=1\linewidth]{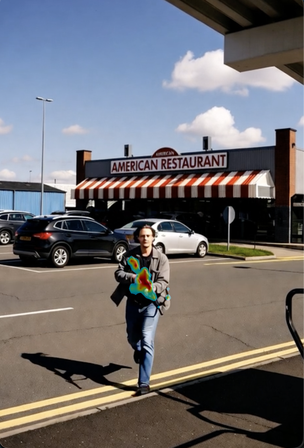} & 
\includegraphics[trim={3.5cm 1cm 4cm 6cm}, clip, width=1\linewidth]{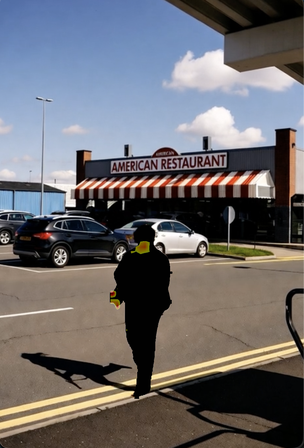} & 
\includegraphics[trim={3.5cm 1cm 4cm 6cm}, clip, width=1\linewidth]{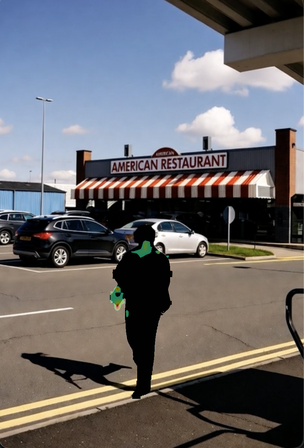} & 
\includegraphics[trim={3.5cm 1cm 4cm 6cm}, clip, width=1\linewidth]{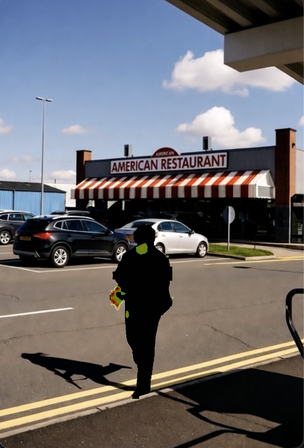} & 
\includegraphics[trim={3.5cm 1cm 4cm 6cm}, clip, width=1\linewidth]{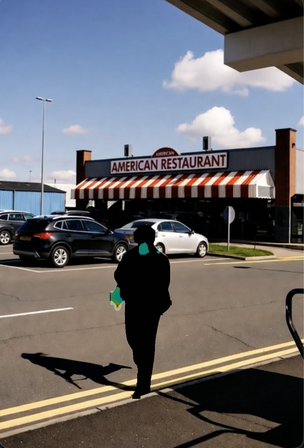} & 
\includegraphics[trim={3.5cm 1cm 4cm 6cm}, clip, width=1\linewidth]{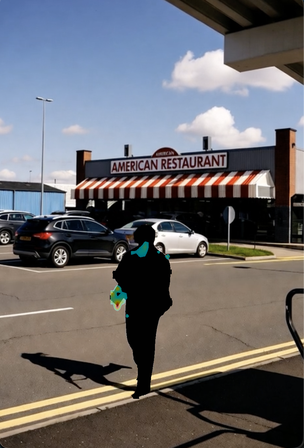} & 
\includegraphics[trim={3.5cm 1cm 4cm 6cm}, clip, width=1\linewidth]{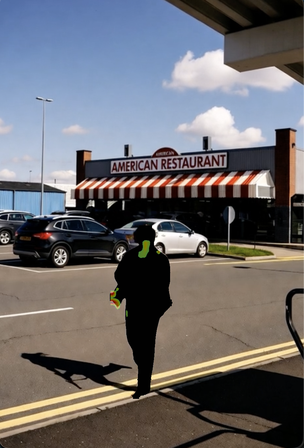} & 
\includegraphics[trim={3.5cm 1cm 4cm 6cm}, clip, width=1\linewidth]{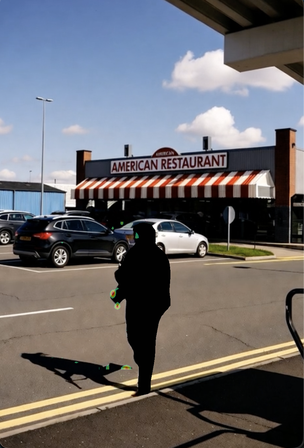} & 
\includegraphics[trim={3.5cm 1cm 4cm 6cm}, clip, width=1\linewidth]{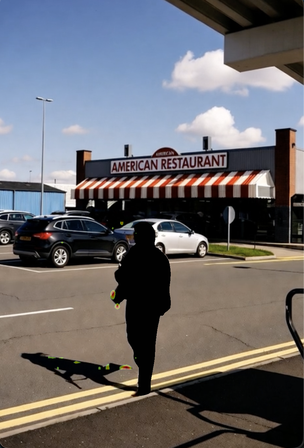} & 
\includegraphics[trim={3.5cm 1cm 4cm 6cm}, clip, width=1\linewidth]{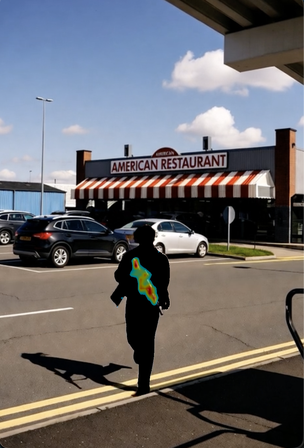}
\\
\small \rotatebox[origin=l]{90}{\parbox{2.3cm}{\centering Scene3\_19.png}} & \includegraphics[trim={4cm 3cm 3cm 4cm}, clip, width=1\linewidth]{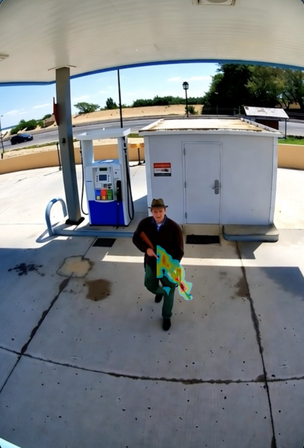} & 
\includegraphics[trim={4cm 3cm 3cm 4cm}, clip, width=1\linewidth]{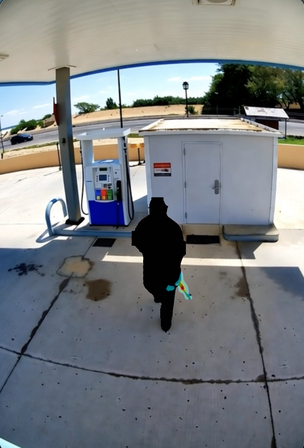} & 
\includegraphics[trim={4cm 3cm 3cm 4cm}, clip, width=1\linewidth]{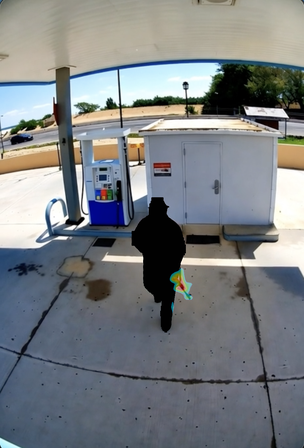} & 
\includegraphics[trim={4cm 3cm 3cm 4cm}, clip, width=1\linewidth]{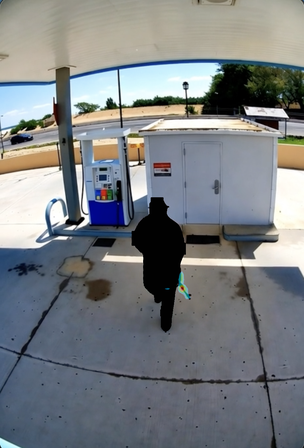} & 
\includegraphics[trim={4cm 3cm 3cm 4cm}, clip, width=1\linewidth]{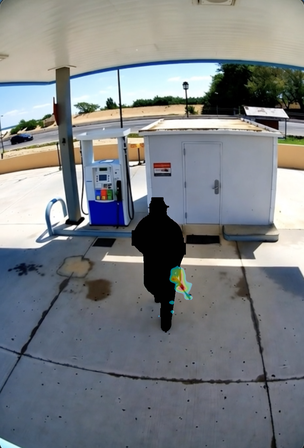} & 
\includegraphics[trim={4cm 3cm 3cm 4cm}, clip, width=1\linewidth]{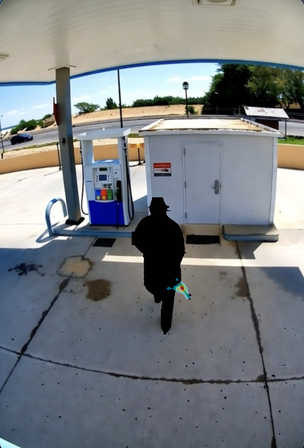} & 
\includegraphics[trim={4cm 3cm 3cm 4cm}, clip, width=1\linewidth]{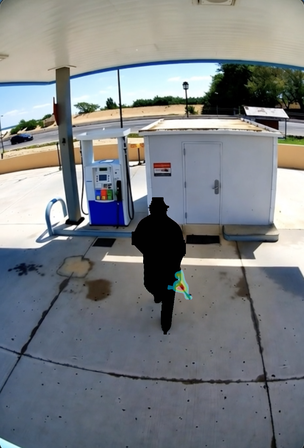} & 
\includegraphics[trim={4cm 3cm 3cm 4cm}, clip, width=1\linewidth]{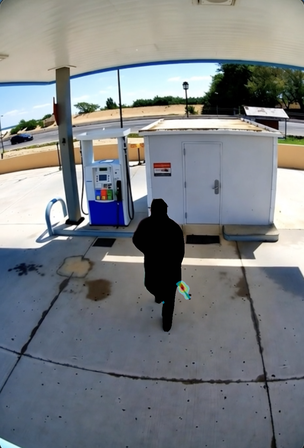} & 
\includegraphics[trim={4cm 3cm 3cm 4cm}, clip, width=1\linewidth]{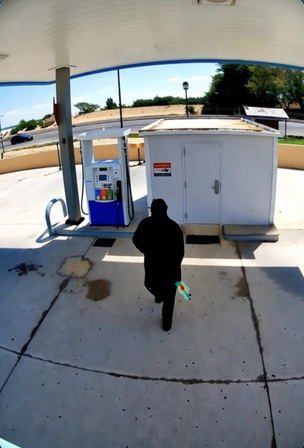} & 
\includegraphics[trim={4cm 3cm 3cm 4cm}, clip, width=1\linewidth]{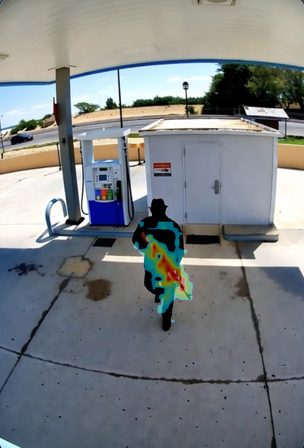}
\\
\\
\small \rotatebox[origin=l]{90}{\parbox{2.3cm}{\centering Knife\_scenario\_53.png}} & \includegraphics[trim={0cm 0cm 4cm 3cm}, clip, width=1\linewidth]{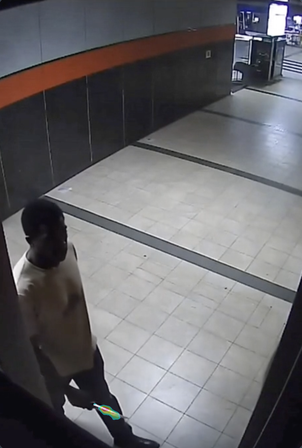} & 
\includegraphics[trim={0cm 0cm 4cm 3cm}, clip, width=1\linewidth]{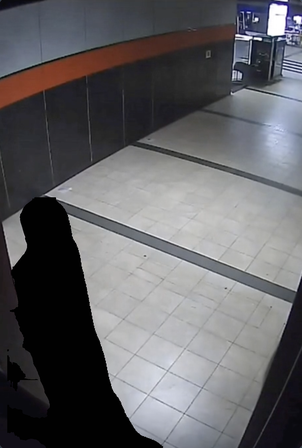} & 
\includegraphics[trim={0cm 0cm 4cm 3cm}, clip, width=1\linewidth]{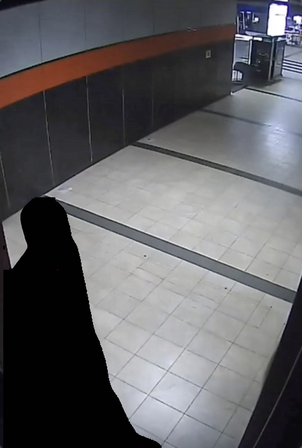} & 
\includegraphics[trim={0cm 0cm 4cm 3cm}, clip, width=1\linewidth]{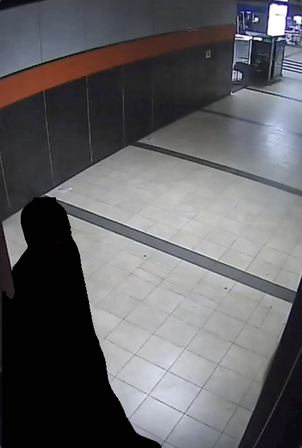} & 
\includegraphics[trim={0cm 0cm 4cm 3cm}, clip, width=1\linewidth]{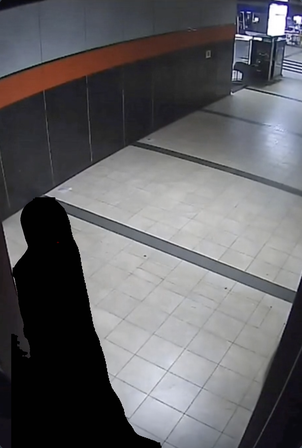} & 
\includegraphics[trim={0cm 0cm 4cm 3cm}, clip, width=1\linewidth]{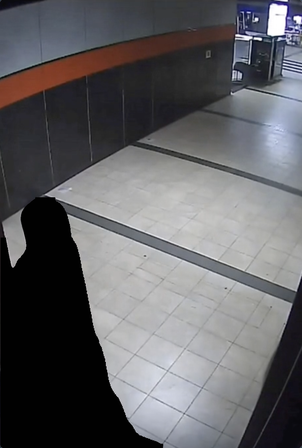} & 
\includegraphics[trim={0cm 0cm 4cm 3cm}, clip, width=1\linewidth]{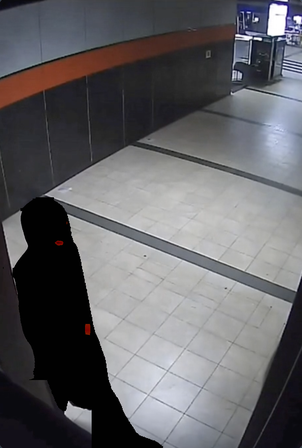} & 
\includegraphics[trim={0cm 0cm 4cm 3cm}, clip, width=1\linewidth]{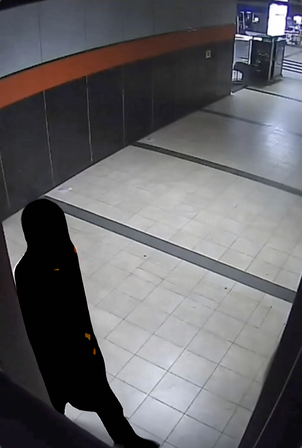} & 
\includegraphics[trim={0cm 0cm 4cm 3cm}, clip, width=1\linewidth]{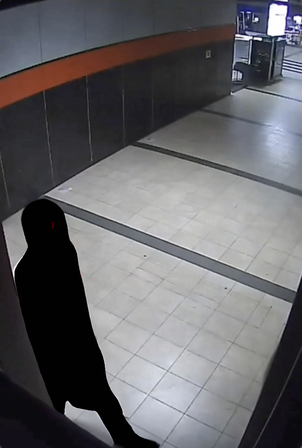} & 
\includegraphics[trim={0cm 0cm 4cm 3cm}, clip, width=1\linewidth]{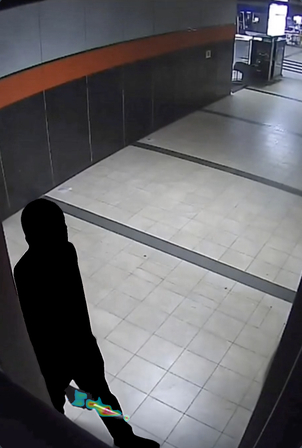}

\end{tabular}

\end{subfigure}
\caption{Weapon occlusion after AN with different person segmentation models: ZSAM with PPA + Grad-ECLIP + Norm on Simuletic CCTV Weapon and Knife datasets~\cite{simuletic_weapon_2025, simuletic_knife_2025}.
% Thresholds:  $\xi_\text{min}=0.05, \xi_\text{max}=0.50, \bar{\xi}_\text{min}=0.30,~\text{and}~\bar{\xi}_\text{bb}=0.5$. 
The YOLO and RF-DETR models conceal objects, such as weapons (which are not in their training MS-COCO datasets), when they overlap with the person's body. Our GDSN with negative segmentation mitigates the overlapping challenge. The images are cropped to enhance visualization.
}
\label{fig:xai_vad_heatmap_simuletic_weapon_knife_occluded_map}
\vspace*{-0.5\baselineskip}
\end{figure*}

\begin{figure*}[hb] 
\centering
\setlength{\kw}{2.0cm}
\begin{subfigure}[]{1\textwidth}
\centering
\begin{tabular}{@{}K{6.6cm}@{}K{6.6cm}@{}K{4.5cm}}
Scene2\_11.png      & Scene2\_24.png        & Scene4\_24.png            \\
\includegraphics[width=1\linewidth]{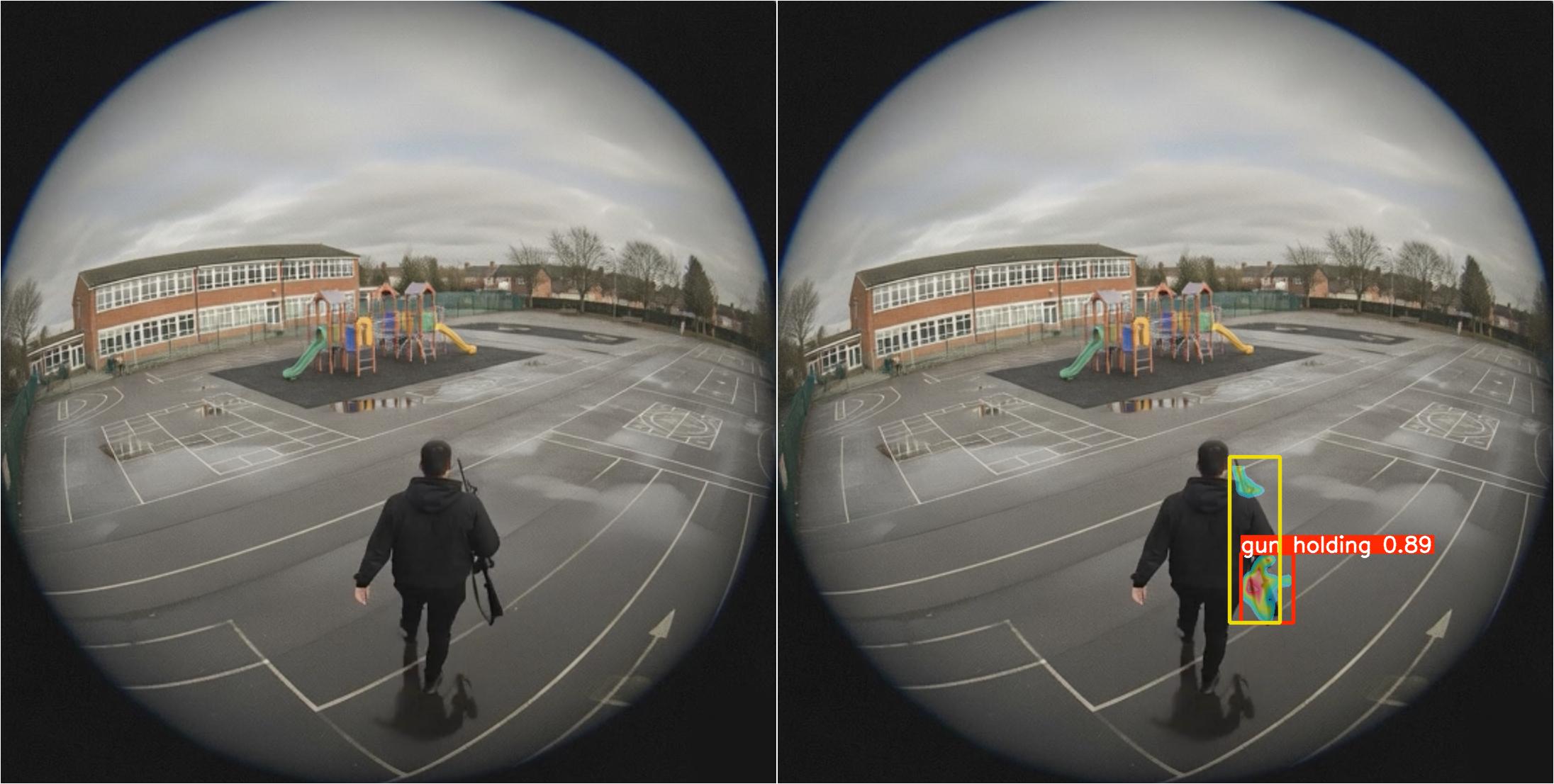} & 
\includegraphics[width=1\linewidth]{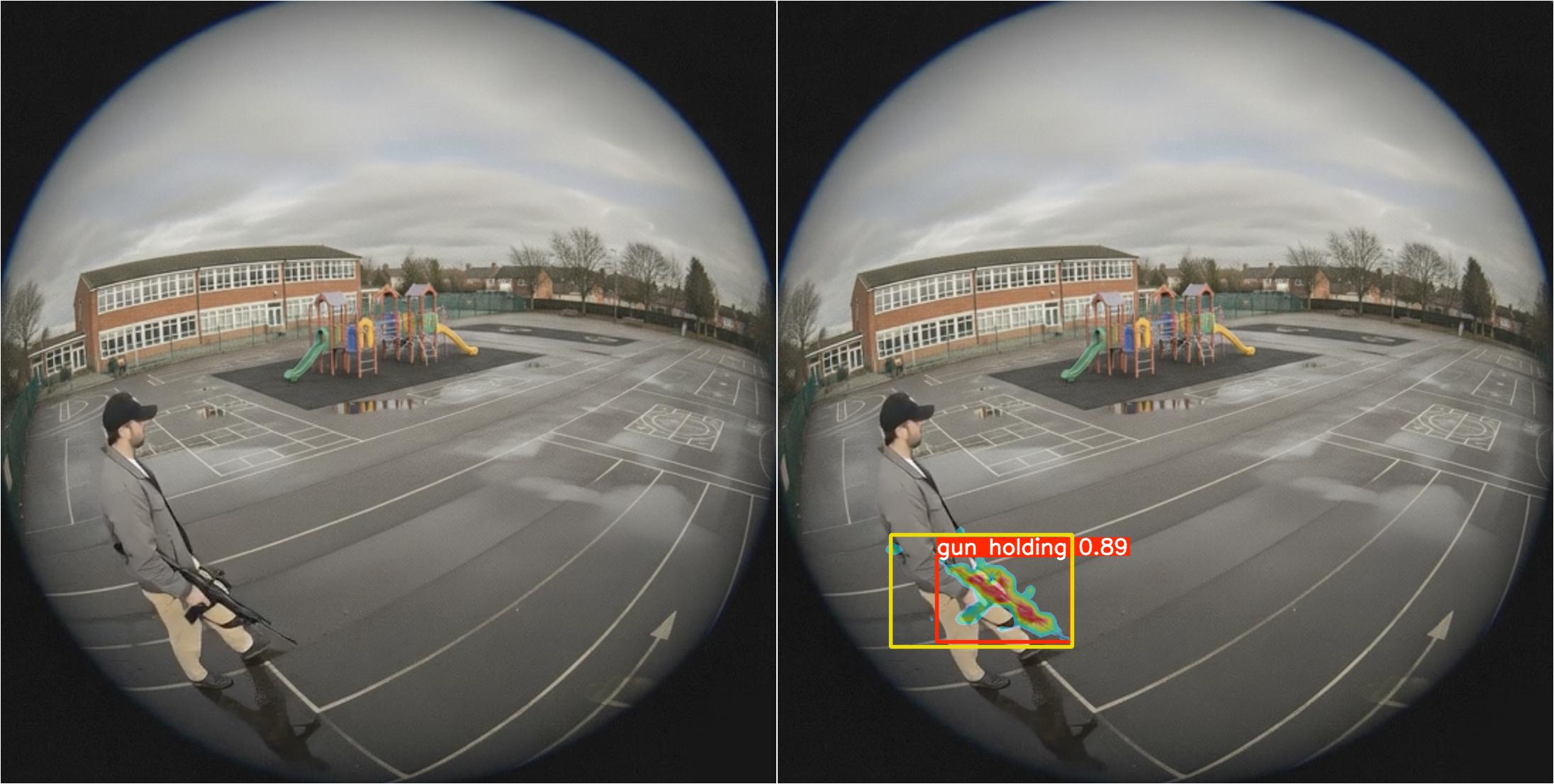} & \includegraphics[width=1\linewidth]{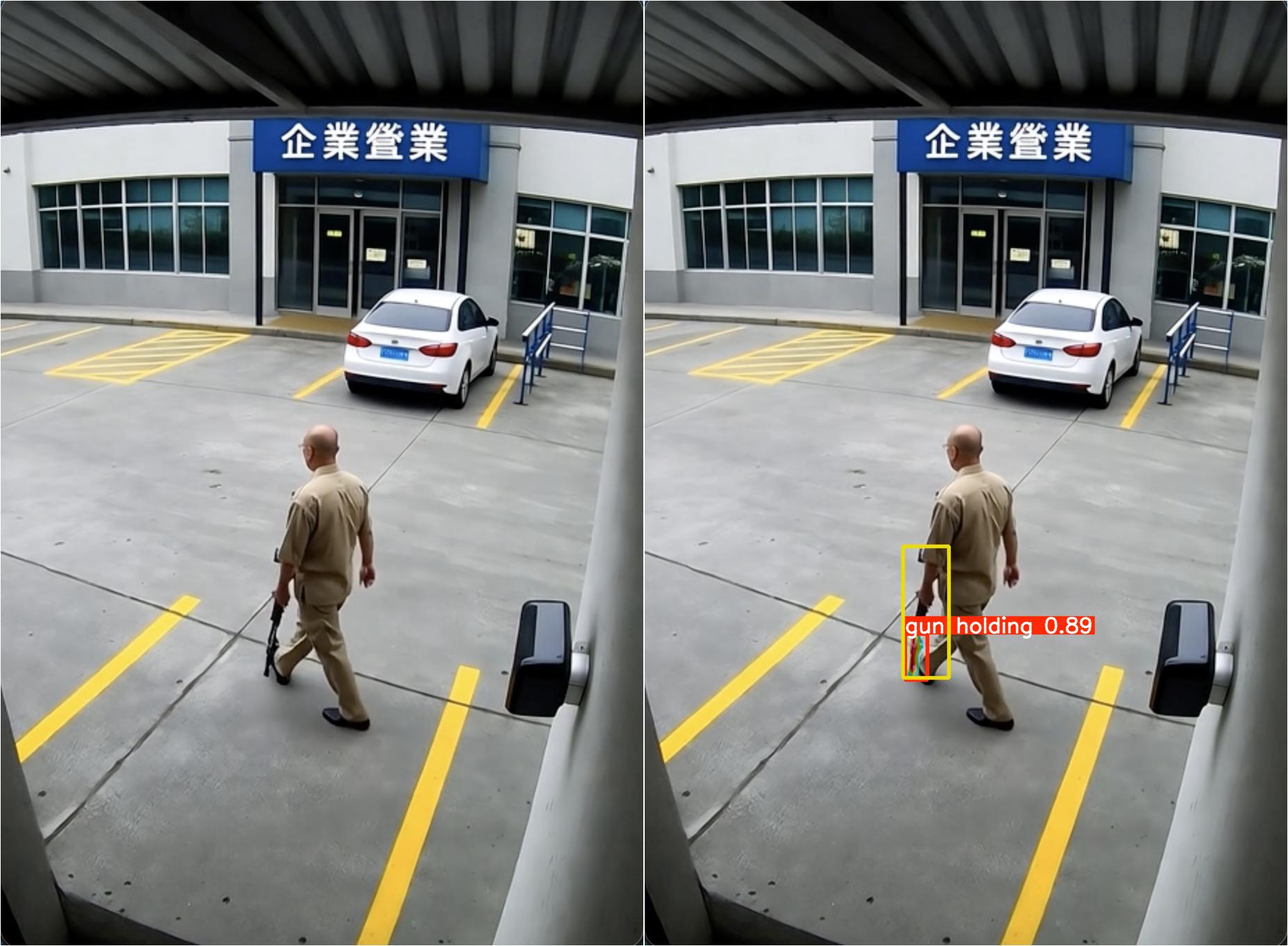} 
\\ 
\end{tabular}
% \begin{tabular}{@{}K{4.5cm}@{}K{4.5cm}@{}K{4.5cm}}
% Knife\_scenario\_42.png      & Knife\_scenario\_58.png         & Knife\_scenario\_78.png             \\
% \includegraphics[width=1\linewidth]{images/simuletic_data/Knife_scenario_42_isauto_Trueoriginal_ispatched_False_448_gradcam.jpg} & 
% \includegraphics[width=1\linewidth]{images/simuletic_data/Knife_scenario_58_isauto_Trueoriginal_ispatched_False_448_gradcam.jpg} & \includegraphics[width=1\linewidth]{images/simuletic_data/Knife_scenario_78_isauto_Trueoriginal_ispatched_False_448_gradcam.jpg} 
% \\ 
% \end{tabular}
% \vspace{-0.1cm}
\end{subfigure}
\caption{Ground truth (yellow) vs. ZSAM-predicted bounding boxes (red) on Simuletic CCTV datasets~\cite{simuletic_weapon_2025, simuletic_knife_2025}. 
The yellow boxes cover extended areas, including occluded regions, while the red boxes focus on high-saliency regions.
}
\label{fig:xai_vad_heatmap_simuletic_weapon_knife_map_bboxes_mismatch}
\vspace*{-0.5\baselineskip}
\end{figure*}

% \subsection{Limitations and Future Recommendations}
% % We recommend considering the sequence of video frames (tracking-based) instead of a single image and a human-in-the-loop to improve the performance further after AN. 

% \begin{itemize}
%     \item Video frame and patch compression of similar regions to speed up ZS XAI. Recent efforts in video understanding VLM models, such as VideoLLaMA~\cite{zhang2025videollama}, compress video frames via token similarity. 
%     Such approaches can speed up inference, with the preprocessing load trade-off remaining lower.
%     \item Expanding normal category prompts, using multiple distinct normal events, to reduce potential false positives. For example, adding crowd-context disambiguation can reduce false positives for riot and group fight categorizations due to busy streets, transport hubs, shopping malls, and events/queues. 
%     Many false alerts occur when there is confusion between dense crowds and fighting, congestion and panic, group motion and robbery or assault, and people running and fleeing.
%     % \item Improving AN masking and XAI map generation approaches that consider the sequence of video frames (tracking-based) instead of a single frame to improve the performance further. 
%     \item \textupdate{LLM-as-Judge for ZSAD evaluation would provide a more accurate assessment for VLM-generated anomaly descriptions. Existing tools such as BLEU and BERTScore are sensitive to words and text length, and LLM-based eval can mitigate this and capture semantic similarity robustly. 
%     }
    
% \end{itemize}

\subsection{Further Results for ZSAD}
\label{sec:zsam_appendix}

\textupdate{Figure~\ref{fig:xai_vad_vlm_hyper_lineplot_internvl} presents ZSAD of InterVL3.5 VLM model across different hyperparameter settings. Setting $temperature=0$ removes the impact of sampling, delivering consistent results across different $top\_k$ and $top\_p$. Increasing $N_\text{vlm}$ improves accuracy at lower values but can deteriorate at higher values, possibly due to VLM context window saturation. A potential mitigation mechanism for this challenge could be to presample the input video using lower FPS to capture frames from a longer time-scale span. This improve information content by preventing similar frames from filling the context window.
}

\begin{figure*}[!t]
\centering
\begin{subfigure}[]{1.0\linewidth}
\centering
\includegraphics[width=1\linewidth]{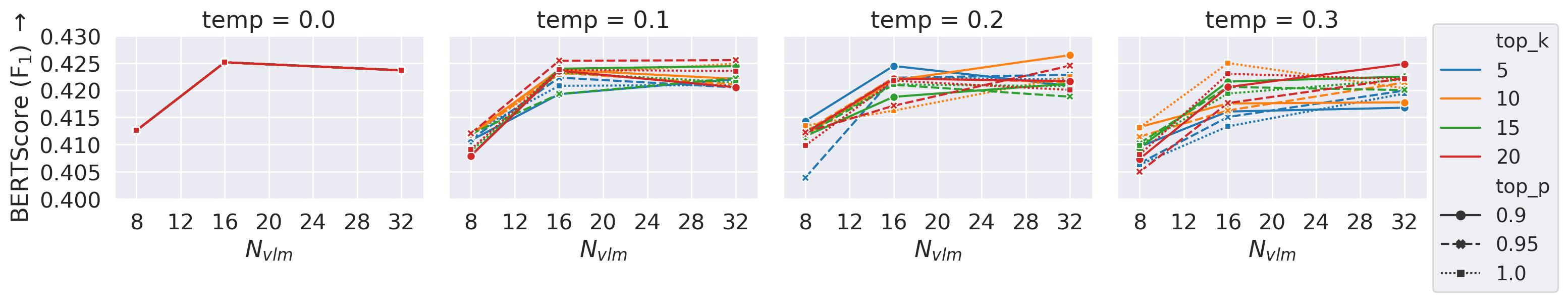} 
\label{fig:xai_vad_vlm_hyper_xd_lineplot_internvl}
\caption{}
\end{subfigure}
\begin{subfigure}[]{1.0\linewidth}
\centering
\includegraphics[width=1\linewidth]{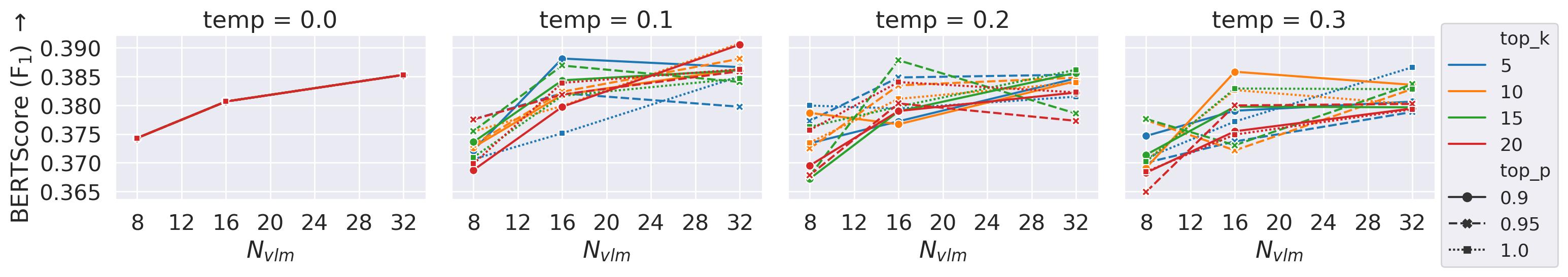} 
\label{fig:xai_vad_vlm_hyper_ucf_lineplot_internvl}
\caption{}
\end{subfigure}
\caption{\textupdate{Performance of ZSAD using VLM InternVL3.5 (8B) model with different hyperparameters: a) HIVAU-70k Dataset [XD Violence] and b) HIVAU-70k Dataset [UCF-Crime]. 
}}
\label{fig:xai_vad_vlm_hyper_lineplot_internvl}
\vspace*{-0.5\baselineskip}
\end{figure*}

\end{document}